\documentclass{article}
\usepackage{arxiv}
\usepackage{libertine} 
\usepackage{lmodern}        
\usepackage{hyperref}       
\usepackage{url}            
\usepackage{booktabs}       
\usepackage{amsfonts}       
\usepackage{nicefrac}       
\usepackage{microtype}      
\usepackage{lipsum}		
\usepackage{graphicx}
\usepackage[square, sort, comma, numbers]{natbib}
\usepackage{doi}    
\usepackage{adjustbox}
\usepackage{amsmath}
\usepackage{geometry}
\usepackage{multirow}
\usepackage{lscape}
\usepackage{rotating}
\usepackage{array}
\usepackage{float}
\usepackage{pdflscape} 
\usepackage{tabularx}
\usepackage{longtable}
\usepackage{makecell, cellspace, caption}
\usepackage[table]{xcolor}
\usepackage{enumerate}
\usepackage{enumitem}
\usepackage{titlesec}
\usepackage{subcaption}
\usepackage{amssymb}

\title{ \emph{Pixels} to \emph{Prose}: Understanding the art of Image Captioning}

\author{ 
    \hspace{1mm}Hrishikesh Singh,\hspace{6mm}
    \hspace{1mm}Aarti Sharma,\thanks{Shared First Authorship -- Equal contribution}
    \hspace{6mm}\hspace{1mm}Dr. Millie Pant\\
        Department of Applied Mathematics \& Scientific Computing,\\
	Indian Institute of Technology - Roorkee, India \\
    {
        \href{mailto:hrishikesh.hsk@gmail.com}{\textit{hrishikesh.hsk@gmail.com}}, \href{mailto:sharmaaarti50528@gmail.com}{\textit{sharmaaarti50528@gmail.com}}, \href{mailto:pant.milli@as.iitr.ac.in}{\textit{pant.milli@as.iitr.ac.in}}
    }
}

\renewcommand{\headeright}{Preprint}
\renewcommand{\shorttitle}{}

\hypersetup{
    pdftitle={Pixel to Prose: Understanding the art of Image Captioning},
    pdfsubject={cs.AI, cs.CV, cs.LG},
    pdfauthor={Hrishikesh Singh, Aarti Sharma},
    pdfkeywords={Deep Learning, Image Caption, Literature Review},
}

\begin{document}

\maketitle
\begin{abstract}
{
    In the era of evolving artificial intelligence, machines are increasingly emulating human-like capabilities, including visual perception and linguistic expression. Image captioning stands at the intersection of these domains, enabling machines to interpret visual content and generate descriptive text. This paper provides a thorough review of image captioning techniques, catering to individuals entering the field of machine learning who seek a comprehensive understanding of available options, from foundational methods to state-of-the-art approaches.
    
    Beginning with an exploration of primitive architectures, the review traces the evolution of image captioning models to the latest cutting-edge solutions. By dissecting the components of these architectures, readers gain insights into the underlying mechanisms and can select suitable approaches tailored to specific problem requirements without duplicating efforts. The paper also delves into the application of image captioning in the medical domain, illuminating its significance in various real-world scenarios.
    
    Furthermore, the review offers guidance on evaluating the performance of image captioning systems, highlighting key metrics for assessment. By synthesizing theoretical concepts with practical application, this paper equips readers with the knowledge needed to navigate the complex landscape of image captioning and harness its potential for diverse applications in machine learning and beyond.
}
\end{abstract}

\keywords{Deep Learning \and Image Captioning \and Literature Review}

\section{Introduction}
{
    In this paper, we survey the existing machine-learning literature for the task of image captioning. We present a comprehensive review of the key techniques used in resolving this task. 
    
    Recent works in image captioning which involve deep learning techniques are based on \emph{supervised learning} techniques\cite{vinyals2015tellneuralimagecaption, mao2014deep, Sur2019TPsgtRNT} which uses the paired image caption data to generate image captions or descriptions. We also include \emph{unsupervised} and \emph{reinforcement learning} approaches in our literature review for a holistic coverage of methods for image captioning \cite{9171997,Vinyals_2017, wang2020visual, wang2018show}.

    Since image captioning lies at the intersection of two major domains of machine learning, \emph{i.e.} \emph{computer vision} and \emph{natural language processing} (NLP), a combined architecture is used in the majority of the image captioning models.

    Mainly \emph{encoder-decoder} architecture is used, the \emph{encoder} extracts one or multiple feature vectors from the input image. Extracting the features is an important step as these features provide the context based on which the decoder in the next generative step, generates meaningful captions.\cite{10.1145/3617592}

    The vanilla encoder-decoder architecture comprises \emph{Convolutional Neural Network} (CNN) for visual encoding and \emph{Recurrent Neural Network} (RNN) for the language decoding part, over the years this architecture has seen considerable progress, and the quality, diversity, and richness of captions have improved considerably.

    In place of the simple RNN architectures, more enhanced models such as \emph{Long short-term memory} (LSTM) and \emph{Gated Recurrent Units} (GRUs) were employed. Various attention mechanisms to capture context, spatial, and semantic relations within an input image are used \cite{Zohourianshahzadi_2021,anderson2018bottomuptopdownattentionimage,lu2017knowinglookadaptiveattention, pedersoli2017areasattentionimagecaptioning}. 
    
    Region-aware interaction learning framework\cite{9521159} explicitly models the interplay between image regions and objects using dual-attention, leading to improved contextual understanding and more accurate caption generation.
    
    \emph{Transformers} with self-attention mechanisms, employed in both encoder and decoder modules have enabled the generation of more sophisticated captions \cite{cornia2020meshedmemorytransformerimagecaptioning, luo2021dual}.
    
    With the advancement in AI models and technologies, it is expected that these models will be at par with human intelligence when it comes to processing and generating multi-modal information. Just like humans are able to make sense of the real world by processing the multi-modal information in their surroundings and generating a language or visual response accordingly, in a similar way various vision-language models were developed which have the capability to process inputs from multiple modalities\cite{chen2020uniteruniversalimagetextrepresentation}. The vision-language models are capable of various multi-modal tasks such as \emph{Visual Question Answering} (VQA), image captioning, and video captioning. Thus the shift is from task-specific architectures towards more generalized models which are capable of various multi-modal operations. The language models like \emph{Bidirectional Encoder Representations from Transformers} (BERT)\cite{devlin2019bertpretrainingdeepbidirectional} pre-trained on massive training datasets are used as language decoders, in the vision-language models \cite{gan2022visionlanguagepretrainingbasicsrecent, herdade2020imagecaptioningtransformingobjects, cho2021unifyingvisionandlanguagetaskstext, yang-etal-2011-corpus}.
    
    The image captioning task has various applications in important domains such as healthcare, autonomous vehicles, remote sensing, and helping the visually impaired to retrieve information, to navigate routes, and even to get a sense of their environment. With the machines today imitating human-like intelligence, image captioning is closer to this human-like trait, where a machine is able to ‘see’ the image and generate a caption, which is quite intuitive to a human. This ‘intuition’ capability of a machine being able to identify an image and also describe its contents has huge applications in the healthcare field, where radiologists can use image captioning models to generate detailed reports based on X-rays and MRI scans, deep learning models can generate captions and reports based on various images in various formats, thus it is a huge aid to the domain experts in the healthcare field \cite{selivanov2022medicalimagecaptioninggenerative,9894658}, as they not only get the output as yes/no for a question, whether there is a tumor in an image but also a detailed report explaining the image, this improves the reliability of deep learning models \cite{yuan2019automaticradiologyreportgeneration,8970668}.

    With more high-resolution and large amounts of data being created through social media, we need models that are computationally efficient, scalable, diverse, and accurate. Because for use cases such as autonomous vehicles and social media, the devices are less computationally efficient.

    We also need the models to be more personalized, and human-like, we want them to capture details but also have human-like qualities.
    
    \vspace{+15pt}

    Through this systematic literature review, we aim to answer the following \emph{research questions}:

    {\textbf{RQ1}.\label{rq1}Which image captioning models achieve superior performance for distinct caption types, including those requiring semantic and spatial alignment, and emotional expression?}

    \textbf{RQ2}. What are the approaches in unsupervised, self-supervised, and reinforcement learning?
    
    \textbf{RQ3}. What are the primary datasets utilized for image captioning tasks, what evaluation metrics are commonly employed, how do these metrics assess caption quality, and what are the distinct characteristics of these datasets?
    
    \textbf{RQ4}. Considering the evaluation metrics employed, which image captioning models demonstrate superior performance?
    
    \newpage
    \begin{figure}[H]
        \centering
        \includegraphics[width=1\linewidth]{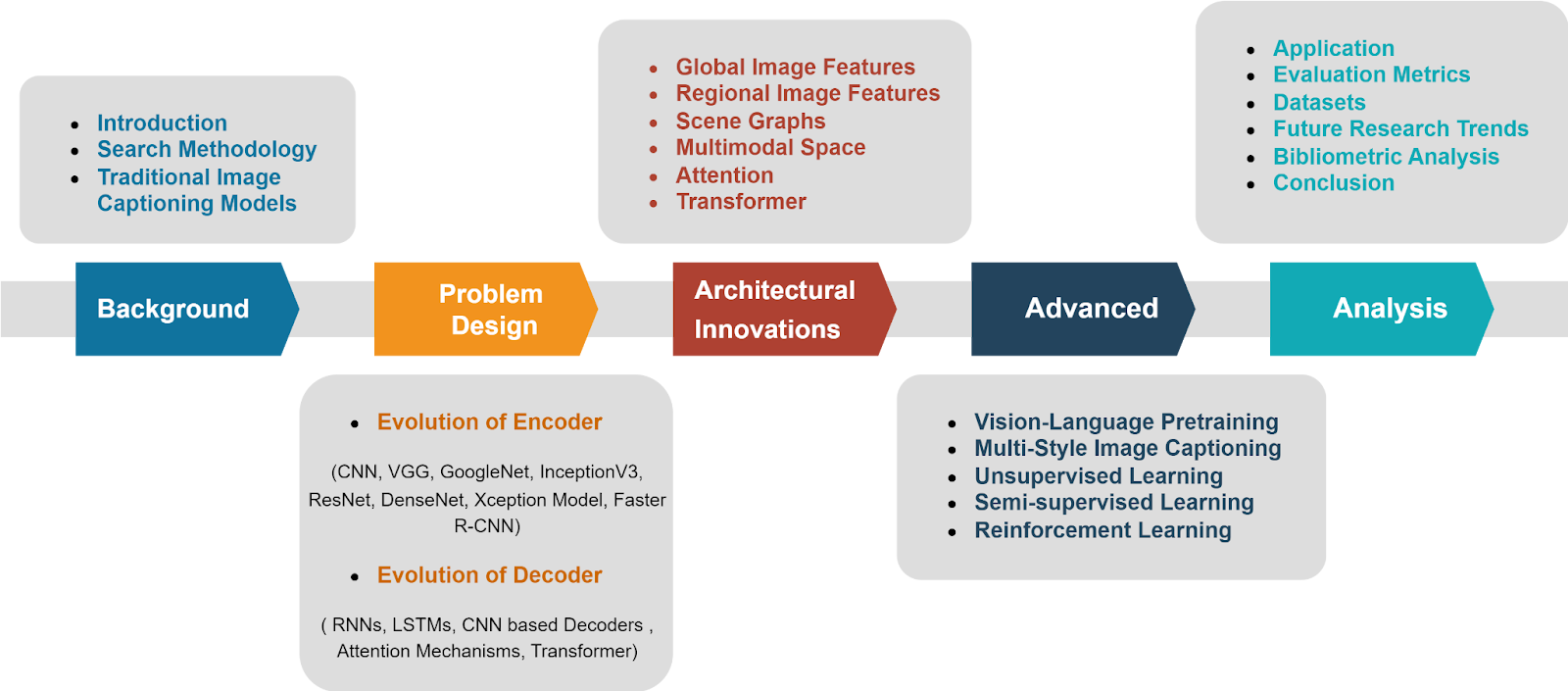}
        \caption{Review Paper Visual Outline}
        \label{fig:paper-overview}
    \end{figure}
    To summarize our contributions are as follows:
    \begin{itemize}[itemsep=3pt, topsep=0pt, partopsep=0pt, parsep=0pt]
        \item We begin with reviewing the evolution of architectures of both encoder and decoder, listing down key CNN based models for visual encoding and key enhancements in language decoders.
        \item We survey various methodologies used to enhance the encoder-decoder architecture such as attention, transformers, scene graphs, and global versus regional features.
        \item We further, comprehensively review the vision-language models which have led to evolved architectures that are capable of handling multiple modalities.
        \item We review various evaluation metrics and datasets used for evaluating the quality of captions generated through various models.
    \end{itemize}
}

\section{Search Methodology}
{
    For the purpose of our review, we collected data from various databases such as \href{https://www.scopus.com/search/form.uri#basic}{\emph{Scopus}}, \href{https://ieeexplore.ieee.org/Xplore/home.jsp}{\emph{IEEE Xplore}}, and \href{https://arxiv.org/}{\emph{arXiv}}, using a combination of relevant keywords such as "\emph{image captioning}", "\emph{deep learning}", "\emph{neural networks}", and "\emph{natural language processing}". Boolean operators were employed to refine the search queries and ensure the retrieval of pertinent literature. Only studies directly related to image captioning, featuring empirical evaluations or significant contributions, were considered for inclusion. Additionally, snowballing techniques were employed to identify additional relevant studies by reviewing the references of selected papers. As a result of this approach we selected 122 papers for the literature review, the bibliometric analysis is presented in section 13 and the \emph{Fig \ref{fig:Search-methodolgy}} depicts the flow diagram of the selection procedure.

    \begin{figure}
        \centering
        \includegraphics[width=0.8\linewidth]{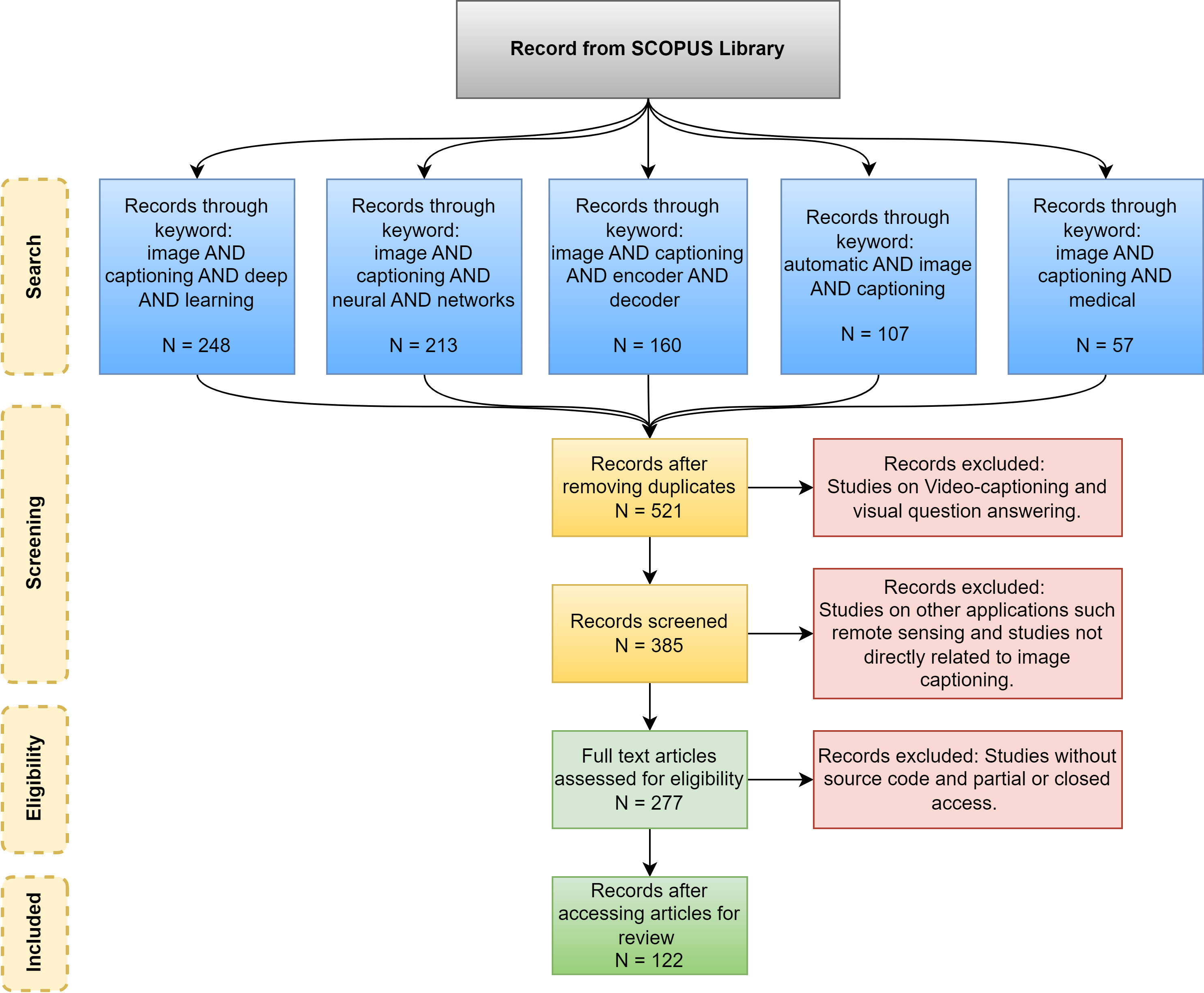}
        \caption{Literature search methodology}
        \label{fig:Search-methodolgy}
    \end{figure}
}

\section{Supervised Traditional approaches to image captioning}

    \subsection{Retrieval Based}
    {
        Retrieval-based methods in image captioning aim to locate images similar to a target image within a training dataset of image-caption pairs and use sentences associated with these similar images to generate captions for the target image. To obtain a candidate set of images from the existing database, various methodologies have been explored. One common approach involves utilizing the nearest neighbor rule to identify images most closely resembling the target image, followed by matching the corresponding sentences using a criterion such as the Tree-F1 rule\cite{10.1007/978-3-642-15561-1_2}. The Tree-F1 rule, which accounts for both accuracy and specificity, offers a balanced assessment by effectively distinguishing between positive and negative instances. Another strategy involves computing the global similarity between the target image and images in the database.\cite{NIPS2011_5dd9db5e}
        
        Another technique involves leveraging \textit{k}-nearest neighbor images from the training set to establish a set of candidate captions. These candidate captions are then utilized to generate new captions for the target image, with a focus on selectively combining relevant phrases.\cite{devlin2015exploringnearestneighborapproaches}
        
        Retrieval-based methods in image captioning, while offering a pragmatic approach to generating captions, are not without limitations. One notable drawback is their inability to produce novel descriptions and adapt effectively to unseen images. This limitation arises primarily from the over-reliance on annotated databases, resulting in generated sentences that closely mimic existing captions in terms of syntax and style. Consequently, the scope of generated captions remains confined to the existing pool of sentences and phrases, hindering the model's capacity to accommodate new objects or scenes present in unseen images. Moreover, due to the inherent constraints of relying solely on pre-existing annotations, there's a risk of generating captions that are entirely unrelated to the given image, thereby compromising the overall coherence and relevance of the output.
    }

    \subsection{Template Based}
    {
        Template-based image captioning involves the use of predefined syntactic structures or templates to generate captions for images. The main idea is to break down sentences into fragments such as subjects, predicates, objects, etc., and then map these fragments with the content of the target image. This process typically involves extracting single words from the image, such as subjects, predicates, and prepositions, and linking them together to form descriptions.\cite{yang-etal-2011-corpus,10.1007/978-3-642-15561-1_2} Various approaches have been explored within this framework.
        
        One approach utilizes a \emph{Support Vector Machine} (SVM) to identify three key components in the image—objects, actions, and scenes—and represent them as single semantic words to describe the image\cite{10.1007/978-3-642-15561-1_2}. Another strategy involves predicting a quadruplet of Nouns, Verbs, Scenes, and Prepositions via \emph{Hidden Markov Models} (HMM), which are then used to fill the predefined templates\cite{8964487}. These template-based methods offer the advantage of generating grammatically correct captions.
        
        However, template-based image captioning methods also have several drawbacks. Firstly, the templates are predefined, limiting the system's ability to generate captions of variable lengths or adapt to diverse image content effectively. Additionally, the descriptions generated through this method tend to be relatively simple, lacking in complexity and detail. Moreover, since the method relies on annotating numerous objects, attributes, and relationships within the image, it can become cumbersome to apply to large-scale datasets and may not be suitable for images across various domains.
    }

\section{Problem Design: Component Overview}
{
    Image captioning, a task at the junction of computer vision and natural language processing, involves automatically generating textual descriptions for images. To achieve this, a machine learning model is used to comprehend image content and articulate it in natural language, typically as a sentence or short paragraph.
    
    The process involves two main components: the Visual Feature Extractor and the Caption Generator. The Visual Feature Extractor employs CNNs to extract salient features from input images, generating vector representations. These representations, along with annotations from a paired dataset, are utilized to train the model.
    
    On the other hand, the Caption Generator utilizes the extracted features to generate descriptive captions. This component incorporates natural language processing techniques, such as RNNs with attention mechanisms, to craft coherent textual descriptions based on the image features and annotations.
    
    This architecture follows an encoder-decoder paradigm, where the encoder, typically a CNN, extracts features from images and annotations, producing fixed-dimensional vector representations. These representations serve as input to the decoder, often an RNN or LSTM network, responsible for generating captions conditioned on the encoder's output.

    \begin{figure}[H]
        \centering
        \includegraphics[width=0.9\linewidth]{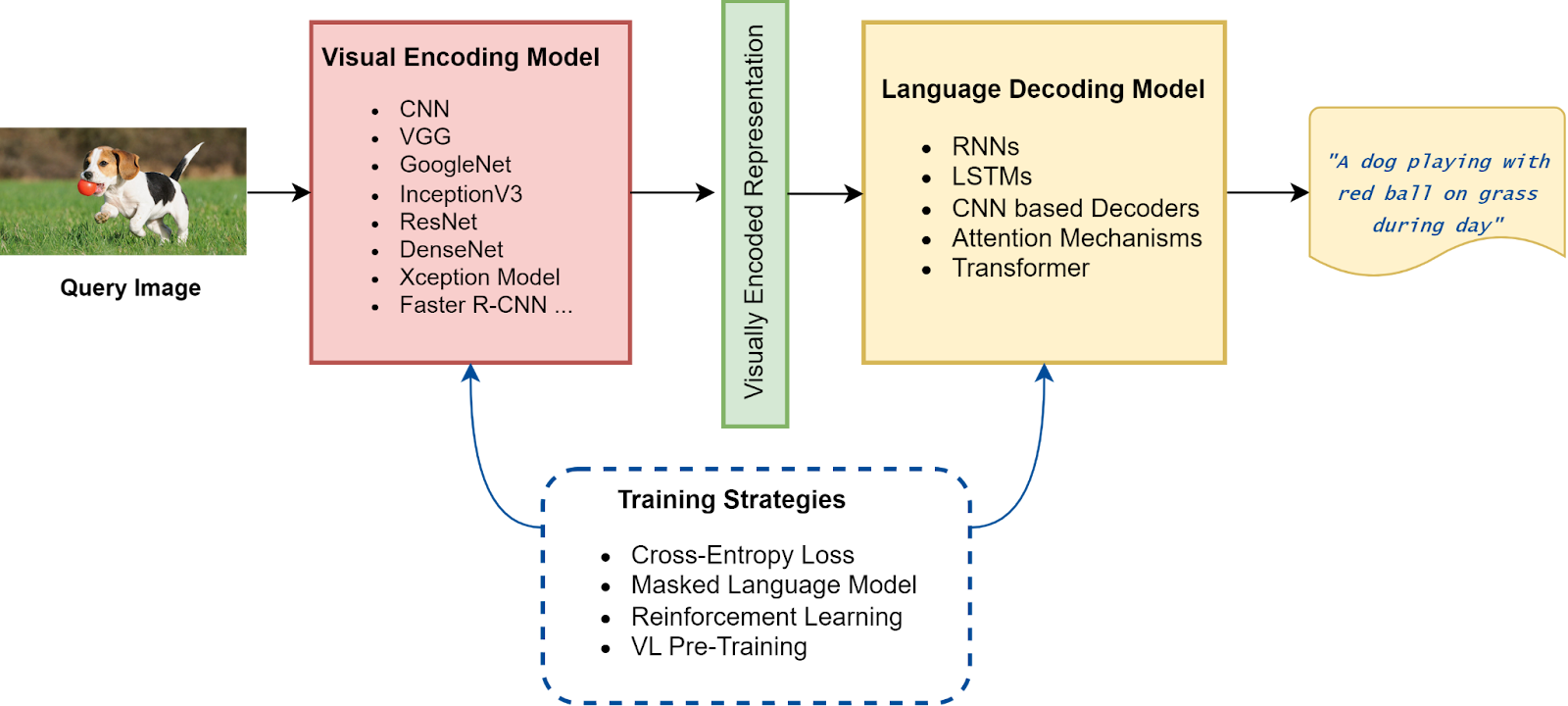}
        \caption{A generalized view of encoder-decoder architecture with sample input and output}
        \label{fig:encoder-decoder}
    \end{figure}

    Overall, image captioning is achieved through the seamless integration of computer vision and NLP techniques, facilitating the translation of visual content into textual descriptions. Here, we review each component of the encoder-decoder architecture separately, outlining the evolution in the CNN architecture as well as the RNN architecture, in the later section, we will review how different evolved architectures with various techniques such as attention and scene graphs were employed to generate richer and coherent captions.

}

    \subsection{Evolution of \emph{Encoder} Architecture in Context of Image Captioning}
    {
        In the image captioning task, the first objective is to extract feature representations from the input image and then these representations are fed to the decoder for caption generation, in between the encoder and the decoder we may have additional blocks such as attention blocks to improve the input-output alignment or capture semantics, intricate details or context within an input image.
        
        In order to generate high-level abstractions of input images, typically a CNN architecture is used. The strong learning capabilities of a CNN architecture, derived from multiple feature extraction stages make it a suitable model for image captioning tasks, owing to this capability a CNN can automatically generate data representations. A typical CNN architecture consists of a series of convolutional and pooling layers, with one or more fully connected layers at the end. One possible substitution for the final fully connected layer is a global average pooling layer. The CNN design incorporates many regulatory units, such as batch normalization and dropout, to optimize performance. The configuration of CNN components is crucial in the construction of novel and advanced systems.In order to enhance the capabilities of CNNs, researchers have investigated various other factors, including the utilization of different activation functions, loss functions, parameter optimization, and regularization techniques.\cite{Khan2020}
        
        \subsubsection{Component of CNN}
        {
            \textbf{Convolutional Layer} -
            The convolutional layer serves as the fundamental element of a CNN architecture. Each image has certain features and patterns such as shapes and textures, filters are employed to extract essential features that can subsequently be utilized to acquire semantic and spatial information. 
            Filters, also referred to as kernels or masks, are matrices that are learned and convolved with the input image. The process of convolution entails moving the filter across the input image and calculating the dot product between the filter and a small patch of the input data, followed by summing the results. 
            Multiple feature maps are generated by employing multiple filters in parallel to capture various features. Different filter sizes capture different levels of features from the input data, in general, large-size kernels capture coarse-grained features, and small-size filters are used to capture fine-grained features. 
            To extract a diverse set of features multiple levels of feature extraction are used. However, certain feature maps may be irrelevant or redundant in the process of distinguishing objects. Additionally, extensive feature sets can result in the presence of irrelevant information and may cause the network to overfit.\cite{Khan2020}
            
            \textbf{Pooling layer} - 
            Pooling layers are strategically positioned after one or more convolutional layers in a CNN to achieve down-sampling. This down-sampling reduces the spatial dimensions while retaining essential features. Pooling contributes to enhancing the network's robustness against minor perturbations in the input image, while also reducing computational overhead for faster training and inference.
            Various pooling techniques can be employed for down-sampling. In max pooling, the maximum value within the pooling window is selected, emphasizing critical features in that region while discarding others. Conversely, average pooling calculates and retains the average value of all elements within the pooling window, capturing general trends within the region.\cite{Khan2020}
            
            \textbf{Activation function} - 
            Neural networks are able to capture complicated relationships and nuanced dependencies within the data due to the non-linearity introduced by activation functions. The research literature uses a variety of activation functions, such as $sigmoid$, $tanh$, \emph{max out}, \emph{SWISH}, \emph{ReLU}, and its variations, \emph{leaky ReLU}, \emph{ELU}, and \emph{PReLU}, to introduce non-linear combinations of features. \emph{ReLU} and its variations are frequently preferred due to their effectiveness in addressing the vanishing gradient problem.\cite{Khan2020}
            
            \textbf{Batch normalization} - 
            Batch normalization addresses the problem of internal covariate shift while training deep neural networks. Internal covariate shift occurs due to changes in the distribution of each layer’s input during training. To mitigate this problem each layer’s inputs are normalized. Particularly in CNNs, batch normalization is applied prior to non-linearities to ensure stable distributions in activations. This practice facilitates the use of higher learning rates and contributes to model regularization.\cite{ioffe2015batchnormalizationacceleratingdeep}
            
            \textbf{Dropout} - 
            Dropout is a technique that resolves two main challenges related to deep neural networks. Over-fitting and combining different neural network architectures. In this technique, a unit along with all its incoming and outgoing connections are temporarily removed randomly.  After randomly dropping units from a network during training a ‘thinned’ network is obtained which contains all the units that were not dropped. In general, if a network has $n$ units, with dropout we have $2n$ possible thinned networks. During training, for each training example, we present to the network, we randomly choose one of these thinned networks and train it. This process repeats for each training example. So, in essence, training a neural network with dropout is like training a large collection of slightly different networks, but they all share weights and are trained infrequently. This approach helps prevent over-fitting and encourages the network to learn robust features.\cite{10.5555/2627435.2670313}
            
            \textbf{Fully connected layer} - 
            A fully connected layer performs a non-linear combination of various features for classification by aggregating inputs from previous layers throughout the network.\cite{Khan2020}
        }

        \subsubsection{Few high-end CNN models used in Visual Encoding}
        {
            \textbf{\underline{Visual Geometry Group}}

            By concentrating on spatial filters, the \emph{Visual Geometry Group} (VGG) network brought about a novel architectural improvement in CNNs. It was found that various filter sizes capture varying degrees of granularity. In comparison to large-sized filters, smaller filters extract more fine-grained information, which enhances feature extraction performance. VGG can capture multiple levels of information in an image by using different sized filters. Notably, significant performance gains were observed when larger filters were replaced with stacked $3 \times 3$ filters. Another advantage of using small-size filters was that they reduced computational complexity by using fewer parameters. VGG uses $1 \times 1$ convolutions between convolutional layers to improve feature-map learning and control network complexity. VGG optimizes the tuning process for image analysis by placing max-pooling layers after the convolutional layer and using padding to preserve spatial resolution. But one significant disadvantage of VGG is that it has a large number of parameters (138 millions), which makes it difficult to implement on systems with limited resources.\cite{Khan2020}
            
            \textbf{\underline{GoogleNet}}

            GoogleNet, devised by Google researchers, aimed to achieve high accuracy while minimizing computational costs. It uses the inception block, which enhances CNNs by integrating multi-scale convolutional transformations through the split, transform, and merge paradigm. 
            This approach involved encapsulating filters of varying sizes ($1 \times 1$, $3 \times 3$, and $5 \times 5$) within compact blocks to capture spatial information across different scales.This helped in addressing a key challenge of learning diverse image variations within the same category having different resolutions.
            
            To streamline computations, GoogleNet introduced bottleneck layers of $1 \times 1$ convolutional filters, before employing large kernels and incorporating sparse connections to mitigate redundant information and omit irrelevant feature maps.
            
            Moreover, by leveraging global average pooling instead of fully connected layers and optimizing parameter tuning, GoogleNet significantly reduced the model's parameter count from 138 million to 4 million, enhancing computational efficiency without sacrificing performance. 
            
            However, GoogleNet's design also posed challenges, notably a representation bottleneck where the dimensionality of the feature space is significantly reduced in the subsequent layer. This reduction in feature space can sometimes lead to the loss of valuable information.\cite{Khan2020}
            
            \textbf{\underline{Inception-V3}} 
            
            One notable enhancement in \emph{Inception-V3} is the replacement of large-sized filters ($5 \times 5$ and $7 \times 7$) with smaller and asymmetric filters ($1 \times 7$ and $1 \times 5$). This strategic adjustment enables the network to capture spatial information more efficiently while minimizing computational costs. Additionally, Inception-V3 incorporates bottleneck layers of $1 \times 1$ convolutions before applying larger filters, effectively regulating computations and enhancing feature extraction. Another key feature of Inception-V3 is the utilization of $1 \times 1$ convolutional operations, which map the input data into smaller-dimensional spaces before applying regular ($3 \times 3$ or $5 \times 5$) convolutions. \cite{Khan2020}
            
            \textbf{\underline{ResNet}}
            
            \emph{Residual Network} (ResNet), introduces a fundamental shift in the architecture of deep neural networks by focusing on learning residual functions instead of directly approximating the underlying mapping from inputs to outputs. Traditional networks aim to learn the entire mapping function from scratch, which can become increasingly challenging as the network depth increases. However, ResNet addresses this challenge by reformulating the learning task to focus on approximating the residual mapping—the difference between the desired output and the input.
            In ResNet, each layer aims to learn a residual function that captures the difference between the input and the desired output. Instead of expecting each layer to directly fit the desired mapping, ResNet allows the layers to fit a residual mapping that, when added to the input, approximates the desired output. By stacking these residual blocks on top of each other, ResNet forms a deep network capable of learning complex mappings while mitigating issues like the vanishing gradient problem.
            
            Formally, ResNet recasts the original mapping function $H(x)$ as $F(x)+x$, where $F(x)$ represents the residual mapping to be learned by the stacked nonlinear layers. This reformulation allows the network to focus on learning the deviations from the identity mapping, which can be more straightforward to optimize compared to learning entirely new mappings.
            Empirical evidence suggests that ResNet architectures are easier to optimize, particularly as the depth of the network increases. Additionally, ResNet models can achieve higher accuracy by leveraging increased depth, as the residual learning approach facilitates the training of deeper networks without encountering performance degradation.\cite{Khan2020}
            
            \textbf{\underline{DenseNet}}
            
            DenseNet stands out among convolutional neural network architectures for its innovative approach to feature connectivity and parameter sharing. 
            
            In contrast to ResNet, DenseNet uses a dense connection pattern in which each layer concatenates feature maps along the channel dimension after receiving feature maps from all previous layers.
            This design choice promotes extensive feature reuse and facilitates enhanced information flow throughout the network, allowing each layer to directly access and build upon the features learned by all preceding layers. \cite{8099726}
            
            By concatenating features instead of simply adding them, DenseNet explicitly differentiates between new information added to the network and existing information preserved from earlier layers, potentially improving the network's ability to learn complex patterns and representations.
            Furthermore, DenseNet's narrow layer structure reduces the number of parameters in each layer, mitigating issues like the vanishing gradient problem and enhancing computational efficiency. 
            Additionally, DenseNet's direct access to gradients for each layer through the loss function serves as a regularizing mechanism, helping to combat over-fitting, especially in scenarios with limited training data. \cite{Khan2020}

            \textbf{\underline{Xception}}
            
            Xception represents a significant advancement in convolutional neural network architecture, building upon the principles of depthwise separable convolution and linear residual connections. Derived from the concept of Inception, Xception takes this idea to an extreme by employing depthwise separable convolution, which splits the convolutional process into two distinct steps. Firstly, a $1 \times 1$ convolution is applied to capture cross-channel correlations, followed by a separate operation to capture spatial correlations within each output channel. The Xception architecture comprises $36$ convolutional layers, forming a powerful feature extraction base for the network. By leveraging depthwise separable convolution and linear residual connections, Xception achieves state-of-the-art performance in a variety of computer vision applications while maintaining computational efficiency.\cite{8099678}
            
            \vspace{10pt}

            \textbf{\underline{Faster R-CNN}} 
            
            Faster R-CNN is endowed with region proposal modules that tells Fast R-CNN\cite{7410526}where to ‘look’. It accepts any size image as input and produces rectangular object proposals with an objectness score assigned to each. These proposals represent candidate regions in the image where objects might be present.
            
            The process typically involves sliding a small window (called an \emph{anchor}) over the image at various locations and scales. For each anchor, the \emph{Region Proposal Network} (RPN) predicts two scores:
            
            \begin{itemize}[itemsep=3pt, topsep=0pt, partopsep=0pt, parsep=0pt]
                \item {\textbf{Objectness Score}: This score indicates the likelihood of an object being present within the region defined by the anchor.}
                \item {\textbf{Bounding Box Regression}: This predicts adjustments to the anchor to better fit the object if it exists within the region.}
            \end{itemize}
            
            The objectness score helps filter out regions that are unlikely to contain objects, while the bounding box regression helps refine the proposals to more accurately fit the objects present in the image.
            
            Once the RPN generates these proposals, they are passed on to subsequent stages of the object detection pipeline for further refinement and classification. This process significantly reduces the computational burden compared to exhaustively examining all possible regions in the image.The RPN  can be trained end-to-end using backpropagation and stochastic gradient descent.\cite{ren2016fasterrcnnrealtimeobject}
            
            \begin{figure}[H]
                \centering
                \includegraphics[width=0.9\linewidth]{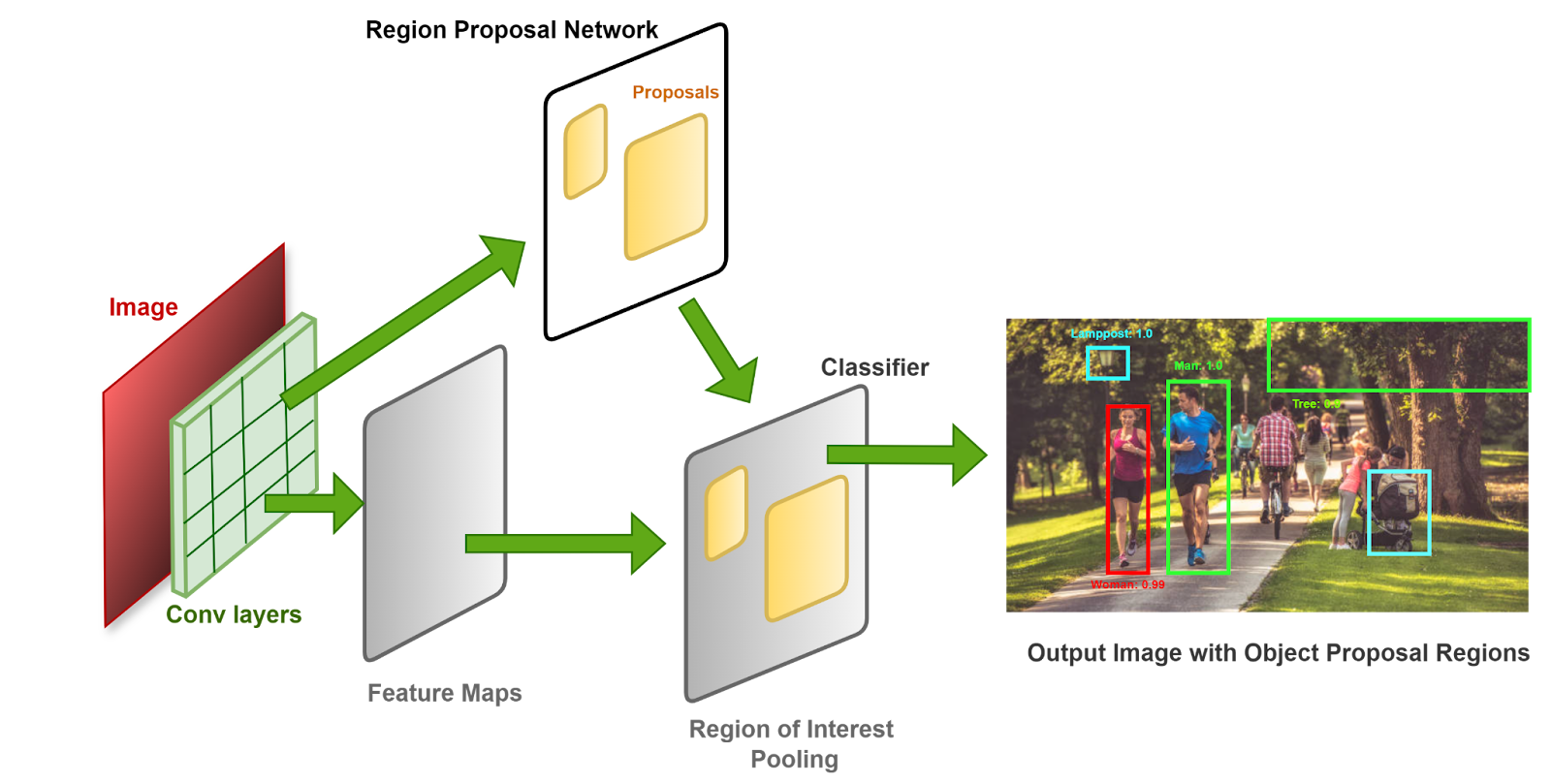}
                \caption{Generalized view of Region Proposal Network}
                \label{fig:region-propose-network}
            \end{figure}
        }
    }

    \subsection{Evolution of Decoder Architecture in the context of Image Captioning}
    {
        The decoder model takes the output of the encoder and generates captions, in the earlier image captioning models RNNs were employed, but RNNs had certain drawbacks such as vanishing/exploding gradient problems and difficulty in maintaining long-range dependencies, to resolve this issue LSTMs and GRUs were used which used memory cells to capture long-range dependencies, LSTMs enhanced the caption generation process by capturing long-range dependencies, however, their recurrent architecture is complex and significant storage requirement while training using backpropagation through time, poses certain challenges. To mitigate these issues CNN based decoders were employed which showed comparable results with LSTMs on various standard metrics. Although the issue of maintaining long-range dependencies was mitigated, the decoders still lacked the ability to generate captions which were aligned with specific features of the image. To enhance the alignment of the input and the output attention mechanisms were being employed in the encoder-decoder architecture. Later on, with the advent of transformers, which were more computationally efficient, encoder-decoder architectures with transformer layers were introduced. Here we provide a comprehensive overview of various architectures employed as decoders for image captioning.

        \subsubsection{Explaining RNNs}
        {
            In the realm of processing sequential data, such as text or time series data, feedforward networks fall short due to their inability to capture temporal dependencies inherent in such data. These networks treat each input independently, disregarding the sequential nature of the data. Moreover, feedforward models necessitate fixed-size input vectors, posing challenges for variable-length sequential data and potentially leading to information loss. To effectively model sequential data while preserving context and capturing long-range dependencies,RNNs and their variants, like LSTMs and GRUs, are commonly employed. In a vanilla RNN, inputs at each time step are multiplied by weight matrices and passed through non-linear activation functions, such as hyperbolic tangent, to compute values for a layer of hidden units; these hidden layers are then used to calculate corresponding outputs. The hyperbolic tangent activation function is often chosen for its desirable properties, such as monotonicity, negative symmetry, and boundedness, making it suitable for encoding complex patterns and preserving gradient flow during training. The distinctive aspect of RNNs lies in their ability to maintain internal state, or memory, across time steps. This memory encodes prior processing information and influences decisions at subsequent time steps, facilitating the capture of context and temporal dependencies. \emph{Backpropagation Through Time} (BPTT) is the mechanism through which RNNs are trained. It extends backpropagation to sequential data by unrolling the network over time and propagating errors through each time step, enabling the network to learn from past interactions and adjust its parameters accordingly.

            \begin{figure}[H]
                \centering
                \includegraphics[width=0.7\linewidth]{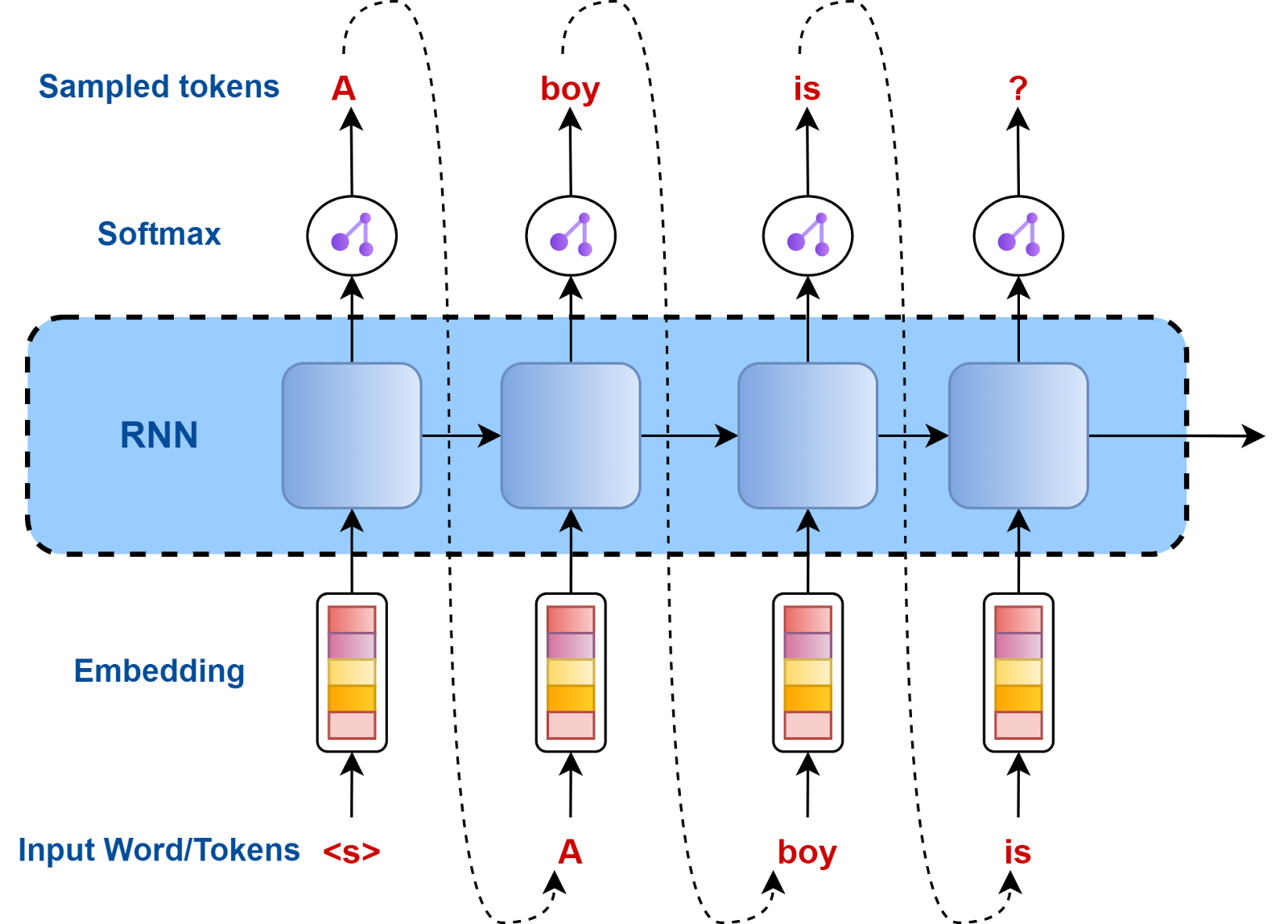}
                \caption{Simplified RNN architecture for token generation}
                \label{fig:RNN}
            \end{figure}
        }
        \subsubsection{Advent of LSTMs}
        {
            RNNs faced challenges in training due to the exploding or vanishing gradient problem, hindering their ability to effectively capture long-term dependencies in sequential data. LSTM networks were developed as a robust solution to mitigate this issue. LSTMs mitigate the vanishing/exploding gradient problem by incorporating specialized units composed of a cell, input gate, output gate, and forget gate.These components collectively enable the LSTM to regulate the information flow inside the network. At the core of the LSTM architecture are memory blocks, a set of recurrently connected sub-networks, which maintain state information over time intervals.The cell within each LSTM unit retains values over arbitrary time intervals, while the input gate controls the flow of information into the cell, the output gate regulates the output flow, and the forget gate manages the retention or discard of information stored in the cell. This design allows LSTMs to effectively capture long-term dependencies while preventing the issues associated with vanishing or exploding gradients.
            
            BPTT, a key training algorithm for LSTMs, requires storing intermediate activations and gradients over sequential time steps. This can result in significant memory consumption, particularly for long sequences, which may limit the scalability of LSTM models and increase training time and resource requirements.\cite{vanHoudt2020}

            \begin{figure}
                \centering
                \includegraphics[width=0.5\linewidth]{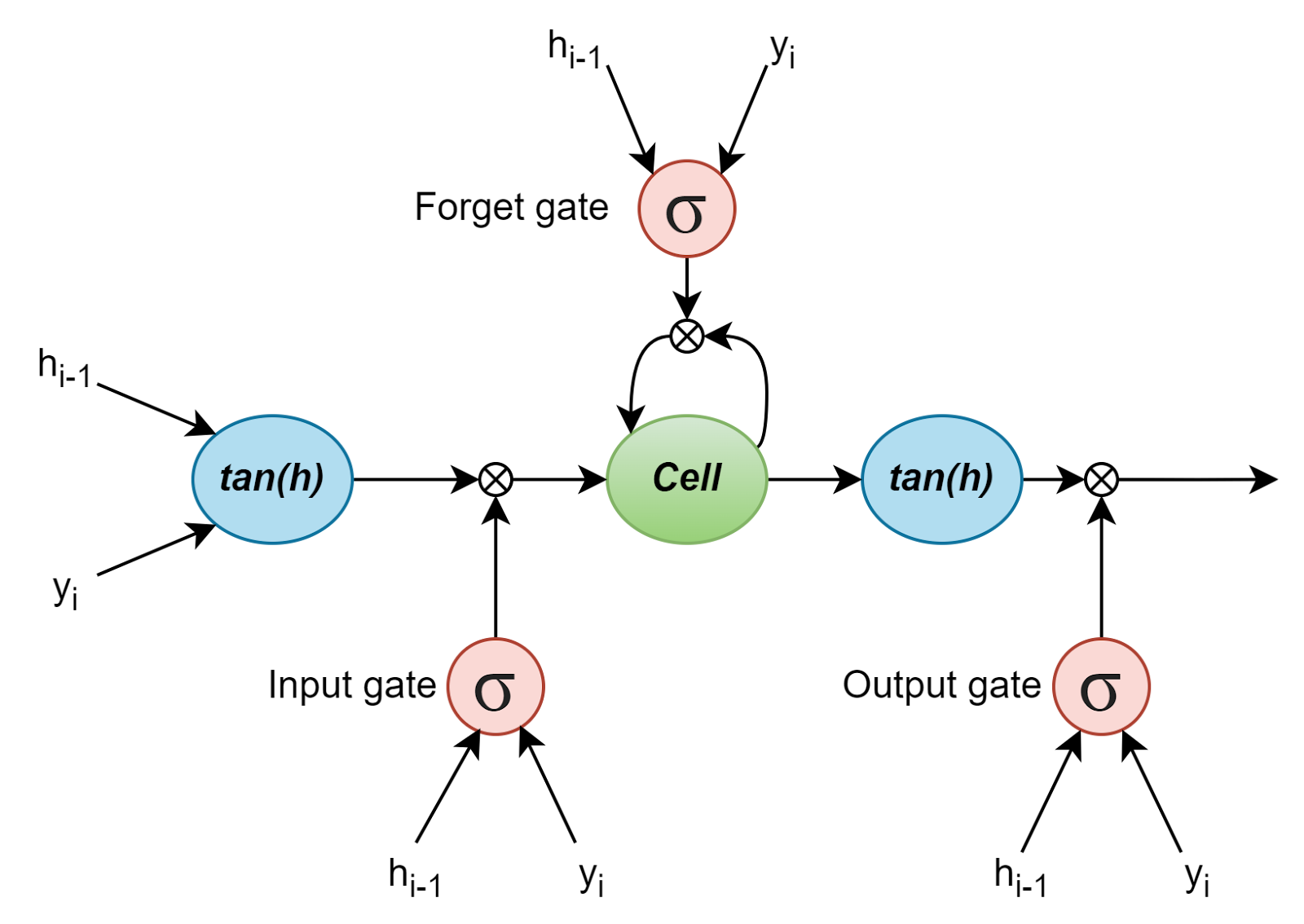}
                \caption{A basic building block of LSTM}
                \label{fig:LSTMs}
            \end{figure}

            
        }
        \subsubsection{CNN based decoders}
        {
            In \cite{aneja2017convolutionalimagecaptioning}, the authors introduced a CNN-based decoder for image captioning, demonstrating comparable performance to LSTM-based methods on standard evaluation metrics such as BLEU\cite{10.3115/1073083.1073135}, CIDEr\cite{vedantam2015ciderconsensusbasedimagedescription}, SPICE\cite{anderson2016spice}, and ROGUE\cite{lin2004rouge}. Instead of employing LSTM or GRU units, this approach utilizes masked convolutions, which are feed-forward networks. To predict a word based on previous words or their representations, they used a feed-forward deep neural network, as opposed to the RNN formulation, where the probabilistic model unfolds over time. This allows for sequential inference, with one word processed at a time. Masked convolutions are used that operate solely on past data, ensuring that future word tokens are not incorporated into the processing. Since all ground truth words are available at every time-step and there are no recurrent connections, this model can be trained in parallel for every word. This architecture provides a viable substitute for conventional recurrent-based decoders in image captioning applications by enabling effective training and inference.
            
            \begin{figure}[H]
                \centering
                \includegraphics[width=0.6\linewidth]{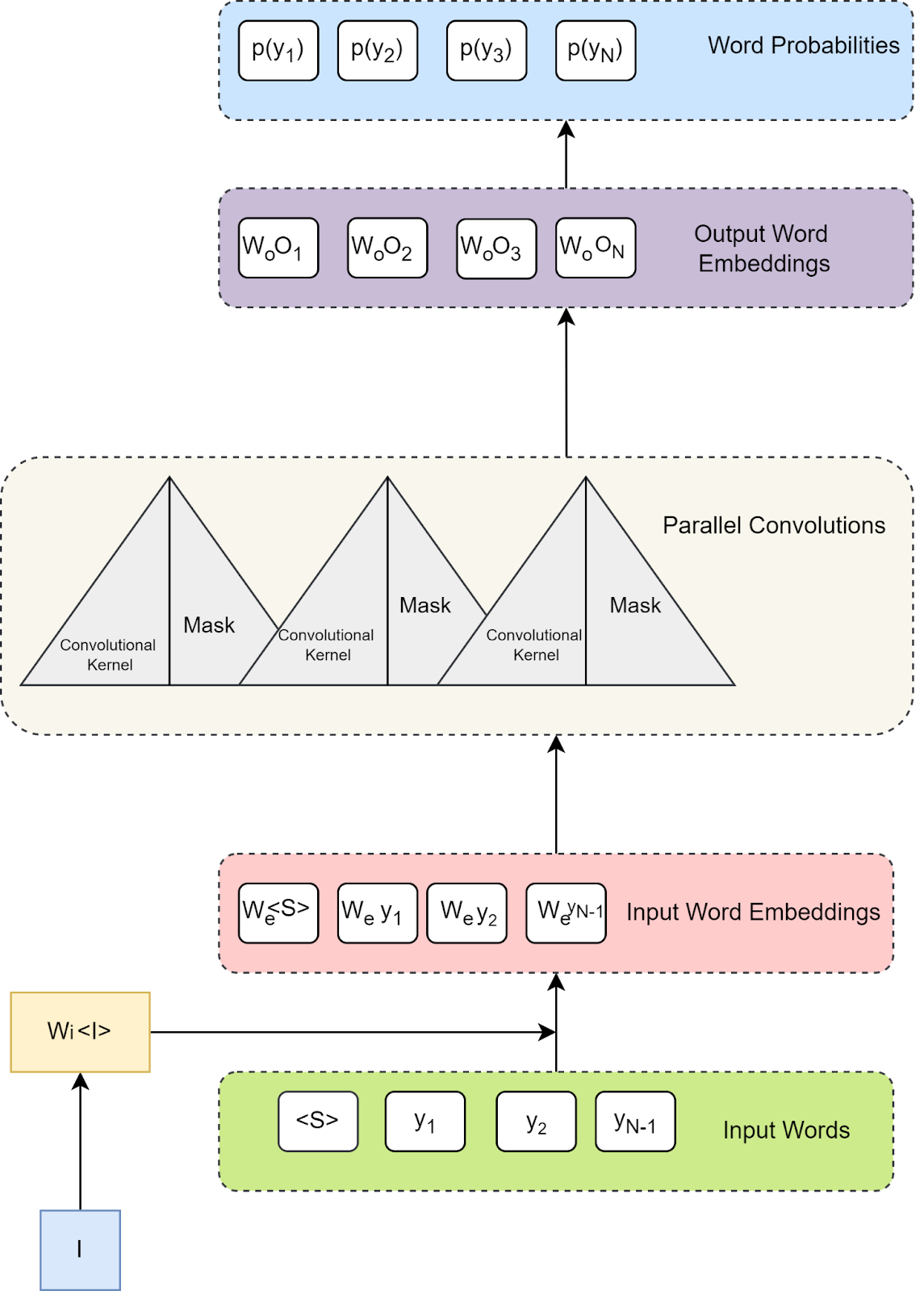}
                \caption{feed-forward with masked convolution \cite{aneja2017convolutionalimagecaptioning}}
                \label{fig:feed-forward}
            \end{figure}
        }
        \subsubsection{Advent of Attention Mechanisms}
        {
            In image captioning tasks using a typical encoder-decoder architecture, the key challenge lies in aligning the input (images) and output (captions). It is expected that certain characteristics of the image will have an impact on the generated captions. When producing the individual words of the caption, the decoder does not, however, have a mechanism for spatially attending to salient portions of the image.

            The attention mechanism addresses the alignment issue between the output and input sequences by introducing attention weights across the encoded input sequence. These weights highlight the components of the input sequence containing essential information to produce the next token in the output sequence.
            Specifically, an attention block is inserted between the encoder and decoder modules. This block autonomously learns attention weights, capturing the relevance between the encoder's hidden states and the decoder's hidden state.\cite{10.1145/3465055} 
            

            
        }
        
        \subsubsection{Explanation of  Attention Model}
        {
            Two basic phases are involved in the implementation of the attention mechanism for image captioning: computing the attention distribution on the input data and computing the context vector based on the attention distribution. The neural network first calculates the input data's attention distribution, where the source data feature is encoded as \emph{keys} $(K)$. These keys represent various aspects of the input, such as specific areas of an image. Additionally, a task-specific representation vector $q$, termed \textit{the query}, is introduced. Score function $f$ is used to obtain the energy score $e$, which assesses the correlation  between keys and queries .This function, crucial to the attention model, determines the importance of queries in relation to keys for generating the next output. \cite{NIU202148}
            
            After that, an attention distribution function $g$ maps the energy scores $e$ to attention weights, normalizing the scores into a probability distribution. In the computation of context vectors, a new data feature representation $V$, referred to as values, is introduced. Each element of $V$ corresponds to a specific element of $K$. Upon computing attention weights and values, a single context vector $c$ is determined. This context vector is primarily influenced by higher attention weights linked with the corresponding values, facilitating effective attention and context selection in image captioning tasks.\cite{NIU202148}
            
            \begin{figure}[H]
                \centering
                \includegraphics[width=0.8\linewidth]{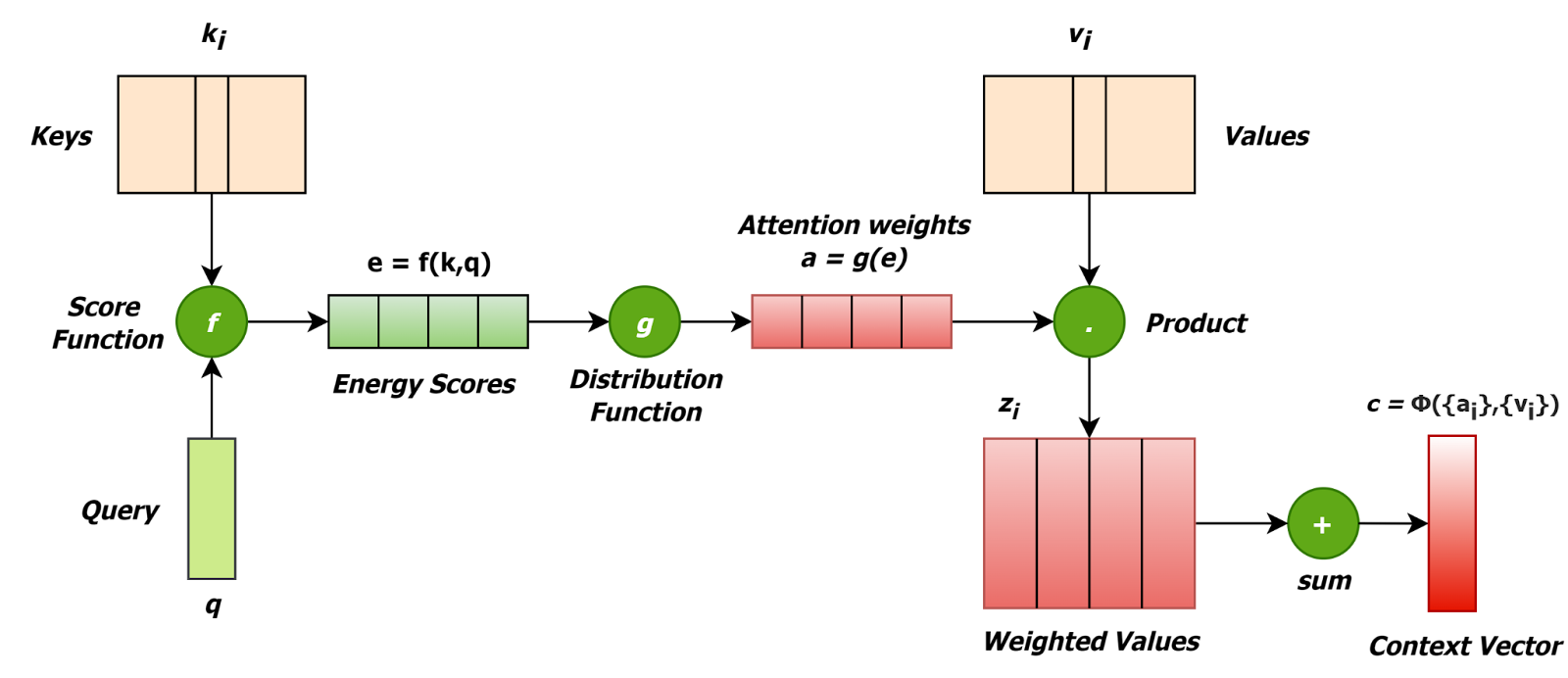}
                \caption{Unified Attention Model’s generic Architecture}
                \label{fig:unified}
            \end{figure}
        }
        
        \subsubsection{Transformers}
        {
            The transformer design uses an attention method to identify global dependencies between input and output, hence eliminating the need for RNNs and CNNs.\cite{vaswani2023attentionneed}

            Comprising two essential components—the position-wise \emph{Feed-Forward Network} (FFN) layer and the \emph{Multi-head Attention layer}; the Transformer achieves its transformative capabilities through an encoder-decoder structure.
            
            Within the Transformer, the encoder consists of stacked self-attention and feed-forward layers. This structure allows the model to capture intricate relationships inside the input sequence. The decoder employs self-attention for individual words and cross-attention over the encoder's output, ensuring comprehensive understanding and effective generation of output sequences.
            
            The position-wise FFN layer functions as a fully connected feedforward network, treating each position within the sequence uniformly. This approach maintains the positional information of each symbol throughout the operation, enhancing the model's ability to obtain context.
            
            Multi-head attention, an important aspect of the Transformer, enables focused information extraction from diverse representation subspaces across various positions. By stacking multiple self-attention layers, akin to multiple channels in a CNN, the model gains the ability to capture patterns and dependencies within the input sequence. Moreover, the self-attention mechanism's complexity remains constant, even when establishing long-range dependencies, contributing to the Transformer's parallelizability and computational efficiency.\cite{NIU202148}
            
            \begin{figure}[H]
                \centering
                \includegraphics[width=0.6\linewidth]{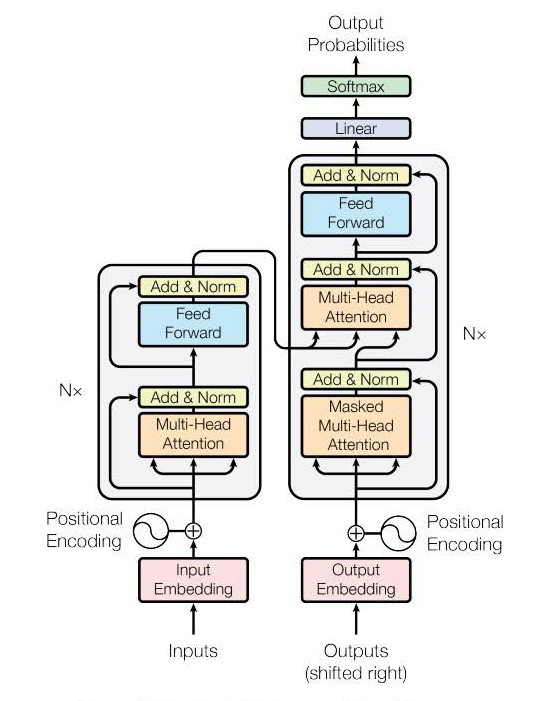}
                \caption{The Transformer Architecture \cite{vaswani2023attentionneed}}
                \label{fig:transformer}
            \end{figure}
        }
        
    }


\section{Coming together: Encoder-Decoder Architecture}
{
    The task of image captioning involves transforming visual inputs into textual outputs, thus fitting into the encoder-decoder architecture paradigm. Here, the encoder, typically a CNN-based model, extracts vector representations from images, while the decoder, an NLP-based architecture, generates sequential text outputs. To enhance caption quality, numerous modifications have been introduced. These include incorporating attention mechanisms within both the encoder and decoder to capture semantic and syntactic nuances. Additionally, some models utilize techniques like scene graphs to enrich the encoder's understanding of the visual content. Furthermore, to provide more fine-grained captions, dense-localization layers are employed after the convolutional layer to predict regions of interest within images. 
    
    In certain models, the vector representations from images and annotations are mapped into a multi-modal space, facilitating a holistic representation of the visual and textual information. All of these methods work together to enhance the resulting captions' richness and relevancy, adapting the encoder-decoder framework to better handle the complexities of image-caption generation.

    \subsection{Encoder-Decoder with Global Image Features}
    {
        In order to effectively represent visual content, various approaches to visual encoding have been employed. These approaches include using global CNN features, regional features, scene graphs, and attention-based mechanisms. Global CNN features, extensively used in image encoders, offer several advantages. Their simplicity and compactness allow them to capture and summarize information from the entire input, providing an overall context of an image. 
        
        To acquire abstract representations of an input image, it's common practice to leverage the activations of a CNN's later layers. These representations act as valuable contextual hints for the language model. In the \emph{Show and Tell} \cite{vinyals2015tellneuralimagecaption} approach, global CNN characteristics are derived from GoogleNet and subsequently inputted into the language model's initial state. In \cite{karpathy2015deepvisualsemanticalignmentsgenerating} the authors utilized global features from AlexNet, whereas \cite{mao2014deep}  and \cite{donahue2016longtermrecurrentconvolutionalnetworks} employed global features from VGG, integrating them at every time-step of the language model.
        
         The NIC model, as discussed in \cite{Vinyals_2017}, leverages GoogleNet's last fully connected layer to activate and extract fixed-size, high-level representations of visual features. However, alternative strategies have been proposed by different researchers. For instance, the authors in\cite{10.5555/3045118.3045336} diverge from using the last fully connected layer output and instead utilize the output of underlying convolutional layers as the image feature vector. Additionally, in \cite{donahue2016longtermrecurrentconvolutionalnetworks}, a \emph{Long-Term Recurrent Convolutional Networks} (LRCN) module is introduced after the initial two fully connected layers to accommodate varying-length visual input, enhancing the model's adaptability. Building upon this, authors in \cite{wu2016valueexplicithighlevel} and \cite{9578290} adopt pre-trained CNNs to obtain global grid features while simultaneously learning high-level attribute representations, thereby improving caption accuracy. Furthermore, in \cite{mao2016generationcomprehensionunambiguousobject}, a multi-modal layer is integrated following the feature layer to fuse visual and textual representations, reinforcing their association and contributing to overall model performance.
        
        Despite these advancements, a noted drawback is that the global grid representations, although comprehensive, uniformly divide the content of an image. Consequently, this uniform fragmentation may treat relevant and non-relevant objects and regions alike, posing challenges in generating specific and accurate descriptions and leading to excessive compression of information.\cite{9849164}
    }

    \subsection{Encoder-Decoder with Regional Image Features}
    {
        In order to obtain finer understanding of the visual features, fine-grained analysis and multi-step reasoning processes are employed on the input images. This entails encoding visual features of prominent regions within images. By focusing on significant regions within an image researchers are able to delve deeper into the visual content, meaningful insights such as spatial semantics extracted from the regional image features help in generating more detailed and coherent captions.

        By leveraging multi-modal embedding techniques, researchers in \cite{karpathy2015deepvisualsemanticalignmentsgenerating} align sentence snippets with specific visual regions, thus enhancing the integration of language and image understanding. This focused analysis on particular visual areas facilitates a more nuanced comprehension of the image content. Utilizing the Faster R-CNN\cite{ren2016fasterrcnnrealtimeobject} detector as a foundation, an increasing number of methodologies are being developed to capture salient visual regions and extract high-level semantic information. Notable examples include\cite{anderson2018bottomuptopdownattentionimage} and \cite{NEURIPS2018_d2ed45a5}. Building upon this, authors in \cite{datta2019align2groundweaklysupervisedphrase} propose a method that utilizes \emph{Regions of Interest} (ROIs) and phrases within the caption to establish latent correspondences between them, thereby generating a discriminative image representation based on these matched ROIs. Additionally, in\cite{9573322}, \emph{Region Proposal Network} (RPN) is employed to obtain object proposals, followed by a combination layer that pairs these proposals and assigns them to subject and object regions to extract contextual information. Moreover, the method presented in \cite{Yang_Yang_Hsu_2021} integrates the spatial coherence of objects into a caption-generating model by combining raw visual features of overlapping object pairs to create directional pair features. These features are then optimized using learned weights, resulting in relation-aware object features that enhance the language-generating model's performance.
        
        These methods for regional feature encoding effectively address the challenge of learning fine-grained information within visual features. Beyond capturing visual representations based on global grids or salient regions, these convolutional techniques aim to identify semantic features within various regions of an image and encode them into higher-level abstractions.
        
        However, a limitation observed across many approaches is, they focus solely on entity regions, overlooking interactions between different regions. Consequently, these methods lack the ability to capture topological structure and paired relationships.\cite{9849164}
        
        To enhance the granularity of visual encoding, attention mechanisms are employed, a topic which will be explored extensively in section 5.5.
    }

    \subsection{Encoder-Decoder with Scene Graphs}
    {
        \begin{figure}[H]
            \centering
            \includegraphics[width=0.7\linewidth]{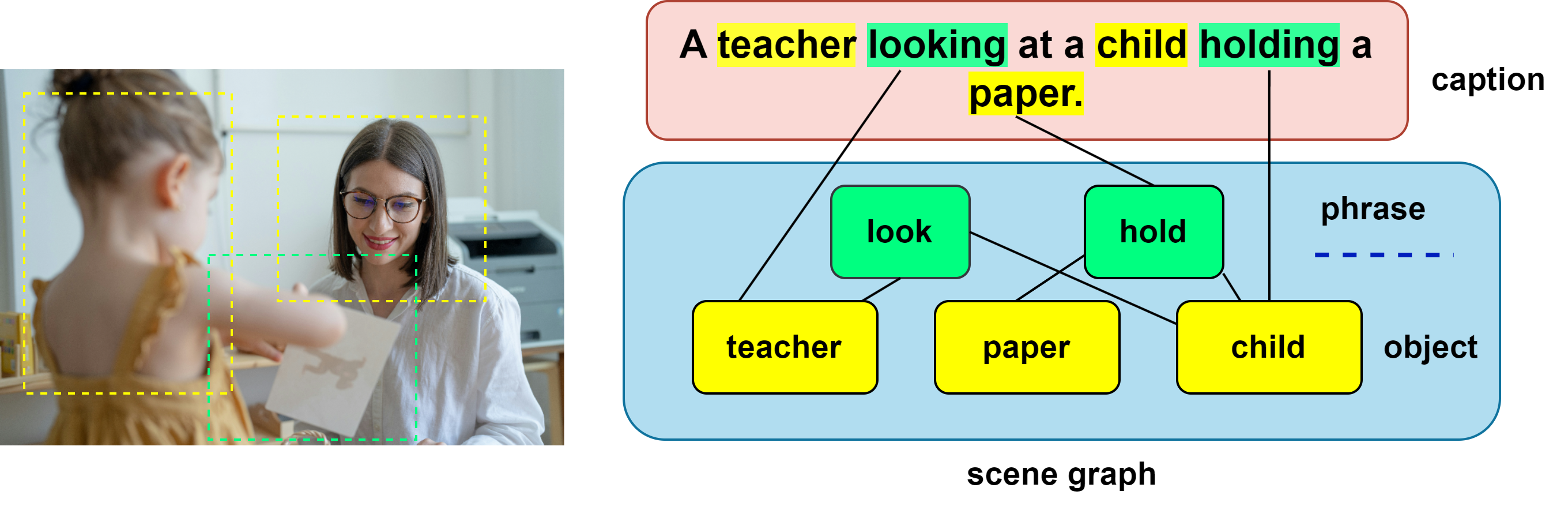}
            \caption{A general scene graph generation process\cite{li2017scenegraphgenerationobjects}}
            \label{fig:scene-graph}
        \end{figure}

        Scene graphs, operating as directed graph data structures, play a crucial role in capturing the visual semantics of images by abstracting objects and their relationships. This abstraction facilitates the clear representation of items, attributes, and the interactions between them within a scene, aiding in the comprehension of image content.\cite{9661322}
        
        A \emph{Scene Graph} has several components, an object instance, attributes and relations. The object instances can be a person, place, thing or a part of another object, the attributes are employed to define the current object’s state which could be shape, color and pose. Relations are used to characterize the relationships, such as actions or placement, between pairs of objects.
        
        A scene graph can be described in the form of a tuple, \textit{(O,R,E)}, where \textit{O} is the set of objects detected  in the image, each object instance can be denoted as a pair of category and attribute. \textit{R} is the set of relations between object instance nodes and \textit{E} represents edges between the object instance nodes and relationship nodes.
        If an object instance is categorized as background or if there is a relationship between two object instances that is deemed irrelevant, an edge is automatically eliminated.
        As a result, the instances of objects localized within an input image by bounding boxes, as well as the interactions between each pair of object instances, make up the scene graph.This clear representation of objects and their relationships forms the foundation for various computational approaches aimed at understanding image content.
        
        Building upon this concept, the \emph{Tensor Product Scene-Graph-Triplet Representation} (TPsgtR), proposed by\cite{Sur2019TPsgtRNT}, introduces a novel technique for caption generation by employing neural-symbolic encoding of scene graphs derived from regional visual information in images. Rather than depending on the model to construct every potential combination,
        neuro-symbolic embeddings of scene-graphs facilitate embedding the identified relationships among various regions of the image into tangible representations.
        Similarly in \cite{chen2020saywishfinegrainedcontrol} the authors introduced the \emph{Abstract Scene Graph} (ASG) structure to address the gap in image captioning models, which often are unable to generate diverse descriptions according to user intentions. The ASG structure regulates the amount of detail in generated descriptions and allows for the fine-grained depiction of user intentions. It consists of abstract nodes representing objects, attributes, and relationships, based on the image without any particular semantic labeling, making it easily obtainable either manually or automatically.
        
        Moreover, in \cite{XU2019477} the authors proposed scene graph captioner which tackles the challenges of using scene graphs for image captioning. The main challenge is to construct scene graphs that capture the rich annotations of objects, attributes, and relationships within one image. The extremely intricate relationships between items and their localizations—which go beyond straightforward pairwise relations—present another difficulty in integrating scene graphs for image captioning. The authors suggest a scene graph captioner that produces logical descriptions by taking into account object interactions and localizations.
        
        They begin by creating an image feature representation for the input image by running it through a CNN. The scene graph is then created by using a number of modules to identify the objects, relationships, and attributes in the input.Based on the scene graph of an image, the authors create the attention region for the image using an attention extraction model. In order to discover tightly clustered regions, it uses a dominant clustering strategy to generalize clusters for the input graph's vertex set. The attention bounding box is defined by the attention graph that is produced, emphasizing important regions of the image. 
        
        To overcome the challenges caused by the lack of paired picture-caption datasets, the researchers in \cite{gu2019unpairedimagecaptioningscene} introduce a scene graph-based method for unpaired image captioning. The framework includes necessary parts like an image scene graph generator, a sentence scene graph generator, a scene graph encoder, and a sentence decoder. The model is trained first on the text modality, and then it uses an unsupervised feature alignment technique to align scene graphs between sentences and images. This is a noteworthy aspect of the training procedure. 
        
        Finally, in \cite{yang2018autoencodingscenegraphsimage} the authors incorporate human-like biases into image captioning through a scene graph auto-encoder. They generate scene graphs for both image and sentences.In the textual domain, a Scene Graph Auto-Encoder \textit{(SGAE)} is employed to learn a dictionary that aids in reconstructing sentences. This dictionary encapsulates the desired language biases or patterns.  In the vision-language domain, this shared dictionary guides the encoder-decoder model in generating captions from images. By leveraging the scene graph representation and the shared dictionary, the inherent biases present in language are effectively transferred across both visual and textual domains, facilitating more human-like and contextually rich captions.
    }

    \subsection{Encoder-Decoder with Multi-Modal Space}
    {
        \begin{figure}[H]
        \centering
            \includegraphics[width=0.9\linewidth]{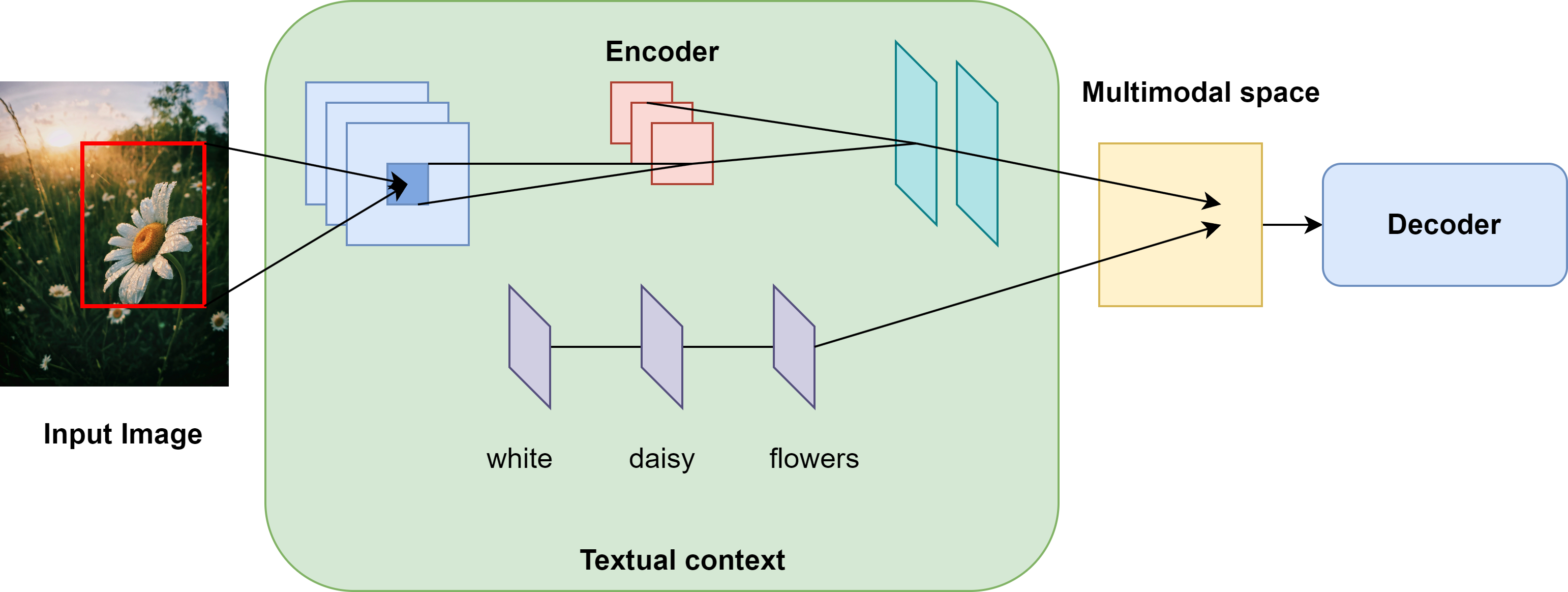}
            \caption{A typical encoder-decoder with multimodal space}
            \label{fig:encoder-decoder with multimodal space}
        \end{figure}
    
        Learning joint feature spaces for images and their captions by projecting  image features and sentence features into a common space, is useful in image search and ranking image captions. 
        
        A language encoder, a vision module, a multi-modal space component, and a language decoder are the main architectural elements of a typical multi-modal space based system\cite{10.1145/3295748}. The vision module uses a deep convolutional neural network to identify visual properties in images by extracting features from them. The language encoder, meantime, is occupied with obtaining word characteristics and generating dense embeddings for every word. These embeddings are then fed into recurrent layers, allowing it to gradually capture semantic context.
        Subsequently, the multi-modal space component plays a crucial role in aligning the extracted image features with the word features and mapping them into a common space. This alignment facilitates the integration of visual and textual information. Finally, the language decoder utilizes this aligned representation to generate captions\cite{10.1145/3295748}.
        
        Expanding on this concept, researchers in\cite{karpathy2014deep} propose a multi-modal image captioning model which retrieves relevant images given a sentence query, and vice-versa. They demonstrated that considering both the global content of images and sentences, as well as the specific details within them, can greatly enhance performance of  image-sentence retrieval tasks. By taking into account both the global  and the finer details, the system becomes more effective at understanding and accurately retrieving relevant information. In this direction, they proposed a model which embeds the fragments of images and fragments of sentences into a shared space  rather than mapping entire images or sentences into a common embedding space. They use two objectives to design their objective function, Global ranking objective ensures that an image-caption pair is consistent with the corresponding ground truth, and all sentence fragment appearances in the visual domain are learned via \emph{Fragment Alignment}. 
        
        In a similar vein, the authors in \cite{kiros} employs two Multi-Modal Log Bilinear Models i.e. \emph{Modality-Biased Log-Bilinear Model} (MLBL-B) and  \emph{Factored 3-way Log-Bilinear Model} (MLBL-F) which are extensions of the log bilinear model. 
        
        Modality-Biased Log-Bilinear Model is used to predict the next word representation with a bias introduced by the features extracted from the image. MLBL-F  model incorporates modality conditioning by gating the word representation matrix using the image features. The joint image-text feature learning mechanism involves using small feature maps  learned using \textit{k-means} to learn convolutional networks instead of using original images. Feature responses from the fully connected layer with ReLU activation are  used either for additive biasing or gating in the MLBL-B and MLBL-F models, respectively.
        Building upon their previous efforts, the researchers extended their work by integrating \emph{Structured Content Neural Language Models} (SC-NLMs) \cite{kiros2014unifyingvisualsemanticembeddingsmultimodal} to learn joint image-sentence embeddings. This approach involved utilizing LSTM networks for sentence encoding and SC-NLMs for caption generation.
        The SC-NLMs were particularly effective in extracting the structural components of sentences, thereby enhancing the model's understanding of the content. This integration led to a notable improvement in caption realism compared to earlier methodologies.
        
        Furthermore, the authors in \cite{7298856} use RNN for bi-directional image-sentence retrieval, incorporating a long-term visual memory to retain information about previously generated words.  The primary objective is to maximize the likelihood of a word and its associated visual features given the context of previous words and their corresponding visual representations.In essence, the network compares its visual memory, representing its prior statements, with the current visual input to predict the next word in the sequence. 
        
        Additionally, an image captioning \emph{Multi-Modal RNN model} (m-RNN) is presented in \cite{mao2014deep}. Each time frame of the suggested m-RNN model consists of the following five layers: two word embedding layers, a recurrent layer, a multi-modal layer, and a softmax layer. The syntactic and semantic meanings of the words are encoded by the two word embedding layers. The word-embedding layer, the recurrent layer, and the image representation are the three inputs of the multi-modal layer. To obtain the activation of the multi-modal layer, the activations of the three layers are mapped to the same multi-modal feature space and then added together.    
    }

    \subsection{Encoder-Decoder with Attention}
    {        
        The integration of attention mechanisms has notably enhanced image captioning quality by effectively capturing image semantics. By dynamically focusing on specific regions of the image, attention mechanisms facilitate the generation of contextually rich captions.
        
        In the earlier \emph{Show and Tell} model\cite{vinyals2015tellneuralimagecaption} the visual features were extracted from a CNN enhanced with batch normalization and were fed into an LSTM decoder. An advancement  in this trajectory was the \emph{Show, Attend and Tell} model \cite{10.5555/3045118.3045336},in which attention mechanism was applied on extracted visual features. These attended visual features were concatenated with word embeddings generated from the caption sentence and were passed on to the LSTM decoder. The use of attention mechanism enables the decoder to attend over specific regions in the image and word embeddings simultaneously.
        This helps in learning patterns between words and visual features in a shared embedded space. The attention function used in show, attend and tell, which is applied on extracted visual features can be a soft deterministic attention function or a hard stochastic attention function which captures the spatial attributes . 
        
        Top-down attention mechanisms use the associated caption in an image caption pair to selectively focus on specific parts of the image, whereas bottom-up attention mechanism works by first extracting salient regions from the input  image and then selectively attending to these regions to generate a caption,\cite{anderson2018bottomuptopdownattentionimage}  combines both these attention mechanisms to generate fine-grained image captions with deeper understanding.
        
        In\cite{anderson2018bottomuptopdownattentionimage}, \emph{Faster R-CNN} is used to implement bottom-up attention in order to produce spatial visual feature vectors. This model makes use of two LSTM layers: a top-down attention layer for the first and a language model for the second layer. The previous output of the language LSTM, along with the mean-pooled image feature and an encoding of the word that was previously generated, are fed into the attention LSTM at each time step.The attended image feature and the attention LSTM's output are combined to form the language LSTM's input.Based on the words that have previously been generated at earlier timesteps, the language LSTM forecasts the likelihood of each potential word that could appear next in the sequence at any time step \(t\). 
        
        Unlike conventional approaches where visual attention is active for every word generation step, adaptive attention model\cite{lu2017knowinglookadaptiveattention} makes a dynamic decision about whether to focus on the image or use a visual sentinel. The model introduces a concept of a visual sentinel, which serves as a proxy for non-visual or less visually dependent words in the caption. When the model decides not to attend to the image, it focuses its attention on this visual sentinel instead, which is stored in the decoder memory. This helps the model to generate words that are less reliant on visual context, such as articles \textit{("the", "a")}, prepositions, and other common language constructs.
        The adaptive attention module plays a crucial role in this process. It computes a new adaptive context vector that combines the visual sentinel vector and the context vector from the spatial attention model. 
        
        An attention mechanism that simulates the pairwise interactions between the RNN state, the image regions, and the caption words is proposed by areas of attention \cite{pedersoli2017areasattentionimagecaptioning}. 
        The region-state interaction term enables the model to emphasize and suppress image regions based on their appearance and the current state.  Additionally, the interactions between RNN state and words contribute to enhancing the model's ability to maintain temporal coherence in the generated word sequence, ensuring consistency as the model recursively conditions on all previous words.
        
        An attention module creates weighted average vectors based on the similarity scores between the key and query. These weighted average vectors may be irrelevant or misleading , to tackle this task attention on attention mechanism was proposed\cite{9008770}. The relevancy of the query and the attention outcome is determined by the \emph{Attention on Attention} mechanism. \textit{AoANet} model is introduced for image captioning which is based on the attention-on-attention mechanism, the attention-on-attention module is used to refine the extracted visual feature vectors. The self-attentive multi-head attention module in this refining module looks for interactions between the objects in the image, and the degree of their relatedness is measured using the \emph{AoA} module.
        
        It was observed that intra-modality information within each modality is complementary to inter-modality information. For instance, in the context of image feature extraction, each region of an image should gather insights not just from the words or phrases associated in the caption, but also from interconnected image regions. Similarly, for captions, gaining a deeper comprehension involves inferring meaning from additional words or phrases. Inter and intra-modality attention mechanisms are used in many vision-language tasks such as Visual Question Answering (VQA) to capture high-level interactions between vision and language domains. In \cite{peng2019dynamicfusionintrainter} a dynamic inter and intra-modality attention flow is proposed for multi-modal fusion.
        
        To deal with the problem of \emph{internal covariate shift} in self-attention mechanisms \cite{guo2020normalizedgeometryawareselfattentionnetwork} proposed normalized self-attention which performs normalization of hidden activations on self-attention to fix their distribution.In order to capture the intra-modal relation within an image in terms of geometry, geometry aware self-attention mechanism is proposed.
        
        Furthermore the authors in \cite{WANG2020107075} propose a context-aware attention mechanism that leverages previously attended visual content to guide attention selection using an additional interpolation gate which fuses the current normalized attention weight with previously produced attention weights.

         \begin{figure}[H]
        \centering
            \includegraphics[width=0.9\linewidth]{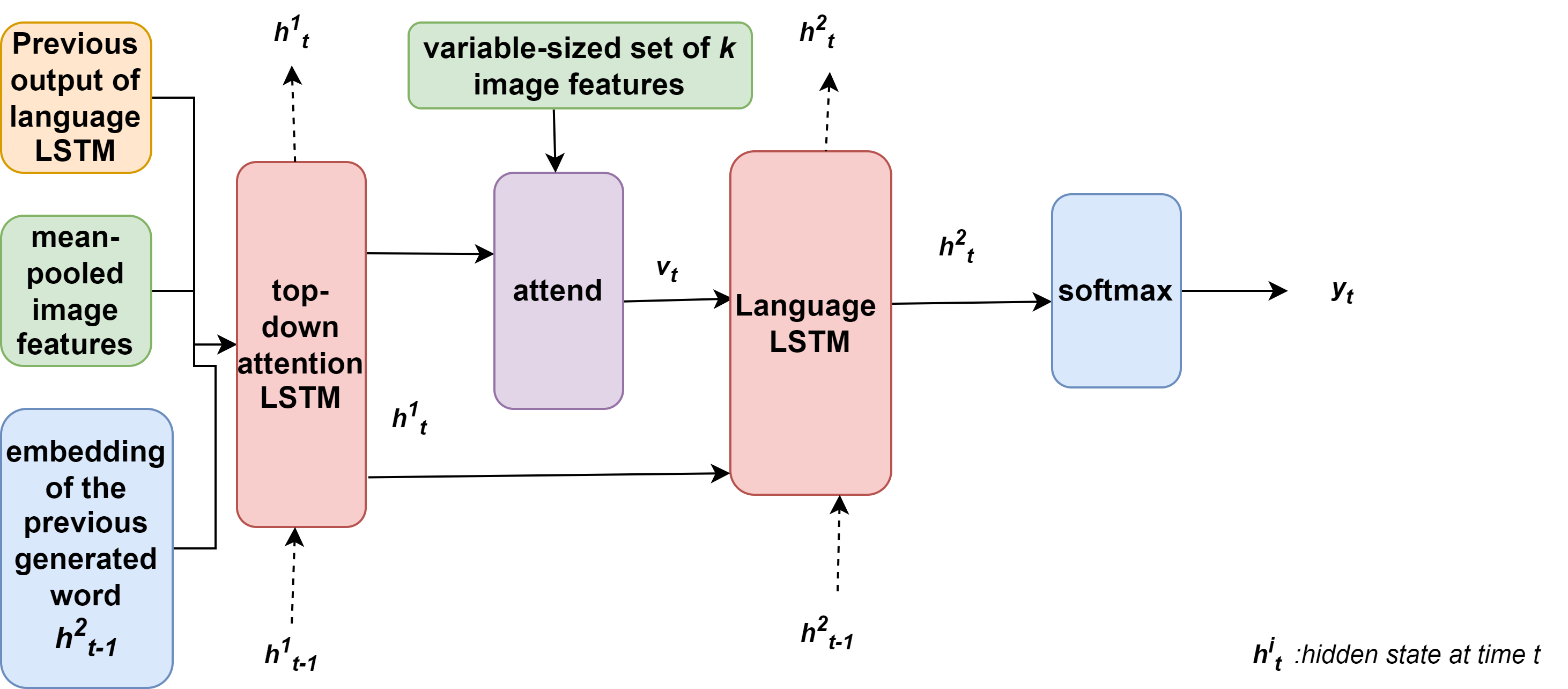}
            \caption{Overview of bottom-up and top-down attention model\cite{anderson2018bottomuptopdownattentionimage}}
            \label{fig:attention}
        \end{figure}
    }

    \newpage
    {

        \setlength{\arrayrulewidth}{0.1mm}
        \setlength{\tabcolsep}{8pt} 
        \renewcommand{\arraystretch}{1}
        \begin{longtable}{ |m{3cm}|m{7cm}|m{5cm}| } 
        \hline
        \multicolumn{3}{|c|}{Table 1: Listing the highlights and limitations of the discussed attention mechanisms} \\
        \hline
        \textbf{Attention Mechanism} & \textbf{Highlights} & \textbf{Limitations} \\
        \hline
        \endfirsthead
        \hline
        \multicolumn{3}{|c|}{Continuation of Table 1} \\
        \hline
        \textbf{Attention Mechanism} & \textbf{Highlights} & \textbf{Limitations} \\
        \hline
        \endhead
        \hline
        \endfoot
        \hline
        \endlastfoot
        \hline
        Show, attend and tell\cite{10.5555/3045118.3045336}
        & {
            \begin{itemize}
                \item Features are extracted from lower convolutional layer
                \item LSTM with attention is used as a decoder
                \item Soft and hard attention mechanisms are used to determine the visual features to attend to
                \item Generates better captions by attending to visual features than the previous show and tell model
            \end{itemize}
        }
        & Attends to visual features from the whole image rather than a particular region \\
        \hline
        Bottom-up and Top-down attention\cite{anderson2018bottomuptopdownattentionimage} 
        & {
            \begin{itemize}
                \item Bottom-up component extracts a diverse set of visual features from the input image
                \item Top-down mechanism employs attention to focus on specific regions of the image that are most relevant to the generation of each word in the caption
                \item Capable of generating semantically rich captions
            \end{itemize}
        }
        & Non-adaptive spatial attention - decoder does not know when to use visual signals and when to use language models for generating the next word in the sentence \\
        \hline
        Adaptive attention\cite{lu2017knowinglookadaptiveattention}
        & {
            \begin{itemize}
                \item Adaptive attention mechanism employed by decoder which doesn't require visual attention to generate non-visual words
                \item Better visual-to-text alignment achieved, comparable with human intuition \cite{lu2017knowinglookadaptiveattention}
            \end{itemize}
        }
        & {
            \begin{itemize}
                \item The model lacks explicit mechanisms to discriminate between words that are truly non-visual and those that are technically visual but have a high correlation with other words\cite{lu2017knowinglookadaptiveattention}
                \item Spatial attention maps derived from coarse-feature maps result in loss of spatial resolution and detail for smaller objects \cite{lu2017knowinglookadaptiveattention}
            \end{itemize}
        } \\
        \hline
        Intramodal and Intermodal attention\cite{peng2019dynamicfusionintrainter}
        & {
            \begin{itemize}
                \item This mechanism enables the information flow within and across different modalities
                \item These types of models are useful in various multimodal vision-language tasks such as VQA and image-text-retrieval tasks
            \end{itemize}
        }
        & Solely integrating the \emph{Intra-Modality Attention Flow} (IntraMAF) harms performance of a VQA model because too many unrelated information flows hinder the learning process \cite{peng2019dynamicfusionintrainter} \\
        \hline
        Context-aware attention\cite{WANG2020107075}
        & {
            \begin{itemize}
                \item Models the relationship between various regions of interests in an image using a graph neural network
                \item Previously attended visual content is memorized and used to guide attention, to prevent attending to the same regions at different time steps
                \item Captures rich visuals\cite{WANG2020107075}
                \item Model attends to more diverse and salient regions, such as the head region of a zebra and the snow region, when generating relation words like "\textit{standing}" and "\textit{riding}" \cite{WANG2020107075}
            \end{itemize}
        }
        & Model parameters grow exponentially upon increasing the hyperparameters \cite{WANG2020107075} \\
        \hline
        Normalized self-attention\cite{guo2020normalizedgeometryawareselfattentionnetwork}
        & {
            \begin{itemize}
                \item Compensates for the lack of transformers to model geometric relations
                \item Uses geometry-aware self-attention for obtaining geometric relationships between objects in an input image
                \item Uses normalized self-attention to eliminate the problem of internal-covariate shift
            \end{itemize}
        }
        & Relative geometry information is advantageous in improving the performance of self-attention network over absolute geometry attention\cite{guo2020normalizedgeometryawareselfattentionnetwork} \\
        \hline
        Areas of Attention\cite{pedersoli2017areasattentionimagecaptioning}
        & {
            \begin{itemize}
                \item Novel attention mechanism which uses three pairwise interactions between RNN state, image regions, and word embedding vectors
                \item This mechanism improves the attention focus and generates rich captions
                \item Experimental results indicate that RNN state and region interaction enables the RNN to dynamically adjust its attention to favor certain regions over others based on compatibility scores \cite{pedersoli2017areasattentionimagecaptioning}
            \end{itemize}
        }
        & As the model becomes more complex with additional interactions, interpreting the attention mechanism and understanding how specific interactions contribute to the model's decision-making process becomes more challenging. This can reduce the model's transparency and interpretability \\
        \hline
        Attention on Attention\cite{9008770}
        & {
            \begin{itemize}
                \item AoA adds another attention layer to measure the relevance between the attention result and the query
                \item This refining process ensures that the attention module returns a relevant output and only useful information is fed to the decoder
            \end{itemize}
        }
        & Attention-on-attention might lead to increased model complexity. This complexity can make the model more challenging to train, optimize, and interpret \\
        \hline
        \end{longtable}
    }

\newpage
    
    \subsection{Encoder-Decoder with Transformer}
    {
        In transformer based image captioning models, both the encoder and the decoder employ transformer-like layers and attention mechanisms. In order to obtain the visual representations,mainly two approaches are used, using a pre-trained object detector or a vision-transformer. Vision transformers are becoming compelling alternatives to the pre-trained object detectors as transformer-like architectures can be directly applied to image patches thus limiting or eliminating the need for a convolutional operator. Vision-transformers pre-trained for multimodal retrieval are employed to extract enriched visual features.
        
        A self-attention and a feed-forward layer make up each encoder layer, while a self-attentive, a cross-attentive, and a feed-forward layer make up each decoder layer.\cite{10.3233/AIC-210172}
        
        A typical encoder-decoder architecture with a transformer for image captioning can be improved by using the spatial information, integrating semantics and geometrical relationships between objects.Researchers have developed a number of improvements to produce more detailed captions, which we go over in this section. 
        The authors of\cite{herdade2020imagecaptioningtransformingobjects} suggest an object relation transformer that uses geometric attention to integrate the spatial arrangement among identified objects in the input.The visual encoder is a pre-trained object detector. They use \emph{Faster R-CNN} with \emph{ResNet-101} for object detection and feature extraction, these feature vectors are then input into the transformer model.
        Using the geometry features of the two objects a displacement vector is calculated.This displacement vector is used to generate bounding boxes corresponding to the two objects. In order to compute attention weights based on the geometry of these two objects, a high-dimensional embedding is calculated using the displacement vector. Multiplying the high-dimensional embedding with a learned vector and applying ReLU for non-linearity gives the geometric attention weights.

        \begin{figure}[H]
            \centering
            \includegraphics[width=0.7\linewidth]{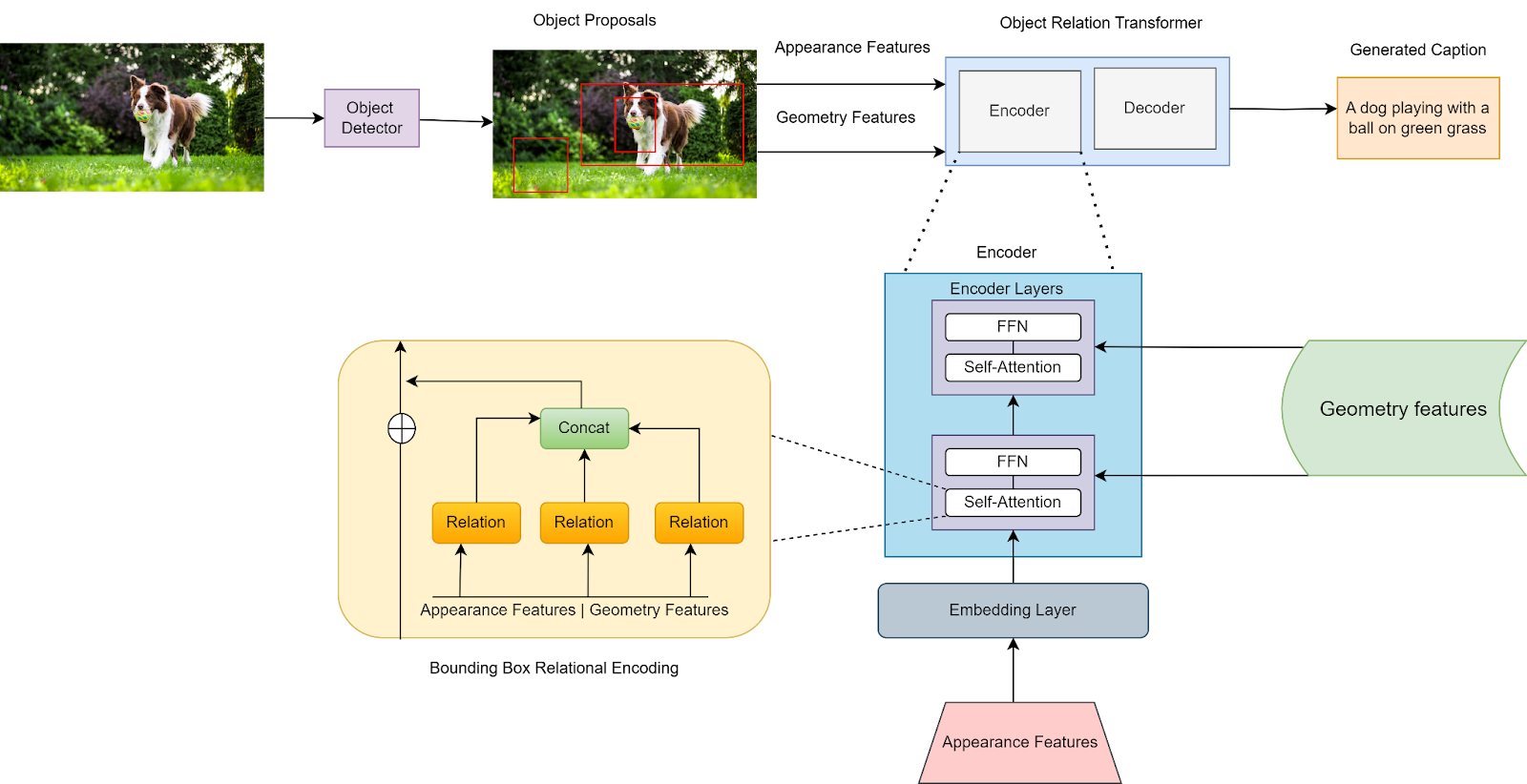}
            \caption{Overview of Object Relation Transformer architecture \cite{herdade2020imagecaptioningtransformingobjects}}
            \label{fig:object-relation-transformer}
        \end{figure}

        Typical attention mechanisms tend to find it difficult to identify equivalent visual signals. The authors in\cite{9008532} proposed the idea of using semantic attributes from language to better explicitly identify the visual signals.They propose a framework which utilizes the vision and language modalities simultaneously to close the semantic gap. Information flow from the visual and language modalities is guided using a \emph{Gated Bilateral Controller} (GBC).

        In order to generate fine-grained captions which capture the semantic attributes in an image, a Transformer-based sequence modeling framework \emph{Entangled Attention} (ETA) is proposed, built only with attention layers and feed forward layers. It utilizes the information from the complementary modalities; the visual and the semantics.They employ two sub-encoders, one dedicated to encoding visual characteristics and another focused on semantic attributes.The output of these two encoders is fed to the multimodal decoder.The multimodal decoder block integrates an ETA module between the self-attention sub-layer and the feed-forward layer. This module enables the decoder to concurrently apply attention to both the visual and semantic outputs of the dual encoder.
        
        Typically, attention is applied separately over multiple modalities and later their representations are fused together, but in this case attention is applied over multiple modalities in an entangled manner. In order to perform multi-head attention on a particular modality, say visual modality, semantic guidance is used and vice-versa.


         In various attempts to enhance the typical transformer based encoder-decoder architecture, the authors in\cite{cornia2020meshedmemorytransformerimagecaptioning} use a different design to link the encoder and the decoder architecture. The authors opt for a novel approach by feeding the decoder with the encoder output in a meshed manner. Specifically, they leverage both low-level and high-level visual relationships from the encoder, enhancing the decoder's input and potentially improving performance. Furthermore, the encoder model incorporates a memory-augmented attention operator. In this setup, the set of keys and values utilized in self-attention is expanded by incorporating additional slots capable of encoding a priori information. These supplementary keys and values are realized as learned vectors to ensure that they do not depend on the input set. Meshed cross-attention is employed in the decoder module, where rather than attending solely to the final encoding layer, cross-attention is conducted across all encoding layers.

        \begin{figure}
            \centering
            \includegraphics[width=0.7\linewidth]{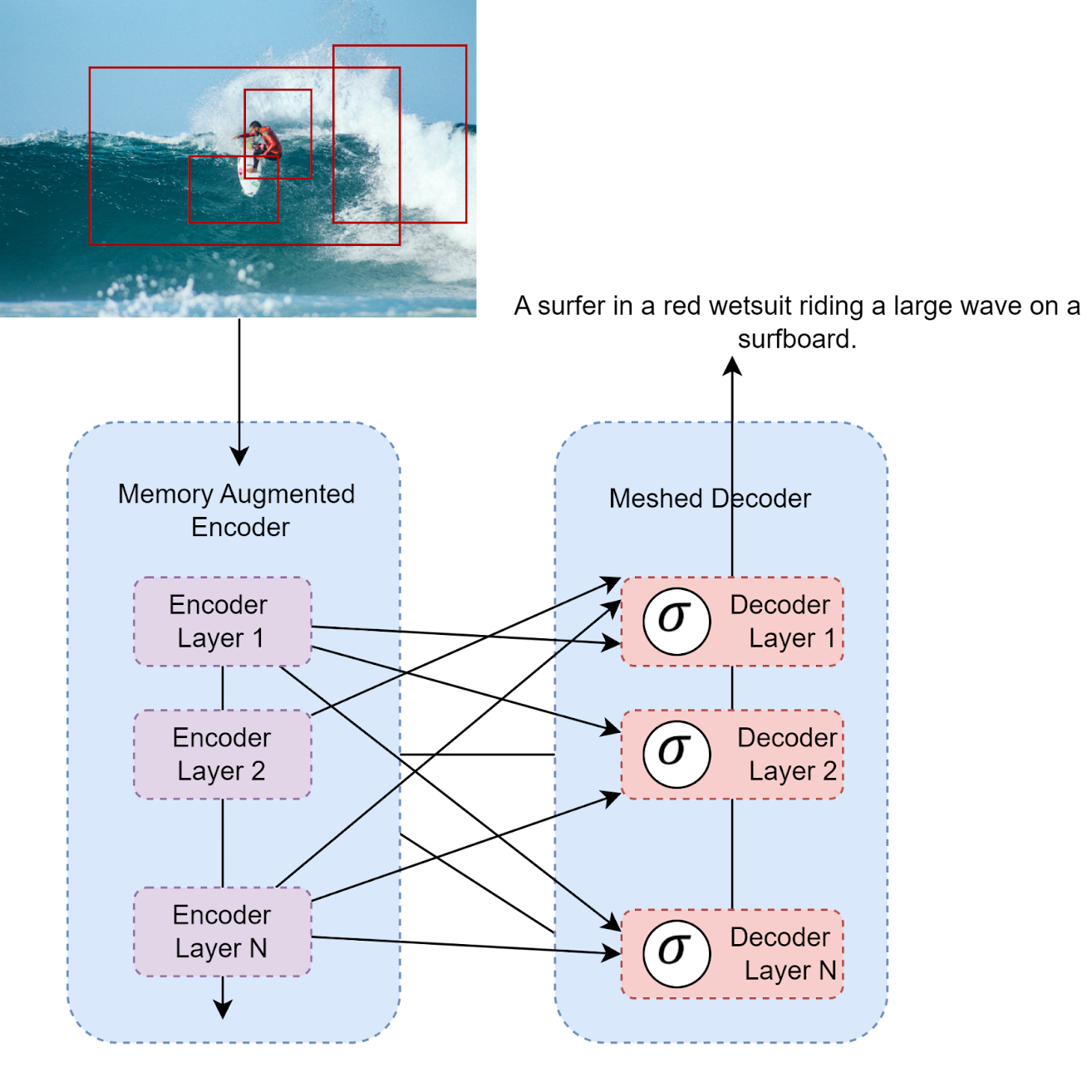}
            \caption{$M^{2}$ Transformer \cite{cornia2020meshedmemorytransformerimagecaptioning}}
            \label{fig:m2-transformer}
        \end{figure}

        The \emph{Dual-Level Collaborative Transformer} (DLCT)\cite{luo2021dual} is another transformer based model which aims to incorporate multiple levels of features to enhance the generation of captions or textual outputs.It incorporates both global grid features and local region features to generate seamless captions. In order to understand both the individual characteristics of each feature type and their spatial relations, Comprehensive relation attention scheme is used.A geometric alignment graph is constructed to guide semantic alignment between region and grid features, facilitating precise interaction.It achieves this by using cross-attention fusions, which involve transferring objectness information from region features to grid features and supplementing fine-grained details from grid features to region features. 
        
        In the Meshed Memory Transformer, the decoder engages in cross-attention with both low-level and high-level encoder layers, enhancing the generation of fine-grained and contextually rich captions. Similarly, in DLCT, cross-attention is employed to merge grid features and regional features of an image, leading to the creation of semantically rich captions.

        \begin{figure}[b!]
            \centering
            \includegraphics[width=1\linewidth]{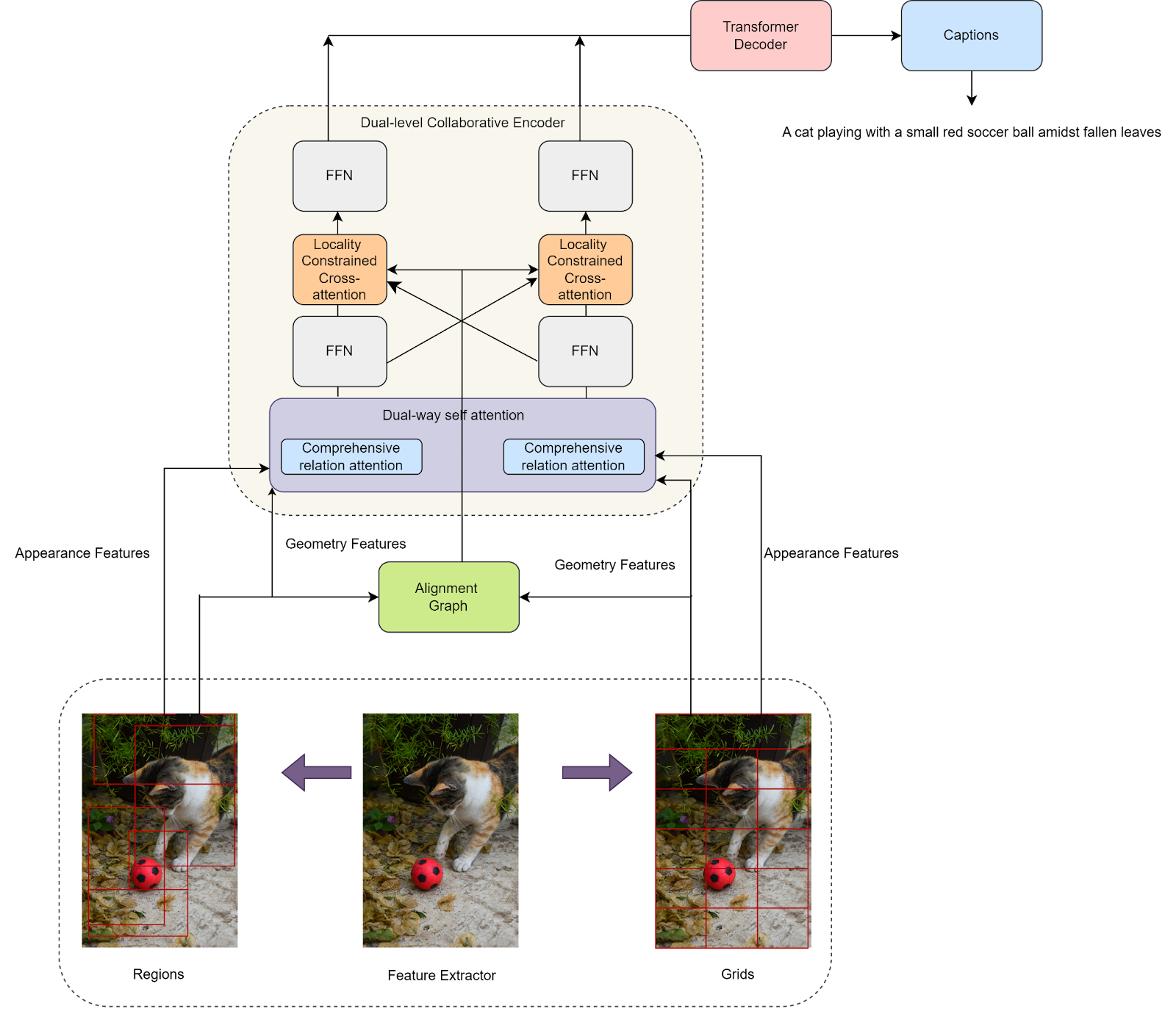}
            \caption{Dual-Level Collaborative Transformer Architecture \cite{luo2021dual}}
            \label{fig:dlc-transformer}
        \end{figure}


        \vspace{-10pt}
        
        Various techniques were employed by the researchers to retrieve the visual features to capture the semantics and spatial information to generate richer and detailed captions since the quality of the visual features extracted has direct correlation with the quality of captions generated. To enhance the quality of captions efforts were also directed towards improving the language decoder, such as employing knowledge graphs to improve the contextuality of the generated captions. In \cite{ZHANG202143} the authors leverage knowledge graphs in transformers to generate image captions. Typically a transformer uses word embedding vectors as inputs. These word embedding vectors use the information  based on themselves. In order to enhance the transformer’s performance the authors construct a knowledge graph for each word in the vocabulary, the knowledge graph enables to use the information not only from the word itself but also its neighboring words.

        \vspace{10pt}
        
        The vocabulary is searched for the top $N$ nearest terms for each word, which are  considered as the word's neighbor nodes. The cosine similarity score is determined using the word embedding as the word representation. The words that rank among the top $N$ greatest values of cosine similarity for each word are chosen to be its neighbor nodes. 
        We summarize the transformer based architectures reviewed in this section in table~\ref{tab-2}.

        \vspace{30pt}

        \vspace{10pt}
        
        \begin{figure}[H]
             \centering
           \includegraphics[width=1\linewidth]{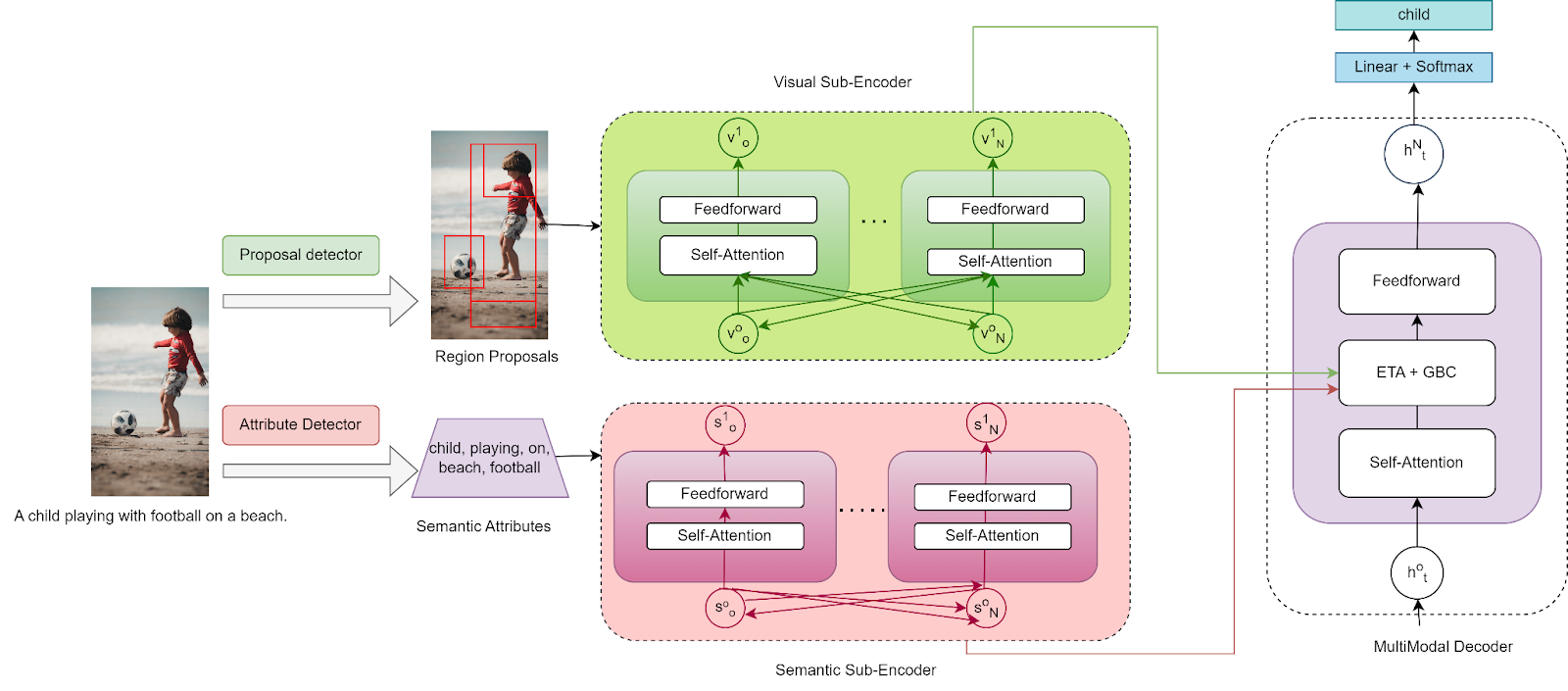}
            \caption{Overall Architecture of ETA-Transformer \cite{9008532}}
             \label{fig:ETA-transformer}
         \end{figure}

          \begin{table}[h!]
            \centering
            \begin{adjustbox}{max width=\textwidth} 
            \begin{tabular}{|m{4cm}|m{6cm}|m{6cm}|m{1.5cm}|m{1.7cm}|m{1.5cm}|m{1.5cm}|}
                \hline
                \multicolumn{3}{|c|}{} & \multicolumn{4}{|c|}{\textbf{Evaluation Metric}}\\
                \hline
                \textbf{Transformer Based Model} & \textbf{Highlights} & \textbf{Limitations} & \textbf{BLEU-4} & \textbf{METEOR} & \textbf{CIDEr} & \textbf{SPICE} \\
                \hline
                
                Object Relation Transformer\cite{herdade2020imagecaptioningtransformingobjects} & Captures spatial relationships for context understanding. Dynamically adjusts the importance of different regions based on their spatial relationships. Utilizes positional encoding for sequencing logic & Errors in identifying rare and unusual objects, relations, and attributes. As compared to the ground truth captions, the captions generated by this model were less descriptive & 38.6 & 28.7 & 128.3 & 22.6 \\
                \hline
                ETA\cite{9008532} & Uses entangled attention to utilize the visual and language information simultaneously. Visual modality guides attention to the language modality and language modality guides attention to the visual modality. Generates more semantically rich captions & Activation functions such as $Tanh$ and $Sigmoid$ considerably impair GBC performance & 39.3 & 28.8 & 126.6 & 22.7 \\
                \hline
                Meshed Memory Transformer\cite{cornia2020meshedmemorytransformerimagecaptioning}  & Meshed connectivity brings an improvement of 7.6 points in CIDEr score over standard transformers. Using a priori information improves visual encoding. Capable of out-of-domain-captioning & In the $M^{2}$ Transformer, complicated non-linear dependencies affect how much a single region contributes to the output, in contrast to recurrent-based captioning models where attention weights over regions are used which can be easily recovered. It is difficult to properly discern the relevance of particular sections in the generated captions due to this intricacy. & 39.1 & 29.2 & 131.2 & 22.6 \\
                \hline
                Dual Level Collaborative Transformer \cite{luo2021dual} & Uses cross-attention to merge grid and regional features to generate semantically rich captions & In comparison to the X-Transformer model\cite{pan2020xlinearattentionnetworksimage}, DLCT performs slightly worse in SPICE score. DLCT (23.3) versus X-Transformer (23.8), on RestNext-101 backbone. & 39.8 & 29.5 & 133.8 & 23.0 \\
                \hline
                Transformer with Knowledge Graph\cite{ZHANG202143} & Leveraging a knowledge graph during training results in a more robust model due to its exposure to flexible ground truth & A limitation is that the model's optimal performance was achieved on MS COCO and Flickr30k datasets with different hyperparameters, indicating potential challenges in achieving consistent performance across diverse datasets without fine-tuning. & 33.72 & 27.45 & 110.30 & 20.64 \\
                \hline
            \end{tabular}
            \end{adjustbox}
            \vspace{5pt}
            \caption{\label{tab-2}Comparison of Transformer Based Image Captioning Models (\emph{Dataset: MS COCO with Karpathy Split})}
            \label{tab:comparison_transformers}
        \end{table}
        
    }

    {
        \begin{table}[h!]
            \centering
             \begin{adjustbox}{max width=\textwidth}
            \begin{tabular}{|m{3.2cm}|m{6cm}|m{6cm}|}
                \hline
                \textbf{Feature/Technique} & \textbf{Strength} & \textbf{Weakness} \\
                \hline
                Global Image\newline Features & Simple, Compact, provides overall context & Treats significant and non-significant information alike, poses a challenge in generating more fine-grained and accurate captions. \\
                \hline
                Regional Image \newline Features & Localized, captures salient information & Focuses only on entity regions, doesn’t capture the topological structure and object relationships. \\
                \hline
                MultiModal Fusion & Helps in bidirectional image-text retrieval, Enhanced contextual understanding & Poor alignment or fusion of visual and textual embeddings may result in inaccurate captions, Multimodal models may find difficulty in adapting to domain shifts and variations. \\
                \hline
                Scene Graphs & Captures object-attributes relationships and spatial arrangements in an image, provides topological representation of a scene & Limited representation of dense interactions, Complex hierarchical interactions may pose challenges. \\
                \hline
                Attention & Enables selective focus on relevant features, facilitates alignment between image regions and corresponding words in generated captions, incorporates contextual information, enables adaptive focus on different image regions based on their relevance to the current word being generated, ability to handle diverse images with varying compositions, contents, and complexities & Computational overhead. \\
                \hline
                Transformer & Allows parallel processing of input, effective in capturing long-range dependencies, employs attention mechanism to focus on relevant regions when generating captions, scalable, transfer learning & High computational cost, requires large amount of data for pre-training. \\
                \hline
              
            \end{tabular}
            \end{adjustbox}
            \vspace{5pt}
            \caption{Summarization of various architectural innovations discussed in section $5$, along with their strengths and weaknesses}
            \label{tab:summarization}
        \end{table}
    }

}

\newpage

\section{Vision Language Pre-Training}
{
    {
    \begin{figure}[H]
        \centering
        \includegraphics[width=0.8\linewidth]{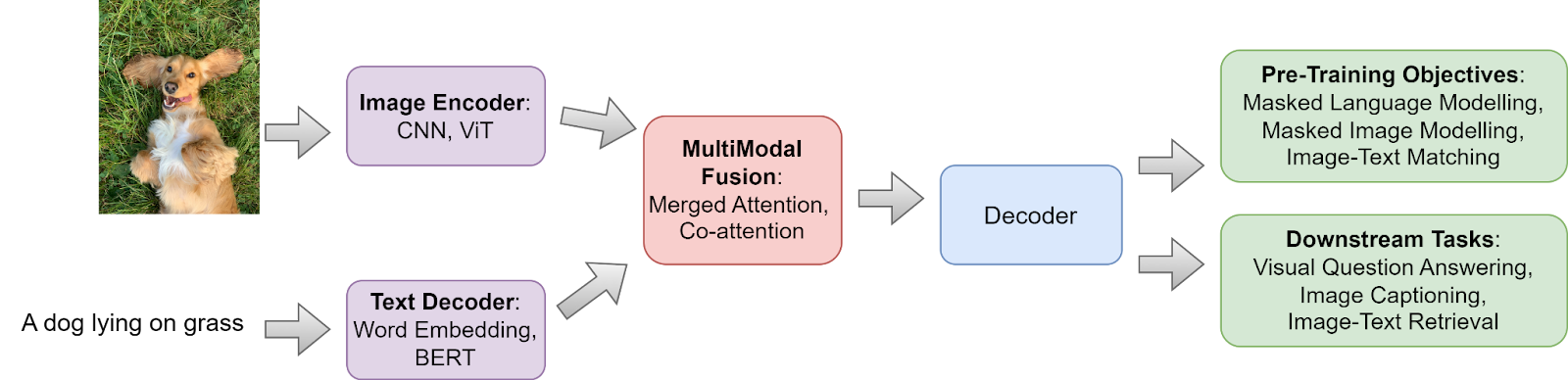}
        \caption{General Framework for Transformer-based vision-language model\cite{gan2022visionlanguagepretrainingbasicsrecent}}
        \label{fig:vlp}
    \end{figure}
    }
    {
    With the rapid progress in the world of AI, we expect artificial intelligence models to learn about the world just like humans do. Humans use information from multiple channels  such as vision and text and extract meaningful information from them for their learning. Based on the similar concept we have seen a new training paradigm Vision-language pre-training which mirrors human cognition's ability to synthesize information from diverse channels such as vision and language. This novel training paradigm enables the seamless integration of visual and textual data, facilitating tasks like image captioning, VQA, and image-text retrieval.
    
    We will be looking at the task of image captioning from the lens of vision-language pre-training(VLP). We have looked at image captioning models which were task-specific. They were mainly based on sequence to sequence modeling based on machine translation framework. VLP models are multi-modal in nature and can perform many vision language tasks such as image captioning, VQA, and video captioning seamlessly. Tasks involving vision and language can be divided into two categories: understanding and generation-based tasks.\cite{gan2022visionlanguagepretrainingbasicsrecent}
    
     Understanding-based tasks involve comprehending the content of visual input and making sense of it, often through tasks like object recognition, scene understanding, or VQA. On the other hand, generation-based tasks focus on producing meaningful textual descriptions or responses based on visual input, such as image captioning or text-to-image generation.\cite{gan2022visionlanguagepretrainingbasicsrecent}
    
    Before the development of big-scale VLP, early \emph{Vision-Language} (VL) models had a straightforward architecture: a pre-trained visual encoder extracted the image attributes first, and a textual encoder extracted the textual features. In order to obtain cross-modal representations, multi-modal fusion was performed on these extracted features. Major work has been done on the use of attention on the multi-modal fusion module. These VL models are categorized based on the attention mechanisms used to enhance the multi-modal fusion module. Eventually with the advent of transformers as universal computation engines, VL models evolved into transformer based architectures.\cite{gan2022visionlanguagepretrainingbasicsrecent}
    
    The introduction of BERT\cite{devlin2019bertpretrainingdeepbidirectional}[50] in NLP marked a pivotal moment for vision-language problems, driving a transition towards transformer-based multi-modal fusion models. These models are pre-trained on extensive image-text datasets, encompassing up to 4 million images and roughly 10 million image-text pairs. Examples include UNITER\cite{chen2020uniteruniversalimagetextrepresentation} and OSCAR\cite{li2020oscarobjectsemanticsalignedpretraining}. More recently, the landscape has been reshaped by large-scale multi-modal foundation models like SimVLM, Florence\cite{yuan2021florencenewfoundationmodel}, Flamingo\cite{alayrac2022flamingovisuallanguagemodel}, CoCa\cite{yu2022cocacontrastivecaptionersimagetext}, and GIT\cite{wang2022gitgenerativeimagetotexttransformer}. These models have been pre-trained on an even larger scale, with over 12 billion image and text pairs. Their extensive pre-training enables adaptation to a broad spectrum of downstream tasks.\cite{gan2022visionlanguagepretrainingbasicsrecent}

    \subsection{Architecture of  Vision Language models}
    
        {
        In the early stages of VL model development for image captioning, architectures were primarily influenced by the encoder-decoder framework derived from \emph{Sequence-to-Sequence} (seq2seq) learning. These models used a visual encoder to extract features from images, which were subsequently used to produce captions using a text decoder. To optimize the utilization of visual features by the text decoder, various multi-modal fusion methods were explored, often incorporating attention mechanisms. These attention mechanisms, including both inter-modality and intra-modality attention, were integrated into the encoder-decoder architecture to enhance performance. Inter-modality attention \cite{lu2017knowinglookadaptiveattention, 9008770,10.5555/3045118.3045336}, implemented on the decoder side, facilitated the generation of captions based on specific image regions or concepts of interest, while intra-modality attention focused on incorporating semantic and spatial object relations into the visual encoder. These advancements played a pivotal role in improving the quality and accuracy of image captioning generated by early VL models.
        
        For instance,the authors in \cite{yao2018exploringvisualrelationshipimage} introduced a novel approach by incorporating a \emph{Graph Convolutional Network} (GCN) into the visual encoder. This integration allowed for the representation of both semantic and spatial relationships among objects within the image. By leveraging graph structures, the model could capture intricate connections between objects, enhancing its understanding of the scene's context.
        
        Similarly, the authors of \cite{herdade2020imagecaptioningtransformingobjects} proposed the use of an object relation Transformer, which utilizes geometric attention mechanisms to explicitly take into account the spatial relationships between input objects. This transformer-based architecture focuses on encoding the interactions between objects based on their spatial proximity and relative positions within the image. By attending to the geometric arrangement of objects, the model can better capture the spatial context of the scene, resulting improved performance in image captioning.\cite{gan2022visionlanguagepretrainingbasicsrecent}
        }

    \subsection{Vision Language Pre-Training Methods}
        {
         Within the realm of image-text tasks explored in scholarly literature, our investigation centers on the application of VLP to image captioning. The growing interest in VLP is due to two key factors. Firstly, the remarkable success achieved in language model pre training has catalyzed interest in extending similar methodologies to multi-modal data. Secondly, the convergence of NLP and computer vision architectures has opened up fresh directions for interdisciplinary research. In practice, VLP entails training models on large sets of image-caption pairs, allowing them to capture diverse multi-modal information.. These pre-trained representations are helpful  for a range of downstream tasks, underscoring the significance of VLP across various vision-language tasks.
         VLP methods can be broadly categorized into two main types: \emph{\textbf{dual encoder}} and \emph{\textbf{fusion encoder}} architectures.
         
         In dual encoder models, images and text are independently encoded, with modality interaction limited to a straightforward cosine similarity calculation between the resulting feature vectors.
         Conversely, fusion encoder architectures employ additional Transformer layers alongside separate image and text encoders to facilitate more intricate interactions between image and text representations. Notable examples of fusion encoder models include UNITER, VinVL\cite{zhang2021vinvlrevisitingvisualrepresentations}, SimVLM\cite{wang2022simvlmsimplevisuallanguage}, and METER\cite{dou2022empiricalstudytrainingendtoend}. While dual encoder models excel in tasks such as image retrieval, fusion encoder architectures demonstrate superior performance in image captioning and VQA tasks. This review will focus primarily on fusion encoder methods within the context of VLP approaches for image captioning.\cite{gan2022visionlanguagepretrainingbasicsrecent}

        \vspace{10pt}
        
        \underline{\textbf{Fusion encoder methods}}:
        
        Fusion encoder comprises multiple layers of transformer blocks which efficiently combine fine-grained features from images-text pairs.First the image features and textual words are encoded into a common embedded space.This process ensures that information from both modalities is represented in a unified format. Then a transformer module is applied to this common embedding to learn contextualized embeddings for each region and word through various pre-trainings tasks. E.g. UNITER, SimVLM, METER.
        
        Fusion encoder methods further have two categorizations - \textbf{Pre-trained object detector} based VLP modes and \textbf{end-to-end pre-training} methods. 
        
        \textbf{Pre-trained Object Detector based Methods} - These techniques extract visual characteristics using pre-trained object detectors, and the multi-modal fusion module generates multi-modal representations using a variety of attention processes, including co-attention and merged attention. Cross-attention is later utilized to enable cross-modality collaborative learning; in co-attention, distinct transformers are used to extract region-based visual and textual information.E.g. ViLBERT\cite{lu2019vilbertpretrainingtaskagnosticvisiolinguistic}, LXMERT\cite{tan2019lxmertlearningcrossmodalityencoder}. In merged attention visual and textual features are fed into a transformer block after being concatenated.E.g. UNITER\cite{chen2020uniteruniversalimagetextrepresentation} ,OSCAR\cite{li2020oscarobjectsemanticsalignedpretraining}.
        
        \textbf{End-to-End pre-training methods} - In end-to-end pre-training methods various methods are used to extract visual features, mainly convolutional neural networks and Vision transformers are used to extract visual features. The model gradients are then back-propagated to the vision backbone for an end-to-end training. Various pre-training tasks are used for end-to-end training such as Masked language modeling and  image-text-matching.
        Pixel-BERT\cite{huang2020pixelbertaligningimagepixels} uses CNN for visual embedding learning and deep multi-modal transformers to jointly learn visual and language embeddings.
        SOHO\cite{huang2021seeingboxendtoendpretraining} extracts visual features using CNN and then aggregates these features into a visual dictionary. Later a transformer is used to fuse visual and language modalities.
        Vision transformers are used in various VLP models to encode visual information. Some significant examples include, ViLT\cite{kim2021vilt} which directly feeds concatenated image and word embeddings into a transformer encoder. 

        }

    \subsection{Pre-training objectives of VLP models}
        {
         \underline{Masked Language Modeling (MLM)}: In MLM, the model processes image-text pairs by randomly hiding words with a 15\% chance and replacing them with [MASK]. The objective is to predict these masked words in order to reduce the negative log-likelihood, taking into account the context of the surrounding words and the paired image.

        \underline{Image-Text Matching (ITM)}: ITM involves determining whether a batch of image-caption pairs match or not. The model learns to associate the correct image with its corresponding caption, distinguishing between matched and mismatched pairs.
        
        \underline{Masked Image Modeling (MIM)}: In MIM, the model reconstructs masked patches or regions within an image based on the visible parts and accompanying text. It learns to fill in missing visual information, optimizing the reconstruction process \cite{gan2022visionlanguagepretrainingbasicsrecent}.

        }

    \subsection{Case Studies of various VLP models:}
    {
         \textbf{Unified Architecture} -

        Normal approaches to combining vision and language tasks typically involve creating separate architectures and objectives for each task. However, there's a growing interest in developing a unified framework that can handle various tasks within a single architecture.
        
        Various vision-language tasks can be accomplished by unifying several image and text related tasks as a single objective, such as text-generation or language modeling, the authors in \cite{cho2021unifyingvisionandlanguagetaskstext} introduce VL-T5 and VL-BART which unifies various image-text tasks with a unified text-generation objective, they use pre trained language models for text generation and eliminate the need for task-specific hand crafted architectures and objectives.
        
        They created a label text by encoding textual and visual inputs using a transformer encoder-decoder architecture. The decoder attends to prior tokens and the encoder outputs to predict the probability of the future text tokens, while the encoder integrates text and visual embeddings to create joint representations. With input text and image taken into account, the model parameters are trained to minimize the negative log-likelihood of label text tokens\cite{cho2021unifyingvisionandlanguagetaskstext}. This process seeks to generate text that maintains coherence and meaning irrespective of the input format—image, audio, or video. The focus lies on generating text that accurately reflects the information conveyed by the various input modalities, maintaining relevance and coherence throughout the generated content \cite{cho2021unifyingvisionandlanguagetaskstext}.
        
        \textbf{Frozen} - this method uses the pre-trained large language model for various multi-modal vision language tasks. The architecture of frozen consists of a visual encoder which takes raw image as input, the vision encoder’s outputs are linearly mapped and reshaped to generate a sequence of continuous embeddings,so that these visual representations can be fed into the pre-trained language models.Frozen is different than other models in a way that to apply this method to various vision language tasks does not need updating the weights of the language transformer as activations move through the model, the system becomes more adept at handling tasks that involve both vision and language, gradually adapting and enhancing its performance on multi-modal challenges. Only the visual encoder's parameters are updated during training. Frozen has shown rapid adaptation to new visual-language tasks, proving to be a multi-modal few-shot learner. Even though Frozen is initially trained on individual image-text pairs, it demonstrates the capability to proficiently handle sequences of multiple images and words. This versatility enables users to present it with various examples of new multi-modal tasks for evaluation or to introduce it to new visual categories before seeking information about them promptly\cite{tsimpoukelli2021multimodalfewshotlearningfrozen}.
        
        \textbf{Video Language Model} -The \emph{Video-Language Model} (VLM) simplifies pre-training by using a single BERT encoder, focusing on token masking without task-specific alignment. Its encoder combines existing models with two new methods for better multi-modal fusion: masked modality model (MMM) and a single masked token loss. MMM masks entire modalities for some examples, encouraging the use of tokens from one modality to predict the masked one. The single masked token loss enhances fusion by jointly leveraging video and text token embeddings. VLM also demonstrates the ability to fine-tune a single encoder for various tasks using task-specific attention masks, improving task performance\cite{xu2021vlmtaskagnosticvideolanguagemodel}.
        
        \textbf{OFA} represents a significant stride towards unified architectures for diverse vision and language tasks, including vision-only, language-only, and combined tasks. It serves as a task-agnostic and modality-agnostic framework, built upon a straightforward sequence-to-sequence learning approach.OFA uses ResNet modules to directly convolve image features into image patches, and linguistic information is processed using byte-pair encoding as followed in GPT and BART. This approach eliminates the need for complex data pre processing and modality-specific adaptors typically used in pre-training multi-modal transformers.OFA achieves task-agonistic behavior by formulating pre-training and fine tuning tasks in a unified sequence-to-sequence abstraction.For modality-agnostic behavior transformers are used as universal compute engines,with a constraint that no modality-specific component will be added by downstream tasks.Representations from different modalities are shared globally  across different tasks globally in a multi-modal vocabulary.Finally to achieve task comprehensiveness,that is its ability to apply the learned knowledge to a variety of unseen tasks,the model is pre-trained on a variety of uni-modal and cross-modal tasks such as image classification,VQA and  image captioning. OFA achieves competitive performance in zero-shot learning, and is able to adapt to out-of-domain information without fine tuning \cite{wang2022ofaunifyingarchitecturestasks}.
        
        \textbf{SimVLM} - Previous approaches have captured the images and text alignment by heavily relying on human-labeled information from various sources. This involved training a supervised object detector (OD) on object detection datasets to extract region-of-interest (ROI) features from images. Subsequently, datasets containing aligned image-text pairs were used for Masked Language Modeling (MLM) pre-training of a fusion model. The concatenated paired text and ROI features are fed into the fusion model. Additionally, task-specific auxiliary losses were introduced to address the limited scale of human annotated data and enhance performance. However, these design choices complicated the pre training protocol for VLP and created a bottleneck for additional quality enhancement. Furthermore, such pre training-fine tuning approaches often lacked zero-shot capability. SimVLM does not use auxiliary models such as Faster R-CNN\cite{ren2016fasterrcnnrealtimeobject} for region detection; instead, it uses only the language modeling loss on raw image inputs.Unlike other VLP approaches that require numerous pre training stages and auxiliary objectives, it just requires one pass pre training utilizing a single language modeling loss.All model parameters are pre-trained from scratch on noisy image-text data which helps in achieving zero-shot generalization. SimVLM achieves notable success in zero-shot image captioning tasks, effectively describing visual inputs with rich detail. Demonstrating a strong grasp of real-world concepts, it offers elaborate descriptions of complex scenes containing multiple objects. Its ability to comprehend fine-grained abstractions allows it to provide nuanced explanations, showcasing its proficiency in handling diverse visual contexts\cite{wang2022simvlmsimplevisuallanguage}.

        \begin{figure}[H]
            \centering
            \includegraphics[width=0.8\linewidth]{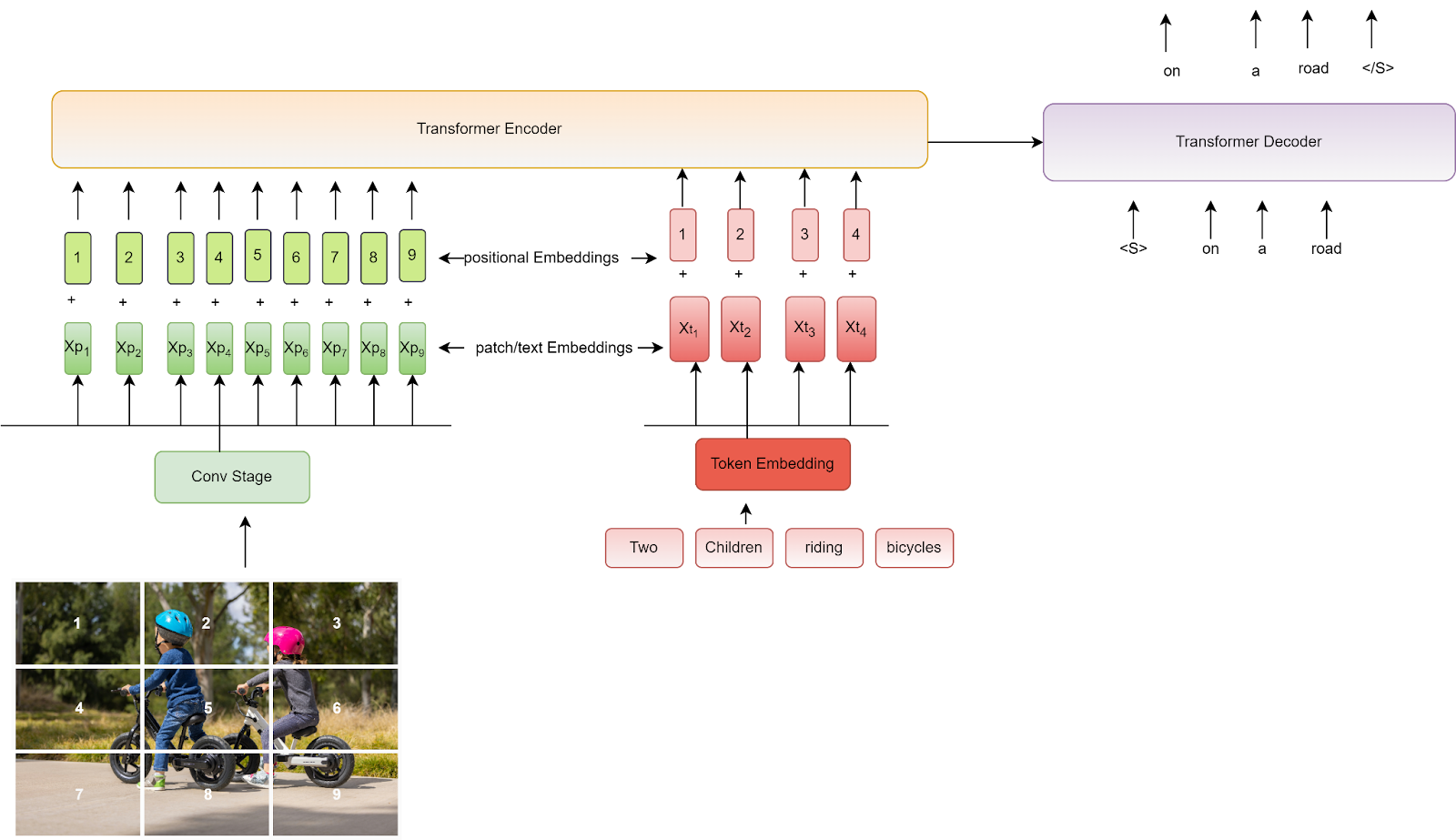}
            \caption{SIM-Vision-Language-Model \cite{wang2022simvlmsimplevisuallanguage}}
            \label{fig:simvlm}
        \end{figure}

        \textbf{FLAMINGO} - Pre-trained language models provide two fold benefits, firstly they save the computation power and time and secondly they store a large amount of knowledge in LM weights. Flamingo leverages these capabilities of pre-trained language models, to handle both open ended tasks such as VQA or image captioning and closed ended tasks such as classification. It accepts input as image or video and generates descriptive text. The authors leverage the power of pre-trained language models, to have strong generative language abilities.The architecture of Flamingo includes a vision encoder pre-trained with an approach similar to \emph{Contrastive Language-Image Pre-Training} (CLIP)\cite{jin2023selfsupervisedimagecaptioningclip}. Vision encoder extracts the semantic and spatial features from the given input images or videos. To bridge the vision encoder and pre-trained large language model together, their weights are frozen and two learnable architectures perceiver-resampler are used \cite{alayrac2022flamingovisuallanguagemodel}.

    \vspace{10pt}
    {
        \begin{table}[ht]
        \centering
         \begin{tabular}{ |p{3cm}||p{2cm}|p{2cm}|p{2cm}|p{2cm}|  }
         \hline
         \multicolumn{5}{|c|}{Evaluation Metric} \\
         \hline
         Model& BLEU-4 & METEOR & CIDER & SPICE\\
         \hline
         VL-T5\cite{cho2021unifyingvisionandlanguagetaskstext}  & 34.5   & 28.7 &  116.5 & 21.9\\
         
         V-BART\cite{cho2021unifyingvisionandlanguagetaskstext} &  35.1  & 28.7    &116.6 & 21.5\\
         
         OSCAR\cite{li2020oscarobjectsemanticsalignedpretraining} &34.5 & 29.1 & 115.6 & 21.9\\
         
         OFA\cite{wang2022ofaunifyingarchitecturestasks}    &43.9 & 31.8 & 145.3 & 24.8\\
         
         SimVLM\cite{wang2022simvlmsimplevisuallanguage} & 40.6  &33.7 &143.3 & 25.4\\
         \hline
        \end{tabular}
        \vspace{5pt}
        \caption{\label{tab-4}Experimental results on MSCOCO Image Captioning with cross entropy loss. The results are reported on the Karpathy test split.}
        \label{tab:evaluation_metrics}
        \end{table}
    }
       
    }
}
}

\newpage
\section{Multistyle Image Caption Generation, controllable captions, multilingual caption generation}
{
          \begin{figure}[H]
            \centering
            \includegraphics[width=0.8\linewidth]{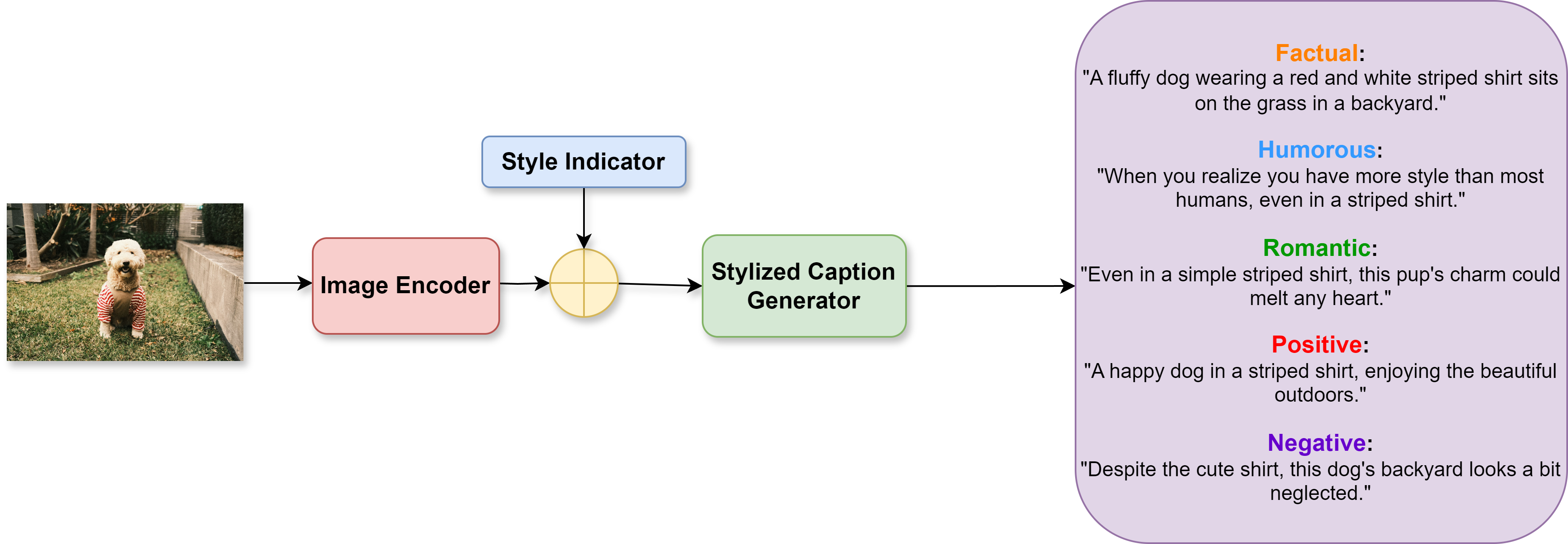}
            \caption{Overview of a multi-style captioning model.}
            \label{fig:multi-style captioning model}
        \end{figure}
        
    \textit{ \textbf{Note}: Here we address RQ1. (Which image captioning models achieve superior performance for distinct caption types, including those requiring semantic and spatial alignment and emotional expression?)}\\
    
    A challenge in image captioning involves generating captions or image descriptions which are not only factually correct, but also capture the abstract idea behind the image, in particular, they must also capture the sentiment that an image may have, making them more human preferable. To generate more human-like captions which capture the various sentiments inside an image or may have a particular style, various approaches have been developed.These approaches delve into understanding the sentiment behind each image,such as humor, striving to create captions that resonate with human perception.
    
    In the same direction, the authors in \cite{mathews2015senticapgeneratingimagedescriptions} introduce SentiCap, a novel approach to image captioning by integrating sentiment analysis into the traditional CNN+RNN architecture. This model, composed of two parallel RNNs, adeptly captures both positive and negative sentiments within image descriptions. While one RNN serves as a background language model, the other specializes in sentiment representation, allowing for nuanced and emotionally resonant captions. A unique word-level regularizer further enhances the model's performance by emphasizing sentiment words during training, effectively blending the outputs of both RNN streams. 
    
    Caption generation is looked upon as composed of two tasks, that is, factual description and other is describing the image with sentiments. A multi-modal switching RNN is used, it sequentially generates words in a sentence. For each word \( y_t \) in the caption a binary sentiment variable \( s_t \) is introduced, at time step \(t\) to indicate whether positive (1) or negative (0) sentiment mechanisms are employed. At each time step t, the model predicts the probability of both the word \( y_t \)  and the sentiment variable  \( s_t \) , given the image feature \( x\)  and the preceding words \( y_{t-1} \) .
    
    The traditional datasets such as COCO\cite{lin2014microsoft} and Flickr30K\cite{7410660}, while factual and informative, lack the ability to evoke human engagement due to their neutral tone.To address this limitation, the authors in \cite{shuster2019engagingimagecaptioningpersonality} introduce the PERSONALITY CAPTIONS dataset, comprising 241,858 captions, each tailored to one of 215 distinct personality traits. By conditioning image captions on controllable personality traits, the study aimed to infuse them with emotion and style, thereby enhancing human engagement. 
    Using text, personality, and image encoders, they project the image, caption, and personality in the same space  to create the TransResNet architecture for caption retrieval.The TransResNet retrieval architecture, attained the highest \emph{R@1 score} on the \emph{Flickr30k} dataset, underscoring the efficacy of their approach. However, challenges persisted, particularly in refining generative models to fully capitalize on the enriched dataset.
    
    The authors in \cite{8953861} address the task of \emph{Multi-Style Image Captioning} (MSCap) using an adversarial learning network.
    
    It seeks to develop a single model that can produce captions in different styles without the need for paired images.The standard image encoder is followed by four key modules that make up the framework: a style-dependent caption generator that generates a statement by using a style and an encoded image. A discriminator for captions that separates produced sentences from actual sentences. A Style classifier which identifies the style of the input sentence and a Back-translation module which ensures visually grounded stylized captions by enforcing cycle consistency between factual and stylized captions.Specifically, the back-translation module includes a text encoder that takes a generated stylized caption \(y\) and the target style label \(s\) as input,the output of the text encoder is fed into a text decoder which generates an output sentence which is expected to be consistent with the factual caption \(y_f\), which was used to generate \(y\).The intuition behind utilizing cycle consistency is to establish a feedback loop that reinforces the relationship between the factual and stylized captions. By iteratively translating between the two representations while maintaining consistency, the model learns to generate stylized captions that are both visually appealing and contextually relevant to the image content. The generator and discriminator are trained adversarially to produce more natural and human-like captions.
    
    In a similar vein the authors in\cite{Zhao_Wu_Zhang_2020} introduce \emph{MemCap}, which employs memory mechanisms to encode knowledge about linguistic styles, diverging from the prevalent reliance on language models. Its memory module stores embedding vectors representing phrases related to style, learned from the training corpus. To identify these phrases, MemCap employs a sentence decomposing algorithm that separates stylized sentences into parts reflecting style and content. During caption generation, the system extracts relevant style information from the memory module using attention mechanisms and integrates it into the language model.

    Most of the captioning techniques often employ black-box architectures, which lack transparency and controllability from an external standpoint.Due to the infinite potential variations in image descriptions depending on the specific objectives and contextual nuances, a higher degree of controllability becomes necessary for captioning systems to function well in complex settings.\cite{8954335}. Controllable captioning methodologies engage users in the image captioning process by soliciting their input to select and prioritize elements to be depicted. This user-provided guidance serves as a guiding agent for the caption generation process.\cite{stefanini2021tellsurveydeeplearningbased}
    
    Recent advancements in captioning methodologies, as outlined in\cite{8954335}, facilitate the generation of diverse captions by integrating both grounding and controllability mechanisms. By leveraging control signals in the form of sequential instructions or sets of image regions, these methodologies employ recurrent architectures to generate captions that explicitly correlate with designated regions. This approach adheres to the constraints imposed by the provided control parameters, thereby enhancing the precision and adaptability of the generated captions \cite{8954335}.
    
    Image captioning is typically done in English, but multilingual captioning tries to make it work in other languages too. There are two main ways to do this: one is to gather captions in various languages for popular datasets, like having Chinese and Japanese captions for COCO images or German captions for Flickr30K. The other way is to train systems that can handle multiple languages without needing paired captions in each language.\cite{stefanini2021tellsurveydeeplearningbased}

    Upon analyzing table ~\ref{tab-5}, we note that for positive style captions, the MSCap model achieves a higher CIDEr score on the SentiCap dataset, implying superior consensus and diversity in its generated captions. Conversely, the SentiCap model outperforms in METEOR score, indicating better fluency in its styled captions.
    
    For negative style captions SentiCap achieves both higher CIDEr and METEOR score. We examine perplexity and style classification accuracy for positive, negative, romantic, and humorous style captions. The Perplexity score measures how well a model predicts a sample. The lower the perplexity scores the better the performance, for the rest of the metrics the higher score is better. The style classification accuracy, refers to the accuracy achieved by a pre-trained style classifier in correctly identifying the style or category of a given caption. It is observed that MemCap demonstrates higher style classification accuracy compared to MSCap. This implies MemCap’s superior ability to correctly identify the intended style of captions.
    
    Additionally, MemCap achieves lower perplexity scores across these styles. Lower perplexity scores signify reduced uncertainty in the language model’s predictions, indicating higher fluency and coherence in the generated captions. Therefore, MemCap not only excels in accurately classifying different styles but also produces more fluent and coherent captions across various stylistic categories. To analyze the performance of various models in terms of semantic and spatial alignment we observe the SPICE scores in table ~\ref{tab-7}. It shows that SPICE score are higher for SimVLM and VinVL in comparison to other methods. 

}

\newpage

{
\begin{sidewaystable}
      
        \centering
        \resizebox{\textwidth}{!}{
         \begin{tabular}{ |m{2.5cm}|m{2cm}|m{2cm}|m{2cm}|m{2cm}|m{2cm}|m{2cm}|m{2cm}|m{2cm}|m{2cm}|m{2cm}|m{2cm}|m{2cm}|m{2.5cm}| } 
            \hline
            \multicolumn{1}{|c|}{} & \multicolumn{5}{|c|}{\textbf{Sentiment}} & \multicolumn{7}{|c|}{\textbf{Evaluation Metric}} \\
            \hline
            \textbf{Model} & \textbf{Positive} & \textbf{Negative} & \textbf{Romantic} & \textbf{Humorous} & \textbf{Personality\newline/Style} & \textbf{BLEU-4} & \textbf{CIDEr} & \textbf{SPICE} & \textbf{METEOR} & \textbf{ROGUE-L} & \textbf{Style classification accuracy (percent.)} & \textbf{Perplexity (lower is better)} & \textbf{Dataset} \\
            \hline
            SentiCap\cite{mathews2015senticapgeneratingimagedescriptions} & yes & - & - & - & - & 10.8 & 54.4 & - & 16.8 & 36.5 & - & - & SentiCap \\
            \hline
            SentiCap\cite{mathews2015senticapgeneratingimagedescriptions} & - & yes & - & - & - & 13.1 & 61.8 & - & 16.8 & 37.9 & - & - & SentiCap \\
            \hline
            Generative model (UPDOWN with image encoder - ResNeXt trained on 3.5 billion Instagram pictures)\cite{shuster2019engagingimagecaptioningpersonality} & - & - & - & - & Traits selected from 215 possible personality traits such as dramatic, anxious, optimistic. & 8.0 & 16.5 & 5.2 & - & 27.4 & - & - & PERSONALITY-CAPTIONS \\
            \hline
            MSCap\cite{8953861} & yes & - & - & - & - & - & 55.3 & - & 16.8 & - & 92.5 & 19.6 & MSCOCO, SentiCap, FlickrStyle10K \\
            \hline
            MSCap\cite{8953861} & - & yes & - & - & - & - & 51.6 & - & 16.2 & - & 93.4 & 19.2 & MSCOCO, SentiCap, FlickrStyle10K \\
            \hline
            MSCap\cite{8953861} & - & - & yes & - & - & - & 10.1 & - & 5.4 & - & 88.7 & 20.4 & M COCO, SentiCap, FlickrStyle10K \\
            \hline
            MSCap\cite{8953861} & - & - & - & yes & - & - & 15.2 & - & 5.3 & - & 91.3 & 22.7 & MSCOCO, SentiCap, FlickrStyle10K \\
            \hline
            MemCap\cite{Zhao_Wu_Zhang_2020} & yes & - & - & - & - & - & 52.8 & - & 16.6 & - & 96.1 & 18.1 & MSCOCO, SentiCap, FlickrStyle10K \\
            \hline
            MemCap\cite{Zhao_Wu_Zhang_2020} & - & yes & - & - & - & - & 59.4 & - & 15.7 & - & 98.9 & 18.9 & MSCOCO, SentiCap, FlickrStyle10K \\
            \hline
            MemCap\cite{Zhao_Wu_Zhang_2020} & - & - & yes & - & - & - & 19.7 & - & 7.7 & - & 91.7 & 19.7 & MSCOCO, SentiCap, FlickrStyle10K \\
            \hline
            MemCap\cite{Zhao_Wu_Zhang_2020} & - & - & - & yes & - & - & 18.5 & - & 7.2 & - & 97.1 & 17.0 & MSCOCO, SentiCap, FlickrStyle10K \\
            \hline
       
        \end{tabular}

        }

    \caption{\label{tab-5}Evaluation metrics of various multi-style image captioning models, based on four sentiments, positive, negative,romantic and humorous. MS COCO is used for factual captions, senticap for positive and negative sentiments and FlickerStyle 10k dataset is used for romantic and humorous sentiments.}
    \label{tab:multistyle}   
    \end{sidewaystable}

}

\newpage
\section{Unsupervised, Self-supervised and Reinforcement learning}
{   
    \begin{figure}[H]
            \centering
            \includegraphics[width=0.35\linewidth]{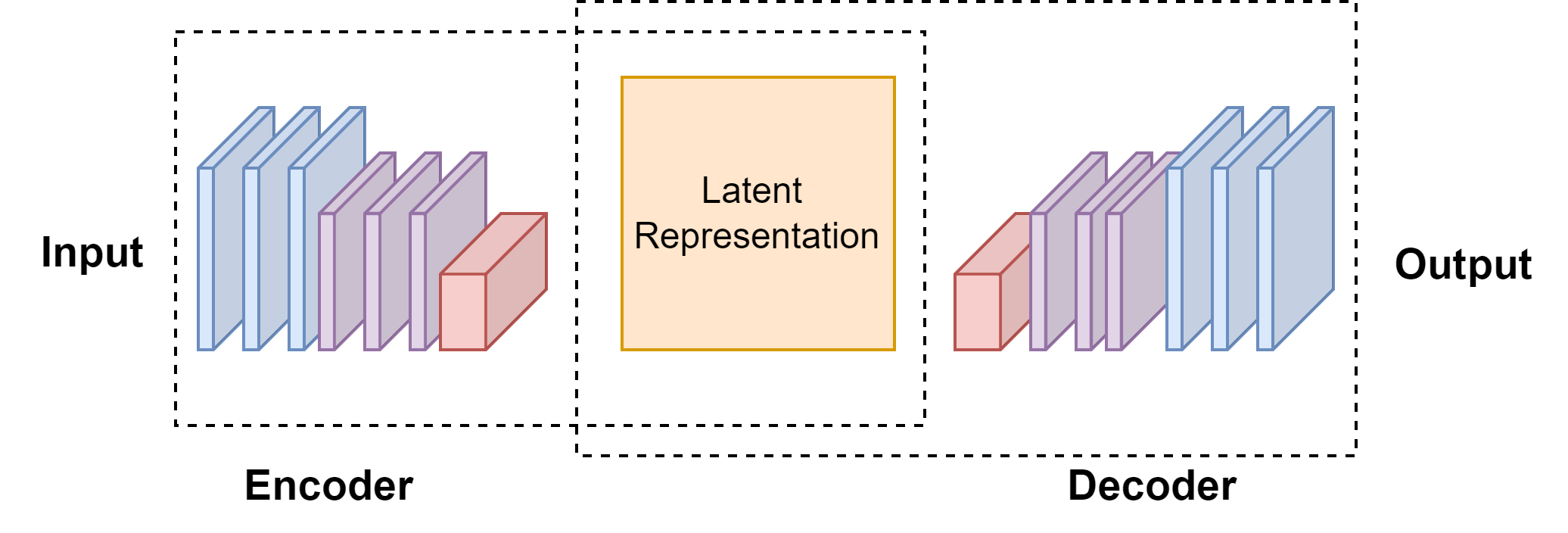}
            \caption{A feed-forward deterministic autoencoder network}
            \label{fig:autoencoder}
        \end{figure}
    \textit{ \textbf{Note}: Here we address RQ2. (What are the approaches in unsupervised,self-supervised and reinforcement learning?)}
    
    Many datasets consist primarily of unlabeled data and it is often impractical to accurately annotate huge quantities of unlabeled data. Unsupervised learning methods eliminate the dependency on high-quality image-caption pairs. Unsupervised and self-supervised learning techniques such as contrastive learning and adversarial learning are employed in image-captioning models.Many unsupervised learning based models use autoencoders and variational autoencoders  due to their ability to learn meaningful representations of data through an encoder-decoder architecture, where the encoder compresses input data into a latent space representation, and the decoder reconstructs the original input from this representation. While autoencoders aim for a close approximation of input data, variational autoencoders incorporate probabilistic principles, allowing them to learn structured latent representations by modeling the distribution of latent variables.
    
    An unsupervised learning method called an autoencoder (AE) trains a model to generate an output that closely resembles the input data. Applications including data dimensionality reduction, image classification, object identification, and image denoising frequently make use of AEs. Three parts usually make up an AE's architecture: a latent space, a decoder, and an encoder. In order to capture important features and reduce information loss, the encoder component of the network compresses the input data into a lower-dimensional latent space representation. The input data is represented succinctly and meaningfully by the latent space. After that, the decoder uses this compressed form to reassemble the original input data. But conventional AEs are unable to provide completely new data samples because they are not explicitly designed for this purpose.\cite{9171997}
    
    A \emph{Variational Autoencoder} (VAE), on the other hand, is a unique kind of autoencoder that integrates ideas from variational Bayes inference.\cite{ZELASZCZYK2023302} The goal of VAEs, which were first introduced by the authors of\cite{kingma2022autoencodingvariationalbayes} and later by the authors of \cite{doersch2021tutorialvariationalautoencoders}, is to discover the training data's underlying distribution. Through sampling from this learned distribution, VAEs are able to produce new samples of data. The main distinction between VAEs and AEs is that the latter are meant to learn to represent the probability distribution of the data, whereas AEs concentrate on learning the compressed representation of input data and reconstructing it. By sampling from the learnt distribution, VAEs can produce fresh data samples, which makes them more suited for generative tasks than regular AEs.\cite{9171997} Several cutting-edge methods for captioning images were developed which employ variational autoencoders to generate captions in an unsupervised manner. 
    
    We look at various approaches in unsupervised,self-supervised and reinforcement learning. In\cite{10.1145/3614435}, the authors present a novel structure for creating a variety of image captions that they name \emph{Conditional Variational Autoencoder} (DCL-CVAE). Contrastive learning and sequential variational autoencoder are integrated in this system. Unlike traditional image captioning models that rely on \emph{Maximum Likelihood Estimation} (MLE) to predict words based on their frequency in the training data, which often leads to biased outputs favoring common words and repetitive phrases, the DCL-CVAE framework aims to address this limitation by enhancing both diversity and accuracy simultaneously. \\
    
    During the encoding stage, consecutive latent spaces for caption pairs are learned by DCL-CVAE using conditional VAEs. It then applies triplet contrastive learning to enhance these latent representations' distinguishability for matched and mismatched image-caption pairs. As we proceed to the decoding phase, a novel objective is developed by combining cross-entropy loss and contrastive learning. Through global contrastive learning, this method not only reduces the generation of frequently used words but also greatly increases caption diversity.\\
    
    {    
        \begin{figure}[H]
            \centering
            \includegraphics[width=0.4\linewidth]{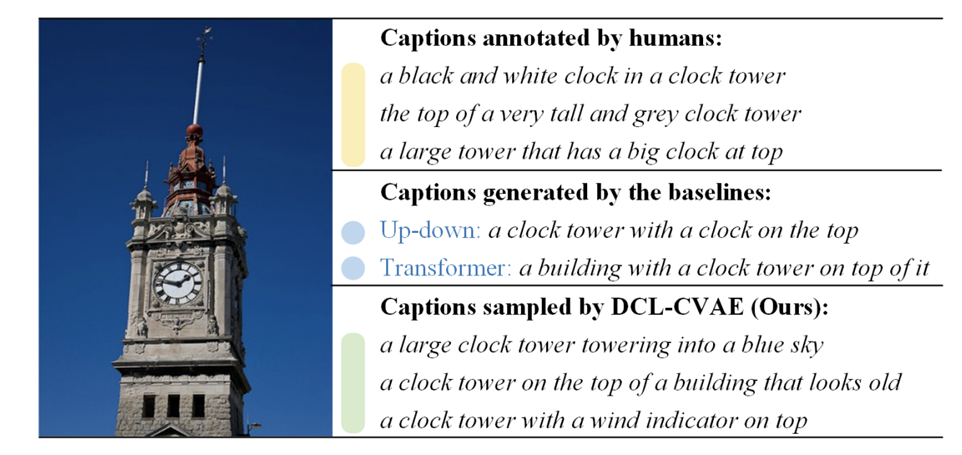}
            \caption{A qualitative comparison of DCL-CVAE with Up-Down and Transformer models in image captioning.\cite{10.1145/3614435}}
        \end{figure}
    }
    
    In \cite{wang2020visual}, another unsupervised learning approach which has shown state-of-the art results in image captioning along with other vision-language tasks is Visual Commonsense R-CNN, The training objective of VC R-CNN differs fundamentally from other methods: it employs causal intervention $(P(Y |do(X)))$ instead of conventional likelihood $(P(Y |X))$. This unique approach enables VC R-CNN to learn \emph{sense-making} knowledge, such as understanding that a chair can be sat on, rather than merely identifying common co-occurrences like the presence of a chair when a table is observed.In their experiment, the authors have shown that discarding the refining encoder from AoANet and using only VC-RCNN with AoANet achieves a new state-of-the-art performance.
    
    {
        \begin{figure}[H]
        \centering
        \includegraphics[width=0.7\linewidth]{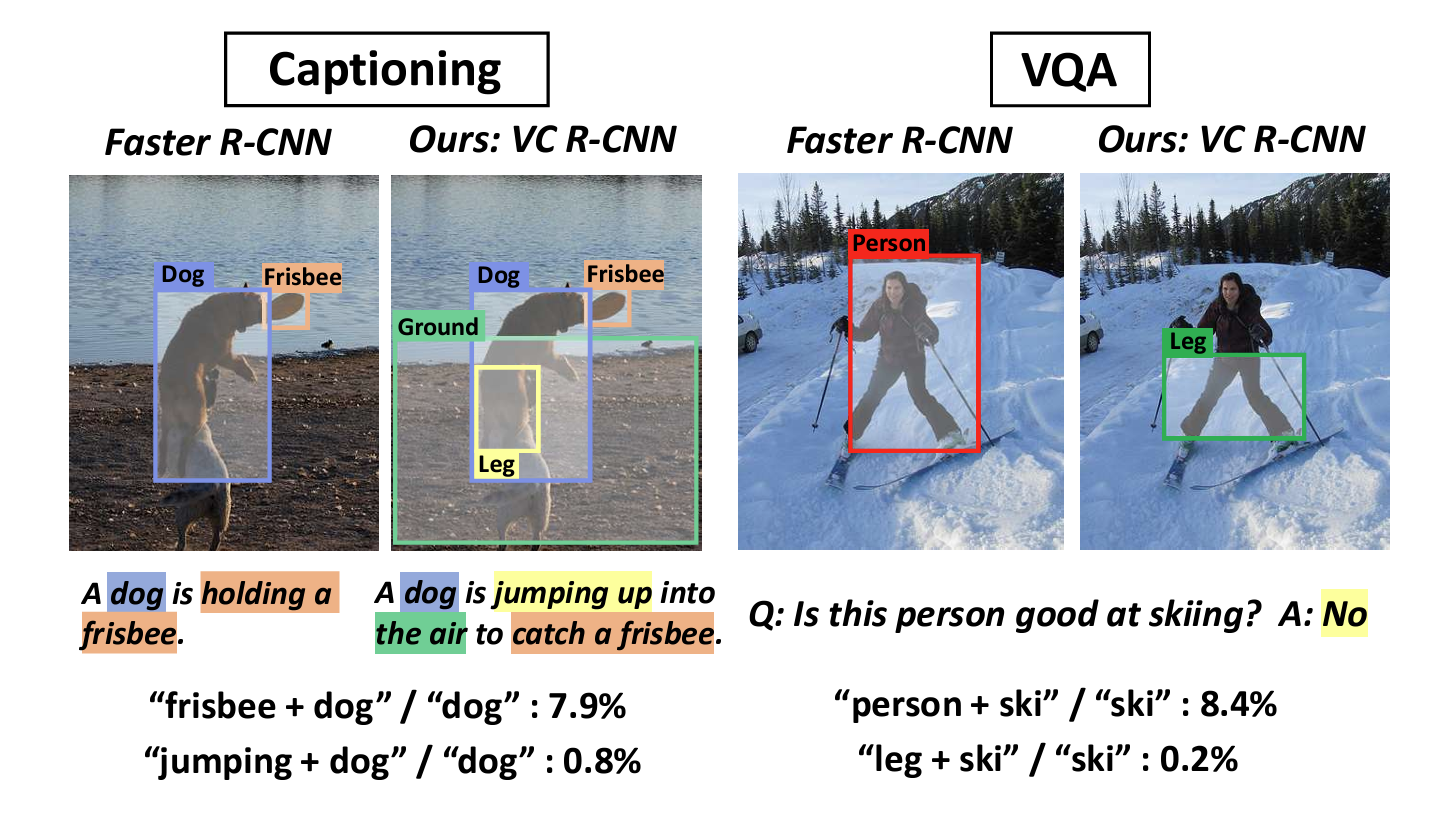}
        \caption{Qualitative comparison of VC RNN with Faster R-CNN in image captioning and visual question answering \cite{wang2020visual}}
        \end{figure}
    }
    
    The authors in \cite{wang2020visual} did a quantitative comparison of VC R-CNN features(denoted as only VC) with Faster R-CNN features(denoted as obj) for two models up-down and AOANet.
    The quantitative results highlight that both the model’s performance enhanced when the VC features were concatenated to the original feature.
    
    {                                          
        \begin{figure}[H]
        \centering
        \includegraphics[width=0.4\linewidth]{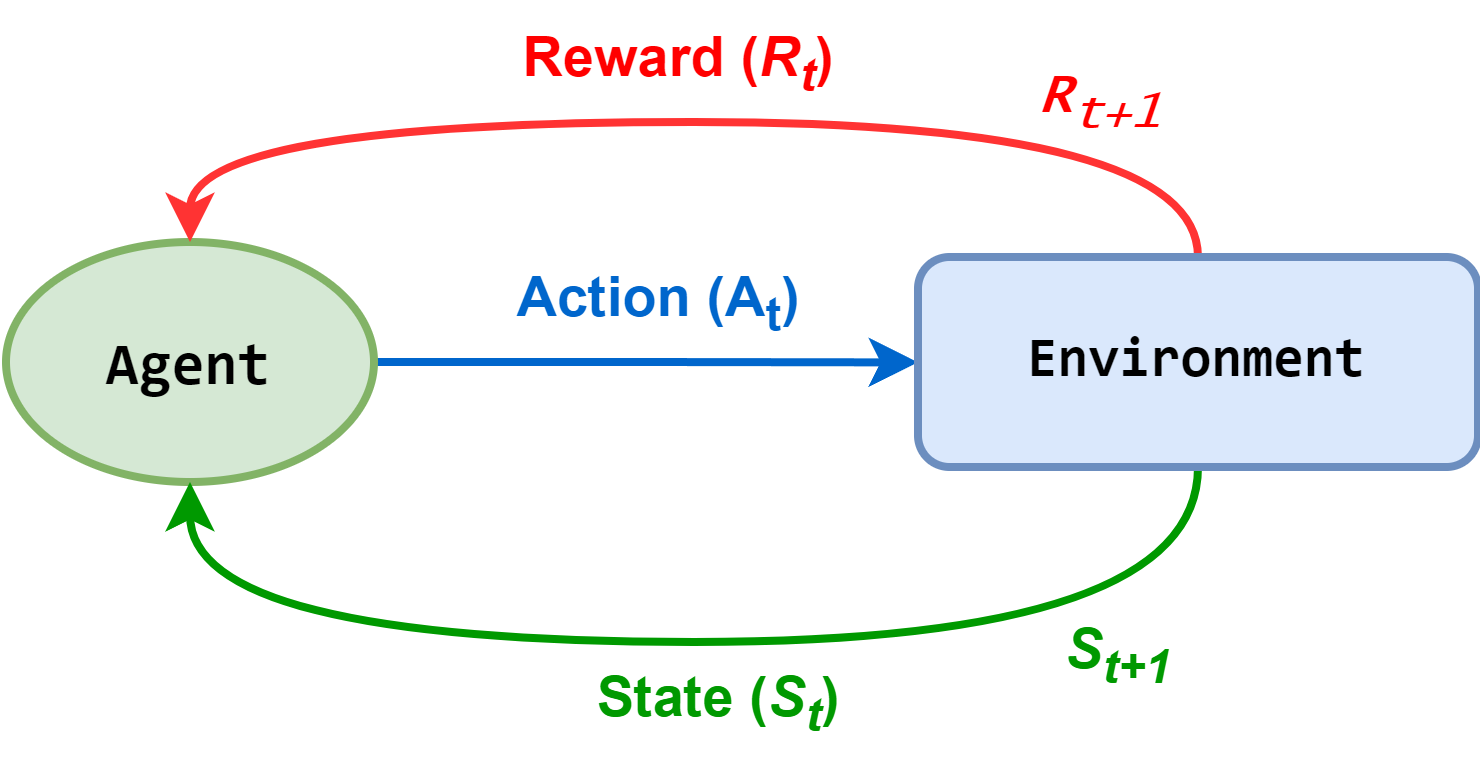}
        \caption{Reinforcement learning model}
        \end{figure}
    }

    Reinforcement learning is a machine learning paradigm where an agent learns to make sequential decisions in an environment to maximize a notion of cumulative reward. The agent interacts with the environment by taking actions, and based on these actions, it receives feedback in the form of rewards or penalties. The goal of the agent is to learn a policy—a mapping from states to actions—that guides its decision-making process to achieve long-term objectives.
    
    Various models leverage reinforcement learning for image captioning to improve the quality and coherence of generated captions. Within this framework, the authors in\cite{wang2018show} examine the difficulty of crafting narrative paragraphs for a series of images, which is more demanding than generating captions for individual images because it requires well-organized narrative and emotive language. The difficulty of this task is increased by the limited training data. To overcome these obstacles, the authors suggest a sequence-to-sequence modeling strategy that is strengthened by adversarial training and reinforcement learning. As a story generator, they provide a hierarchical recurrent neural network that has been improved through reinforcement learning and rewards. \\
    
    In order to incentivize the creation of pertinent and narrative-style paragraphs, rewards are engineered with two critic networks: a multi-modal critic and a language-style discriminator. Both the narrative generator and the reward critics are viewed as adversaries; the generator seeks to produce paragraphs that are identical to human-like stories, while the critics improve the generator through policy gradient. 
}
   
   {                                           
        \begin{figure}[H]
        \centering
        \includegraphics[width=0.8\linewidth]{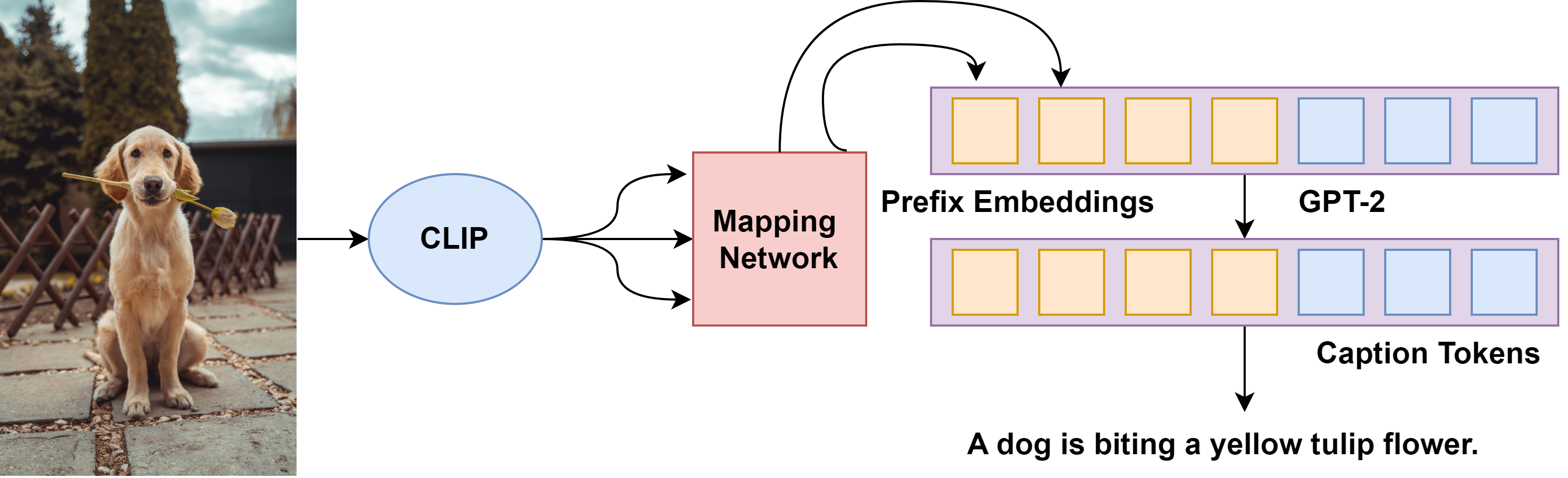}
        \caption{Overview of CLIPCap architecture \cite{mokady2021clipcapclipprefiximage}}
        \end{figure}                                  
    }

{                                                  
    Researchers have also explored the self-supervised learning based models for image captioning, which have shown state-of-the-art performance. \\
    
    Self-supervised learning models for image captioning learn to make predictions about its input data without explicit supervision. Instead of relying on labeled data provided by humans, self-supervised learning algorithms generate their own supervision signal from the input data itself. \emph{Contrastive Language-Image Pre-training} (CLIP)\cite{jin2023selfsupervisedimagecaptioningclip}, is one such self-supervised learning approach which aims to understand the content and context of images in a way that facilitates generating relevant and coherent captions. CLIP operates by embedding the image and text pairs in a shared semantic space where related pairs (like an image and its description) are brought close together, while unrelated pairs are pushed apart. This is achieved through contrastive learning, a type of self-supervised learning that doesn't require human-labeled data but instead generates labels from the input itself. By comparing positive pairs with negative examples, CLIP learns to infer textual representations from unseen images, leveraging a large dataset for learning signals. However, CLIP has limitations, including sensitivity to image quality, difficulty with abstract images, and challenges in handling ambiguity.\cite{wang2024surveylargelanguagemodels}. \\
   
    Various models have employed the CLIP framework which underlines the utility of CLIP in image-captioning task. \cite{cho2023finegrainedimagecaptioningclip, jin2023selfsupervisedimagecaptioningclip}, CLIPCap\cite{mokady2021clipcapclipprefiximage} model utilizes CLIP encoding to initiate the image captioning process. It employs a mapping network to integrate CLIP's semantic features into the caption generation pipeline. Subsequent fine-tuning with a language model like GPT-2\cite{radford2019language} refines these initial representations, yielding informative and contextually relevant captions. Leveraging CLIP's pre-existing understanding of visual and textual data, the model efficiently generates captions in a self-supervised manner. By training only the mapping network while keeping CLIP and the language model frozen, the architecture achieves lighter, faster, and more resource-efficient performance while remaining competitive on benchmark datasets such as Conceptual Captions\cite{sharma2018conceptual} and nocaps\cite{9009481}. \\
    In \cite{cho2023finegrainedimagecaptioningclip} the authors have used the similarity scores generated by CLIP as rewards to provide feedback to the captioning model. By using the similarity scores as rewards, the captioning model can learn to generate captions that are more aligned with the content of the images. \\

}

\section{Application: Medical Image Captioning}
{
    If we look at the application domains of image captioning problem, most significant out of these is the medical or health care domain. Use of artificial intelligence in generating descriptive reports by looking at the medical images of various diseases is a crucial task which can be revolutionized by emerging technologies in the vision-language domain. Medical image captioning task is different from regular image captioning in the way that it highlights the relationship between image and clinical findings, also there are various medical image modalities such as radiography, MRI, ultrasound, computed tomography which have to be taken into consideration for generating detailed automated clinical reports. Medical image captioning can be used for diagnosis, treatment and report generation. For example, radiology report generation requires extensive expertise, hence reliable automatic report generation is much desirable.There are certain bottlenecks in the medical image captioning task, first being the complexity of the task itself, since generating descriptive clinical reports from medical images is a challenging task, second is open access datasets with paired medical images and reports is sparse. Medical reports are descriptive in nature which are longer than typical image captions and they must be precise, reliable and must cover heterogeneous information, which may be difficult in case of rare diseases.
    
    Various image captioning methods have been employed for the task of medical image captioning. Most recent vision-language pre-training paradigm has also been applied to the medical image captioning task. 
    
    One of the main challenges in generating automatic medical reports from medical images is to detect abnormalities, since the training corpora has sparser abnormal cases. Producing thorough medical reports is a difficult task due to the length of medical reports. In order to resolve the first task of detecting patients’ abnormalities \emph{Global label Pooling} (GLP) is proposed\cite{8970668}. Detecting several abnormalities is taken up as a multi-label classification problem. A CNN is used to generate feature maps and GLP mechanism is used to predict abnormality heat maps from feature maps.

    In Global label pooling, a DenseNet except the last global pooling layer and fully connected layer is used and a feature map is obtained, using this feature map a label map is obtained using a convolutional layer. The label map indicates the heat maps for all the abnormalities.

    After detecting all possible abnormalities a hierarchical RNN is used for caption prediction, viz-a-viz; a sentence RNN and word RNN.The sentence RNN responsible for generating sentences from image features determines the number of sentences needed for the resulting paragraph. It also generates a topic vector for each sentence. Then, using this topic vector, word RNN,predicts the words for each sentence individually. The outputs of this word-level RNN are then combined together to form the final paragraph.
    
     \begin{figure}[H]
        \centering
        \includegraphics[width=0.9\linewidth]{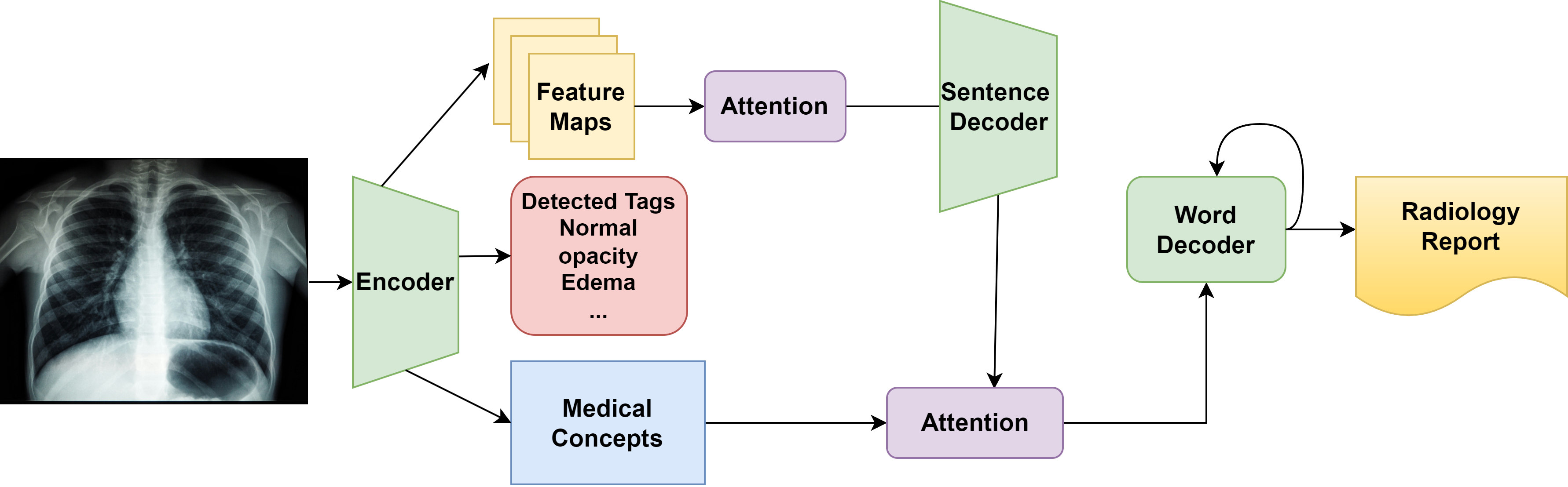}
        \caption{A typical architecture of medical image captioning models}
        \label{fig:medical}
    \end{figure}
    
    Generating radiology reports is an intensive and time-consuming task, automatic image captioning is highly desirable in this domain, but generating detailed reports which highlight the linkage of a particular diagnosis with a region in the image is a challenging task, and it is made even more difficult by the absence of publicly available databases with paired medical images and reports, in order to tackle these challenges a generative encoder-decoder architecture was proposed in \cite{yuan2019automaticradiologyreportgeneration}. Their work mostly focuses on chest x-ray images. By pre-training their image encoder on a large volume of chest x-ray images, the issue of a lack of datasets for this task was overcome, allowing it to correctly recognize 14 typical radiographic observations. The image encoder was fed with multiple views of an image as a single input and in order to enforce consistency in predictions across different views Mean squared loss error was used.In addition to that, they also extracted most recurrent medical concepts from the radiology reports in the training data and used a concept classifier on top of image encoder.
    
    Two hierarchical decoders were used in order to generate reports sentence by sentence.The two layers are composed of a word LSTM decoder that produces natural language output from sentence hidden states as input and a sentence LSTM decoder that generates sentence hidden states. Three different attention mechanisms are proposed to fuse multiview image features; viz-a viz;concatenation of multiview image features, early fusion of frontal and lateral global features with attention and late fusion of frontal and lateral global features with attention.
    
    Finally the word LSTM decoder incorporates the medical concepts along with the output from LSTM sentence decoder and generates natural language output.\cite{yuan2019automaticradiologyreportgeneration}

    In another attempt to enhance medical image captioning the authors in\cite{selivanov2022medicalimagecaptioninggenerative} used the combined capability of seminal show,attend and tell (SAT)\cite{10.5555/3045118.3045336} model and generative pre-trained transformer\cite{NEURIPS2020_1457c0d6} (GPT-3), the SAT model is used to identify the pathologies in the input images, with the help of an attention module, it creates a report based on the information found on x-ray. The GPT-3 model uses the SAT model's output as a seed. Using the MIMIC-CXR dataset\cite{johnson2019mimiccxrjpglargepubliclyavailable}, the refined GPT-3\cite{NEURIPS2020_1457c0d6} model produces a thorough report based on the pathologies identified by the SAT model. 
    
    Vision language pre-training paradigm has also been employed in the domain of medical image captioning. Researchers in \cite{9894658} propose Medical Vision Language Learner (MedViLL). In this model a  joint representation is learned from  an image and its report and various vision-language understanding and generation downstream tasks are performed with fine tuning.After extracting visual information from the last convolution layer of the images, they flatten them along the spatial dimension using a CNN. BERT \cite{devlin2019bertpretrainingdeepbidirectional}is used for textual encoding and language feature embeddings are obtained, these two embeddings are then concatenated to input into the joint embedding component.Joint embedding component generates contextualized embeddings. \cite{9894658}

   \textit{\textbf{ Note}:In sections 10 and 11 we attempt to answer RQ3. What are the primary datasets utilized for image captioning tasks, what evaluation metrics are commonly employed, how do these metrics assess caption quality, and what are the distinct characteristics of these datasets?} 

}

\section{Evaluation Metric}
{
    \label{eval-met}\textbf{BLEU} evaluates the machine-generated translation's quality.It offers a score depending on how similar a machine-generated translation is to one or more reference human translations.When comparing the machine-generated translation to the reference translations, BLEU calculates how accurate the \textit{n-grams} are. Precision is calculated for various values of $n$, usually up to 4. BLEU adjusts the precision score to account for situations in which the machine-generated translation may produce more \textit{n-grams} than the reference translations. Each \textit{n-gram's} count is capped at the most times it can appear in a single reference translation.
    
    By comparing the lengths of the machine-generated and reference translations, it penalizes translations that are too short. This incentivizes the system to produce translations with lengths comparable to those of the reference translations. The final BLEU score is calculated by adding the updated precision scores for each \textit{n-gram} and applying the brevity penalty. A perfect match between the machine-generated translation and the reference translations is indicated by a score of 1, which runs from 0 to 1. A notable drawback of BLEU is that it may not adequately account for variations in word choice, synonyms, or paraphrasing that could affect the overall quality of the translation\cite{callison-burch-etal-2006-evaluating, 10.3115/1073083.1073135}. We first compute the adjusted n-gram precision score on the test corpus before calculating the BLEU score\cite{9849164}.
    
    We calculate the precision score based on the number of matched \textit{n-grams} for each sentence in the test corpus. This entails calculating how many times each \textit{n-gram} appears in the candidate translation and the reference translation. But we \emph{clip} the counts—that is, we take the lowest count of each \textit{n-gram} in both the reference translation and the candidate translation—in order to prevent overestimating precision because of \textit{n-gram} repetition. These clipped numbers are added together for every sentence in the test corpus. We divide the sum of all candidate \textit{n-grams} in all test corpus sentences by the sum of the clipped counts that were calculated.

    {
        Consider the modified precision score \( P_n(X,Y) \) for candidate translations \( X \) and reference translations \( Y \), given by:
        
        \[ P_n(X,Y) = \frac{{\sum_i \sum_j \min \left\{ a_j(x_i), \max_{i \leq M} b_j(y_{ij}) \right\}}}{{\sum_i \sum_j \min \{ a_j(x_i) \}}} \]
        
        Where:
        \begin{itemize}
          \item \( x_i \) denotes the \( i^{th} \) candidate translation.
          \item \( y_{ij} \) denotes the \( j^{th} \) reference translation for the \( i^{th} \) candidate translation.
          \item \( a_j(\cdot) \) represents the count of \( j \)-grams in candidate translations.
          \item \( b_j(\cdot) \) represents the count of \( j \)-grams in reference translations.
          \item \( M \) denotes the maximum number of reference translations.
        \end{itemize}
        
        This formula computes the modified precision score, taking into account the minimum count between candidate \( j \)-grams and the maximum count of reference \( j \)-grams, divided by the minimum count of candidate \( j \)-grams.
    
        Next, we calculate the \emph{Brevity Penalty} (BP) which addresses the issue of penalizing candidate translations that are smaller than the length of reference translations, as follows;
    
        \[
        \text{BP} = 
        \begin{cases} 
        1 & \text{if } c_l > l \\
        e^{(1-\frac{l}{c_l})} & \text{if } c_l \leq l 
        \end{cases}
        \]
        
        \text{Where:}
        \begin{itemize}
            \item \( \text{BP} \) denotes the Brevity Penalty.
            \item \( c_l \) represents the length of the candidate translation.
            \item \( l \) denotes the length of the reference corpus.
        \end{itemize}
        
        \vspace{50pt}
        
        BLEU score is computed as,
        
        \[
        \text{BLEU} = \text{BP} \times \exp \left( \sum_{n=1}^{N} \alpha_n \log p_n \right)
        \]
        
        \text{Where:}
        \begin{itemize}
            \item BLEU represents the BLEU score.
            \item BP denotes the Brevity Penalty.
            \item \( p_n \) represents the modified \( n \)-gram precision.
            \item \( \alpha_n \) denotes the positive weight for \( n \)-grams, summing to one.
            \item \( N \) represents the maximum \( n \)-gram length considered in the computation.
        \end{itemize}
        
    }

    \textbf{METEOR} is predicated on the idea of unigram matching, which matches machine translation with reference human translations. METEOR matches words which originate from the identical stem as well as synonyms. If more than one reference  translations are found, the best score is chosen after the translation scores are individually calculated against each reference.  \cite{10.5555/1626355.1626389,9849164}.

    Penalty Term and the harmonic mean of Precision and Recall (\(F_{\text{mean}}\)) can be represented as

    {
        \begin{align*}
            \text{Penalty} &= \beta \cdot \left(\frac{\text{No.~of~Chunks}}{\text{No.~of~unigrams\_matched}}\right)^\theta &
            F_{\text{mean}} &= \frac{P_m \cdot R_m \cdot \alpha}{P_m \cdot \alpha + (1 - \alpha) \cdot R_m}
        \end{align*}
        
        \begin{align*}
            \text{Precision ($P_m$)}: \quad & P_m = \frac{|m|}{\sum_{k} g_k(c_i)} &
            \text{Recall ($R_m$)}: \quad & R_m = \frac{|m|}{\sum_{k} g_k(s_{ij})}
        \end{align*}

        \begin{flushleft}
        \text{Where:} \(\beta, \theta, \alpha\) \text{ are the hyperparameters set based on several datasets and } $|m|$ \text{ represents the number of matched word pairs.}
        \end{flushleft}
        
        $\sum_{k} g_k(c_i)$ \text{ is the sum of all hypothesis words} and $\sum_{k} g_k(s_{ij})$ \text{ is the sum of all reference words.} \\
        
        \[
        \text{METEOR} = (1 - \text{Penalty}) \cdot F_{\text{mean}}
        \]
    }
    
    The number of unigrams in the system translation that correspond to the reference translation's unigram count is referred to as \textit{unigrams mapped}.
    
    Chunks: The smallest quantity of chunks that the matching unigrams can be grouped into. Every chunk is made up of neighboring system translation unigrams that are also mapped to neighboring reference translation unigrams. \\

    \textbf{ROGUE} is a set of metrics used to evaluate the accuracy of machine-generated summaries. The metrics tally how many overlapping units—like word pairs, n-grams, and sequences of words—there are between the optimal human-written summaries and the computer-generated summary that needs to be assessed. The authors found that \textit{ROGUE-S, ROUGE-L, ROUGE-W, and ROUGE-2} performed well in tasks involving the summarizing of a single document.\cite{lin2004rouge, 9849164}

    {
        Rouge-L defines the longest common sub-sequence (LCS) as the longest identical fragment in the generated and reference sentences. 
        As examples, we consider the generated sentence $G$ and the reference sentence $R$.

        
        

        \begin{align*}
            R_{lcs} &= \frac{{LCS}(G,R)}{l_1} &
            P_{lcs} &= \frac{{LCS}(G,R)}{l_2} &
            ROUGE-L &= \frac{(1 + \beta^2) \cdot R_{lcs} \cdot P_{lcs}}{R_{lcs} + \beta^2 \cdot P_{lcs}}
        \end{align*}
        
        where $l_1$ and $l_2$ represent the lengths of $G$ and $R$ respectively, and $\beta = \frac{P_{lcs}}{R_{lcs}}$. 
        The only component to be taken into account in real-world applications is, $\beta \rightarrow \infty, \text{ } R_{lcs}$. The precision without the $F_{mean}$ function is the main focus of ROUGE.
    }

    \textbf{CIDER} evaluates the quality of image descriptions using human consensus. How frequently the \textit{n-grams} found in candidate sentences are found in reference sentences is indicated by a consensus measure.\textit{n-grams} that are absent from the reference sentences but present in the candidate sentence may indicate divergence or inconsistency, hence they are penalized thereby leading to lower consensus score.Additionally, to account for the overall context of the dataset, common \textit{n-grams} that occur frequently across all images are given lower weight in the consensus measure. These ubiquitous \textit{n-grams} are likely to be less informative as they may represent generic language patterns rather than specific image-related content. \cite{vedantam2015ciderconsensusbasedimagedescription,9849164}

    {
        The average cosine similarity between the candidate phrase $a_i$ and the reference sentences $r_{ij}$ is used to calculate the $CIDEr_{n}$ score for \textit{n-grams} of length $n$. This score takes precision and recall into account. 

        \[
        \text{CIDEr}_{n}(a_i, r_{ij}) = \frac{1}{m} \sum_{j} \frac{h_{n}(a_i) \cdot h_{n}(s_{ij})}{\|h_{n}(a_i)\| \|h_{n}(r_{ij})\|},\\
        \]
        
        where $\|h_{n}(a_i)\|$ is the magnitude of the vector $h_{n}(a_i)$, and $h_{n}(r_{ij})$ is a vector produced by $h_{k}(a_i)$ corresponding to all \textit{n-grams} of length $n$.
    }

    \textbf{SPICE} (Semantic Propositional Image Caption Evaluation) is a noteworthy development in the field of automated caption evaluation metrics. Unlike existing metrics, which primarily focus on n-gram overlap and may not fully capture the nuances of human judgment, SPICE introduces a novel approach by incorporating semantic propositional content. By analyzing scene graphs, SPICE assesses the semantic richness and coherence of model-generated captions, aligning more closely with human evaluation criteria.Moreover, SPICE's unique capability to address specific questions related to color comprehension and numerical counting sets it apart, making it a valuable tool for evaluating the proficiency of caption-generating models in understanding diverse aspects of visual content.\cite{anderson2016spice, 9849164}

    {
        For $SPICE$, a caption $Q$ is initially processed into a scene graph:

        \[
        H(Q) = \langle O(Q), E(Q), K(Q) \rangle,
        \]
        
        where $M$ denotes a set of object classes and $O(Q) \subseteq M$ is the set of objects referenced in $Q$,
        
        The set of edges representing the relationships between the objects is $E(Q) \subseteq O(Q) \times R \times O(Q)$.
        
        A set of relation types is represented by $R$, and $K(Q) \subseteq O(Q) \times A$ is the collection of properties linked to the object.
        
        $A$ denotes attribute type sets.
        
        Next, the function $T$ produces a logical semantic tuple built as follows to assess how similar the candidate scene graph is to the reference one:
        
        \[
        T(H(S)) \triangleq O(Q) \cup E(Q) \cup K(Q)
        \]
        
        The binary matching operator $\otimes$ is used as a function to return matching tuples. Finally, the F-values of objects, attributes, and relations in the sentence are calculated to evaluate the accuracy:
        
        \[
        P(Q,S) = \frac{|T(H(Q)) \otimes T(H(S))|}{|T(H(Q))|}, \\
        \]
        \[
        R(Q,S) = \frac{|T(H(Q)) \otimes T(H(S))|}{|T(H(S))|}, \\
        \]
        
        \[
        SPICE(Q,S) = F_1(Q,S) = \frac{2 \cdot P(Q,S) \cdot R(Q,S)}{P(Q,S) + R(Q,S)},
        \]
        
        where $P(Q,S)$ and $R(Q,S)$ are precision and recall respectively.
    }

}

\section{Datasets}
{
    \textbf{MS COCO Dataset} - widely recognized as one of the most prominent benchmarks for image captioning tasks. It contains a diverse collection of images, each annotated with multiple human-generated captions, capturing various objects, scenes, and activities in real-world contexts. With 1.5 million annotated object instances across 80 distinct categories and 91 labeled stuff categories, MS COCO provides a comprehensive and diverse representation of visual content in real-world contexts. Each image is enriched with multiple annotations, including object segmentation masks, facilitating precise localization and recognition of objects within complex scenes.\cite{lin2014microsoft}

    \textbf{Flickr30K Entities Dataset} -  Flickr30k Entities, an extension of the original Flickr30k dataset, enhances its utility by augmenting the existing 158,000 captions with 244,000 coreference chains. These chains create relationships between references to the same items in several descriptions for the same image, further enriched with 276,000 manually annotated bounding boxes. These annotations play a crucial role in advancing the field of automatic image description and grounded language understanding by facilitating tasks such as entity localization within images.\cite{7410660}
    
    \textbf{Visual Genome Dataset} - In contrast to traditional datasets focused on perceptual tasks like image classification, Visual Genome emphasizes cognitive tasks such as describing image and answering questions. With the help of the deep annotations of objects, properties, and relationships provided by this dataset, models are better able to comprehend and make sense of the visual world. With over 108,000 images, each containing a wealth of object instances, attributes, and pairwise relationships, one of the densest and largest databases of image annotations and descriptions is provided by Visual Genome.\cite{krishna2017visual}
     
    \textbf{TextCaps} - The TextCaps dataset addresses the need for image descriptions that encompass both visual content and accompanying written text, a crucial aid for visually impaired individuals in understanding their surroundings. Unlike previous approaches which primarily focus on automatically describing images and recognizing text separately, TextCaps integrates both aspects, presenting a novel challenge for models to comprehend text within the context of an image.With 145k captions corresponding to 28k images, this dataset requires models to not only recognize text but also to establish its relationship with visual elements and determine the pertinent information to include in the description. This task necessitates spatial, semantic, and visual reasoning, as models must navigate multiple text tokens and visual entities such as objects.\cite{sidorov2020textcaps}
    
    \textbf{VizWiz - Captions} - Comprising over 39,000 images sourced from visually impaired individuals, each paired with five captions, VizWiz-Captions offers a rich and diverse array of visual content. \cite{10.1007/978-3-030-58520-4_25}
    
    \textbf{Nocaps} - The 'nocaps' benchmark includes 166,100 human-generated captions for 15,100 images from the OpenImages validation and test sets. It offers a strong basis for testing image captioning models on a wide range of visual concepts. By combining COCO image-caption pairs with OpenImages\cite{openimages} image-level labels and object bounding boxes in the training dataset, 'nocaps' allows models to learn from diverse visual data, bridging the gap between training and real-world use. The inclusion of  nearly 400 object classes in test images without enough training captions highlights the novelty and complexity of this benchmark.\cite{9009481}

     \begin{table}[h!]
            \centering
             \begin{adjustbox}{max width=\textwidth}
            \begin{tabular}{|m{3cm}|m{2cm}|m{2cm}|m{4cm}|m{2cm}|}
                \hline
                \textbf{Dataset} & \textbf{Images} & \textbf{Captions per Image} &\textbf{Characteristics}& \textbf{Object \newline Categories}\\
                \hline
                MS COCO & 328,000 & 5 & Everyday Scenes & 91\\
                \hline
                Flickr30K Entities & 31,783 & 5 & Contains region-to-phrase correspondences & 44,518 \\
                \hline
               Visual Genome & 108,077 & 50 & Everyday scenes with objects, attributes and relationships & 33,877 \\
                \hline
                TextCaps & 28,408 & 5 & Consists of images with text present in them & - \\
                \hline
               VizWiz-Captions & 39,181 & 5 & Consists of captions of images taken by visually-impaired people & - \\
                \hline
               Nocaps & 15,100 & 10 & Dataset for novel object captioning & ~500\\
                \hline
              
            \end{tabular}
            \end{adjustbox} 
            \vspace{+5pt}
            \caption{Summarization of popular datasets in image captioning along with their characteristics}
            \label{tab:datasets}
        \end{table}

}

\section{Future Research Trends}
{
    Image captioning models have evolved over the past years from earlier vanilla encoder-decoder architectures to more sophisticated and efficient vision-language models which are capable of various vision-language tasks including image captioning. Due to the high efficacy of vision-language models their application is highly desirable in the medical domain to generate comprehensive reports from medical images. Since medical images are derived from diverse image modalities such as MRI scans, X-rays vision-language models are particularly useful in this scenario. But vision-language models are not easier to train and their latent representation is complex. Hence future research trends will be towards developing medical vision-language models which are easier to train \cite{10.1007/978-3-031-43907-0_61}.
    
    Vision-language models require fine-tuning on sizable downstream datasets, in the medical domain where data is scarce, it hinders their capacity to accurately comprehend and generate insights from medical images and accompanying clinical text\cite{moor2023med}. Few-shot and zero-shot models are required to tackle the problem of lack of data and to perform better on unseen inputs. Sophisticated unsupervised and semi-supervised learning methodologies can also be harnessed for mitigating the challenge of high-quality labeled data for image captioning. Image captioning of sensitive data such as medical data, where privacy is a main concern, can benefit from the integration of federated learning approaches in medical vision-language models \cite{LU2023107037}.
    
}

\section{Bibliometric Analysis}
{
    We perform a brief bibliometric analysis of the papers we selected from various sources. In Fig \ref{fig:years}, we display the number of papers selected from each year from 2019 to 2024 for our survey. Fig \ref{fig:sources} depicts a pie chart of the percentage of papers selected from various sources, the top three sources are : proceedings of the IEEE computer society conference on computer vision and pattern recognition, IEEE transaction on Pattern analysis and machine learning and Proceedings of IEEE conference on computer vision. Fig \ref{fig:keywords} depicts the co-authorship network visualization.Fig \ref{fig:territories} and \ref{fig:types} show documents count by country and percentage of document types.

    {
        \begin{figure}[H]
            \centering
            \begin{subfigure}[b]{0.45\textwidth}
                \centering
                \includegraphics[width=\textwidth]{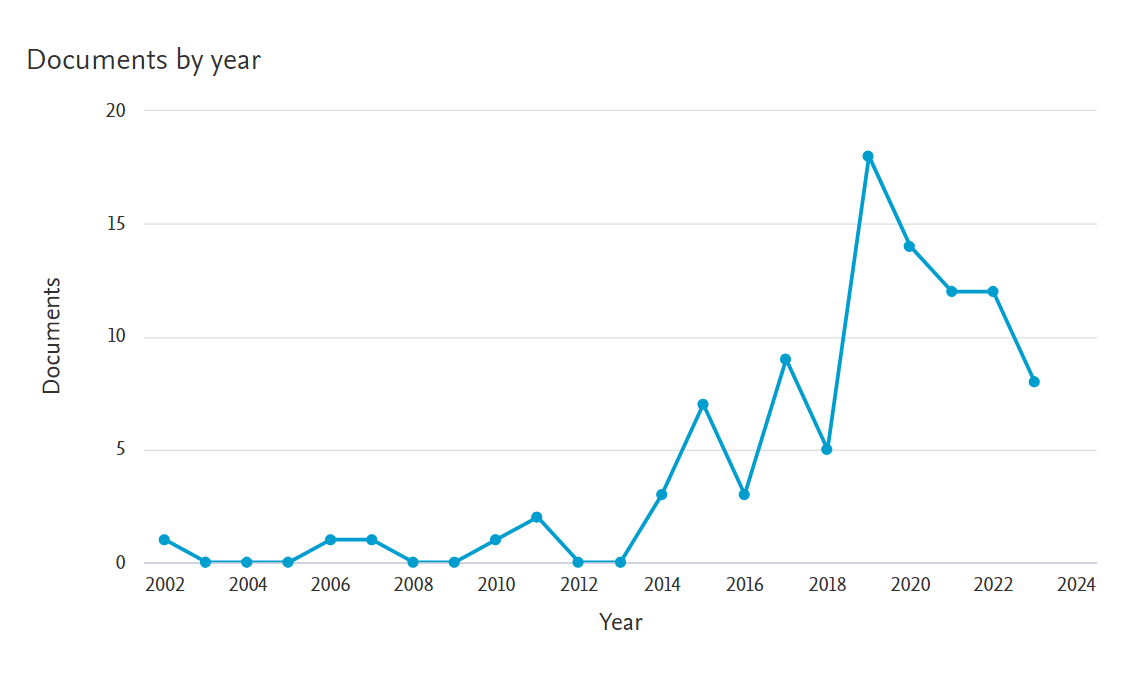}
                \caption{Number of publications by years}
                \label{fig:years}
            \end{subfigure}
            \hspace{0.05\textwidth} 
            \begin{subfigure}[b]{0.45\textwidth}
                \centering
                \includegraphics[width=\textwidth]{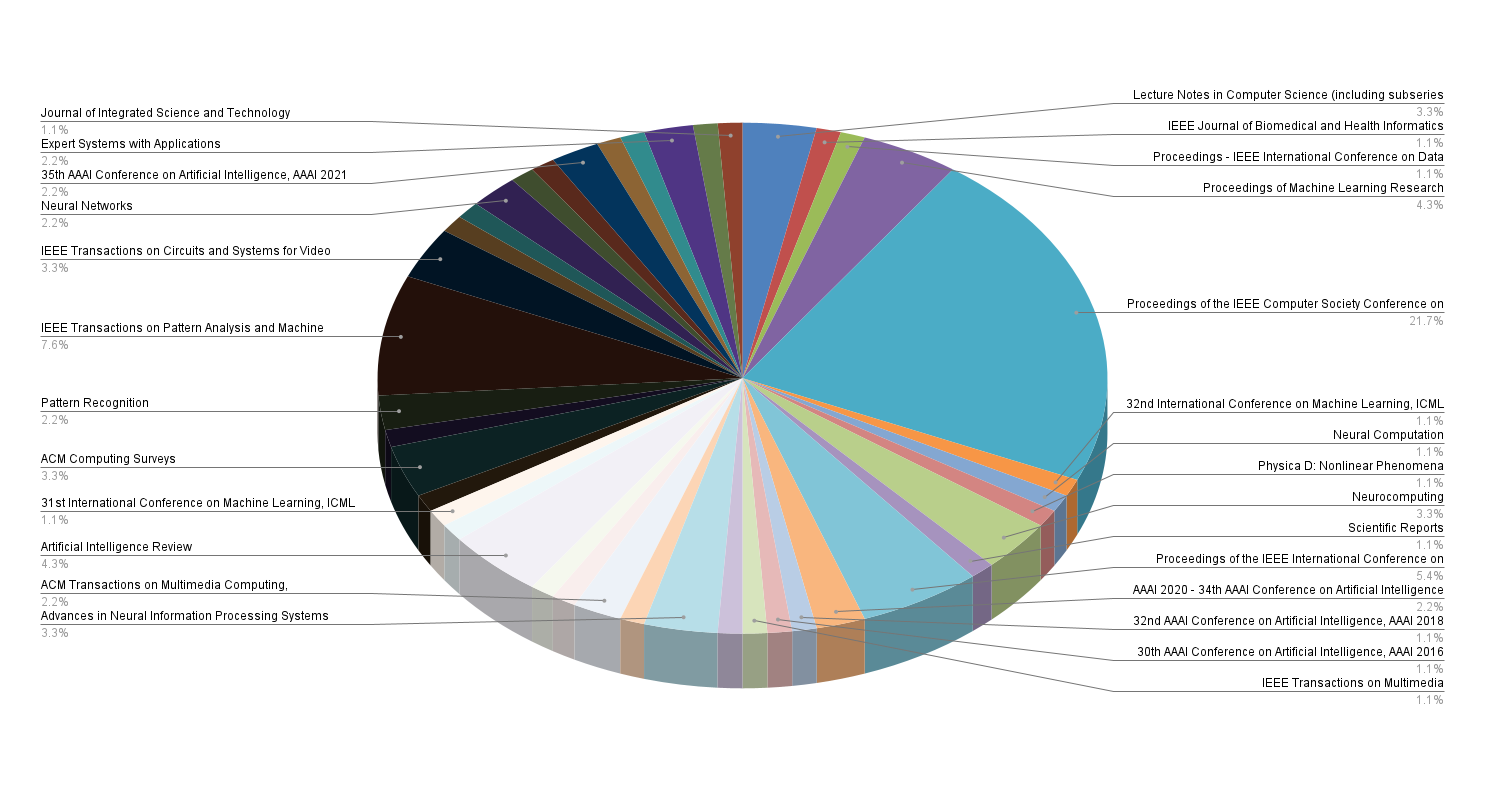}
                \caption{Documents by sources}
                \label{fig:sources}
            \end{subfigure}
            \caption{}
        \end{figure}
    }

    {
        \begin{figure}[H]
            \centering
            \begin{subfigure}[b]{0.45\textwidth}
                \centering
                \includegraphics[width=\textwidth]{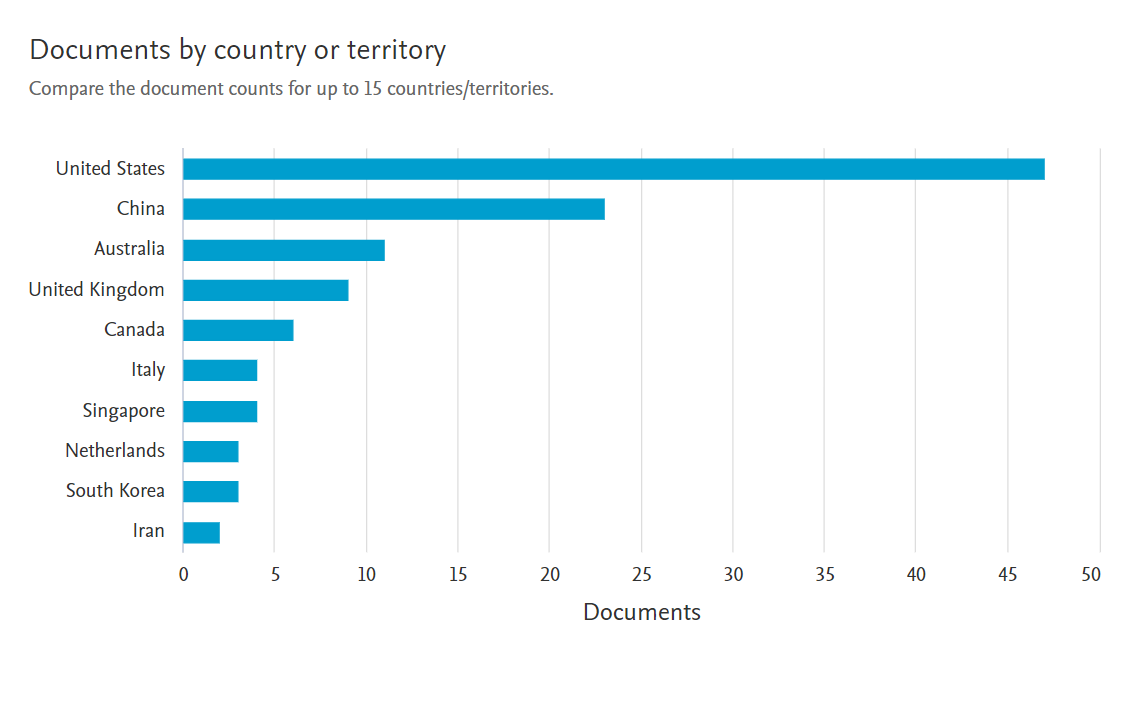}
                \caption{Documents by territories}
                \label{fig:territories}
            \end{subfigure}
            \hspace{0.05\textwidth} 
            \begin{subfigure}[b]{0.45\textwidth}
                \centering
                \includegraphics[width=\textwidth]{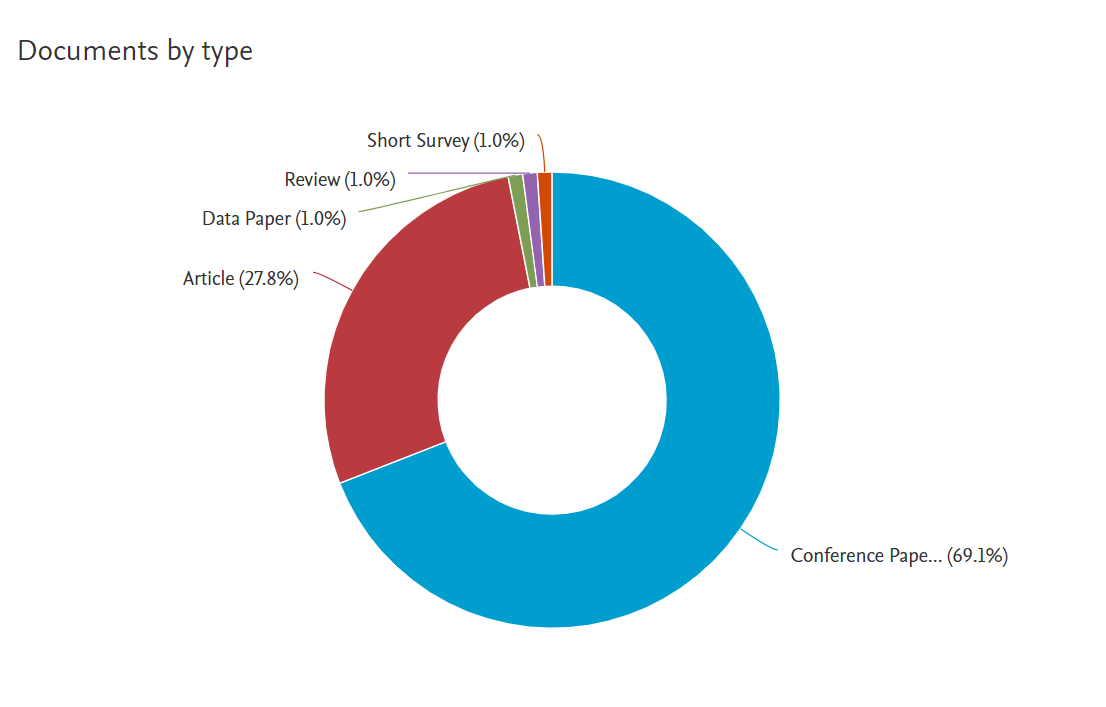}
                \caption{Documents by types}
                \label{fig:types}
            \end{subfigure}
            \caption{}
        \end{figure}
    }

    \begin{figure}[H]
        \centering
        \begin{subfigure}[b]{0.45\textwidth}
            \centering
            \includegraphics[width=\textwidth]{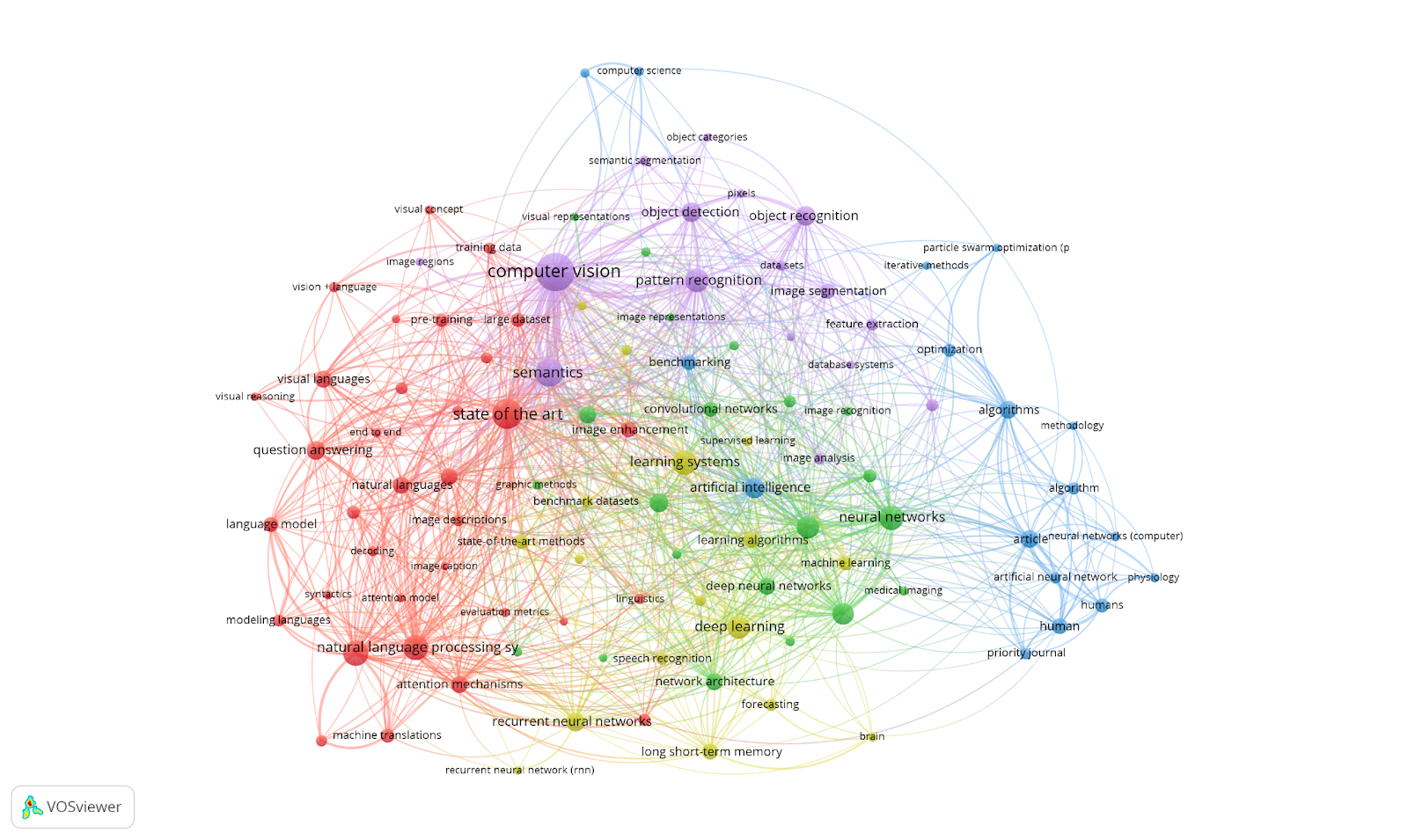}
            \caption{Keywords for image captioning}
            \label{fig:keywords}
        \end{subfigure}
        \hspace{0.05\textwidth} 
        \begin{subfigure}[b]{0.45\textwidth}
            \centering
            \includegraphics[width=\textwidth]{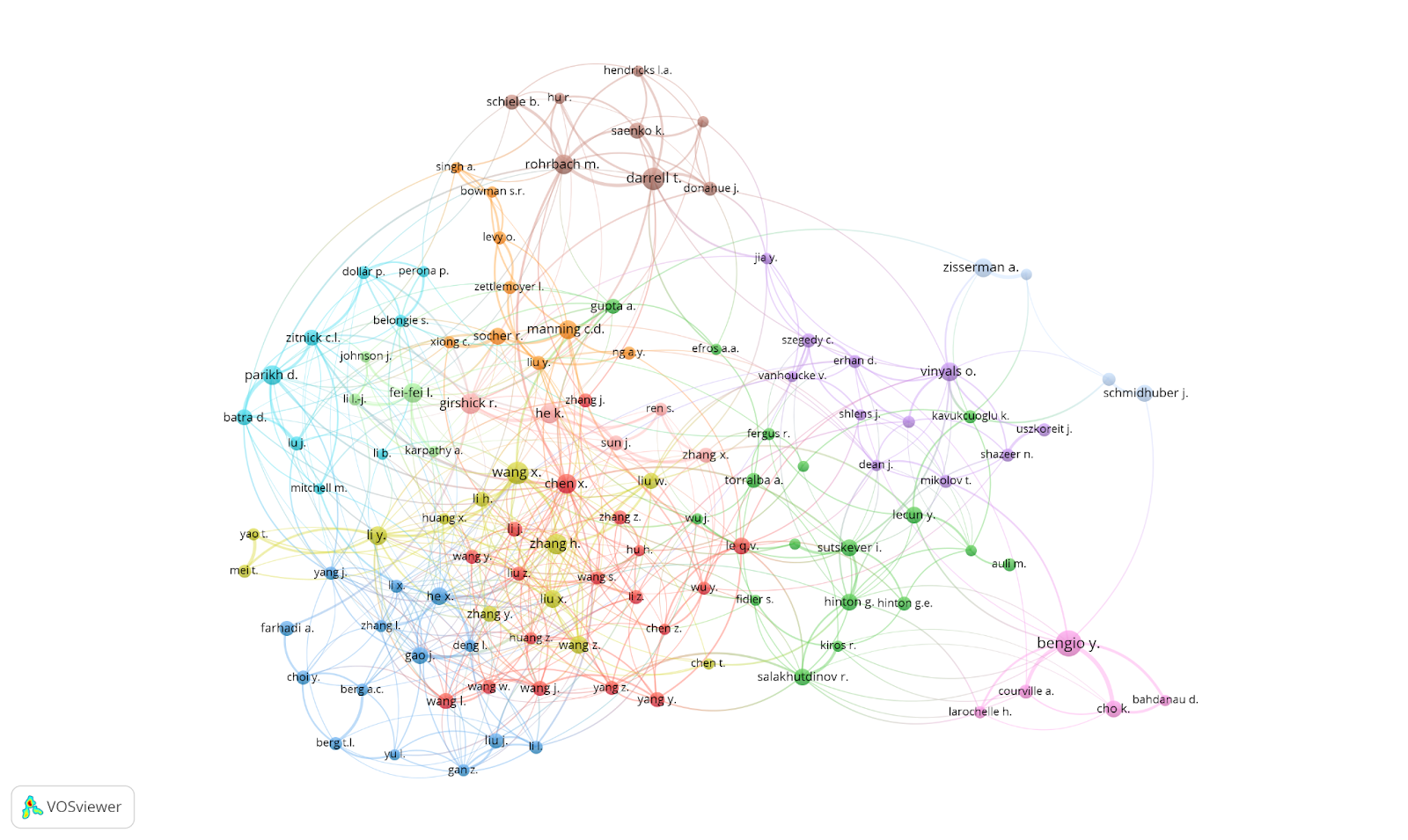}
            \caption{Co-authorship by network visualization}
            \label{fig:coauthor}
        \end{subfigure}
        \caption{}
    \end{figure}

     \begin{figure}[H]
        \centering
        \begin{minipage}{0.45\textwidth}
            \centering
            \includegraphics[width=\linewidth]{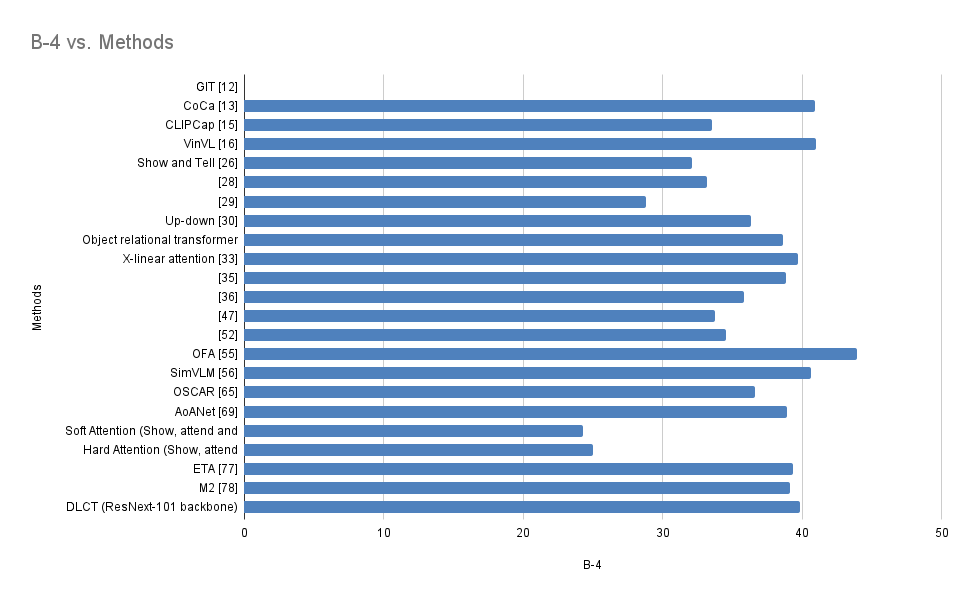}
            \caption{B-4 vs Methods}
            \label{fig:b4}
        \end{minipage}
        \hfill
        \begin{minipage}{0.45\textwidth}
            \centering
            \includegraphics[width=\linewidth]{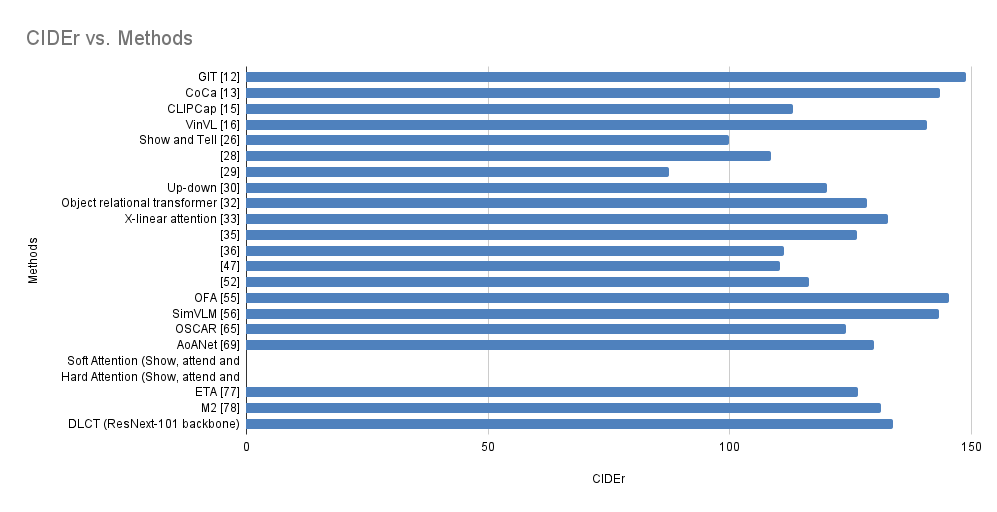}
            \caption{CIDEr vs Methods}
            \label{fig:cider}
        \end{minipage}
        
    \end{figure}

{
\begin{table}[h!]
\centering
\begin{tabular}{|m{3cm}|m{2cm}|m{2cm}|m{2cm}|m{2cm}|m{2cm}|m{2cm}|}
\hline
\multicolumn{1}{|c|}{} &\multicolumn{5}{|c|}{\textbf{Evaluation Metric}} \\
\hline
\textbf{Methods} & \textbf{BLEU-4} & \textbf{CIDEr} &\textbf{METEOR}& \textbf{ROGUE-L}& \textbf{SPICE}\\
\hline
GIT \cite{wang2022gitgenerativeimagetotexttransformer}  & - & 148.8 & - & - & - \\
\hline
CoCa \cite{yu2022cocacontrastivecaptionersimagetext} & 40.9 & 143.6 & 33.9 & - & 24.7 \\
\hline
CLIPCap\cite{mokady2021clipcapclipprefiximage} &  33.53 & 113.08 & 27.45 & - & 21.05 \\
\hline
VinVL\cite{zhang2021vinvlrevisitingvisualrepresentations} &  41.0 & 140.9 & 31.1 & - & 25.4 \\
\hline
Show and Tell \cite{Vinyals_2017} &  32.1 & 99.8 & 25.7 & - & - \\
\hline
\cite{lu2017knowinglookadaptiveattention} &  33.2 & 108.5 & 26.6 & - & - \\
\hline
\cite{pedersoli2017areasattentionimagecaptioning} &  28.8 & 87.4 & 23.7 & - & - \\
\hline
Up-down\cite{anderson2018bottomuptopdownattentionimage}  & 36.3 & 120.1 & 27.7 & 56.9 & 21.4 \\
\hline
Object relational transformer\cite{herdade2020imagecaptioningtransformingobjects}  &  38.6 & 128.3 & 28.7 & 58.4 & 22.6 \\
\hline
X-linear attention\cite{pan2020xlinearattentionnetworksimage}  &  39.7 & 132.8 & 29.5 & 59.1 & 23.4 \\
\hline
\cite{guo2020normalizedgeometryawareselfattentionnetwork} &  38.8 & 126.3 & 29.0 & 58.7 & - \\
\hline
\cite{WANG2020107075} & 35.8 & 111.3 & 27.8 & 56.4 & - \\
\hline
\cite{ZHANG202143}&  33.72 & 110.30 & 27.45 & - & 20.64 \\
\hline
\cite{cho2021unifyingvisionandlanguagetaskstext}&  34.5 & 116.5 & 28.7 & - & 21.9 \\
\hline
OFA \cite{wang2022ofaunifyingarchitecturestasks} &  43.9 & 145.3 & 31.8 & - & 24.8 \\
\hline
SimVLM\cite{wang2022simvlmsimplevisuallanguage}  &  40.6 & 143.3 & 33.7 & - & 25.4 \\
\hline
OSCAR \cite{li2020oscarobjectsemanticsalignedpretraining} &  36.58 & 124.12 & 30.4 & - & 23.17 \\
\hline
AoANet\cite{9008770}  &  38.9 & 129.8 & 29.2 & 58.8 & 22.4 \\
\hline
Soft Attention (Show, attend and tell)\cite{10.5555/3045118.3045336}  &  24.3 & - & 23.90 & - & - \\
\hline
Hard Attention (Show, attend and tell)\cite{10.5555/3045118.3045336}  &  25.0 & - & 23.04 & - & - \\
\hline
ETA \cite{9008532} &  39.3 & 126.6 & 28.8 & 58.9 & 22.7 \\
\hline
M2 \cite{cornia2020meshedmemorytransformerimagecaptioning} &  39.1 & 131.2 & 29.2 & 58.6 & 22.6 \\
\hline
DLCT (ResNext-101 backbone)\cite{luo2021dual}  &  39.8 & 133.8 & 29.5 & 59.1 & 23.0 \\
\hline
\end{tabular}
\vspace{+5pt}
\caption{\label{tab-7}Summarizing various image captioning models reviewed, listing the performance on key evaluation metrics on MSCOCO dataset with Karpathy test split.}
\label{tab:image_captioning_models}
\end{table}
    
}

\newpage
\section{Conclusion}
{
In this paper we comprehensively reviewed the progress and trend in image captioning. We covered various methodologies which have richly contributed towards more sophisticated caption generation and feature extraction. The key contributing factors have been attention mechanisms and transformers, more recently the trend is towards multimodal vision-language models after the great success of large language models. We also surveyed the literature for application of image captioning in the medical domain.

    \textit{\textbf{Note}: Here we answer RQ4, Considering the evaluation metrics employed, which image captioning models demonstrate superior performance?}
    
    We observe that the up-down attention mechanism has been the most widely used in various image captioning models. The transformer architecture has improved the quality of captions and out of the various transformer based models we reviewed the evaluation metrics(B4, CIDER,METEOR,SPICE) of DLCT\cite{luo2021dual} are higher than the other transformer based models, as is shown in table ~\ref{tab-2}. VLP based models are not only task-agnostic but also show enhanced performance on various evaluation metrics as shown in table ~\ref{tab-4}. We also observed that unsupervised, self-supervised and reinforcement learning models show comparable performance with respect to supervised models. CLIP has emerged to be a popular framework for self-supervised image captioning models. In terms of multi-style image captioning models, among the various methods we reviewed it was observed that MemCap shows higher scores in evaluation metrics, with reference to table ~\ref{tab-5}. Finally in the table ~\ref{tab-7} we compare the various models based on their score on key evaluation metrics as discussed in section ~\ref{eval-met}, it was observed that OFA showed higher B-4 scores, GIT performed higher in CIDER scores and CoCa had higher METEOR scores, a general trend shows that VLP based models show superior performance than the other reviewed models.  
    
    Lastly, we looked at various datasets and evaluation metrics for image captioning tasks. We observed that the future research trend will be towards more scalable and computationally efficient vision-language models and domain specific VL models.
    
}

}

\bibliographystyle{plain} 
\bibliography{main}

\end{document}